\documentclass[10pt,twocolumn,letterpaper]{article}

\usepackage{wacv}
\usepackage{times}
\usepackage{epsfig}
\usepackage{graphicx}
\usepackage{amsmath}
\usepackage{amssymb}
\usepackage{enumitem}
\usepackage{subcaption}
\usepackage{txfonts}
\usepackage{multirow}
\usepackage{array}
\usepackage{adjustbox}
\usepackage{authblk}

\newcommand\bstrut{\rule[-1.6ex]{0pt}{0pt}}
\newcommand\tstrut{\rule{0pt}{2.6ex}}
\newcolumntype{C}[1]{>{\centering\let\newline\\\arraybackslash\hspace{0pt}}m{#1}}

\usepackage[breaklinks=true, bookmarks=false]{hyperref}

\wacvfinalcopy 

\def\wacvPaperID{327} 

\ifwacvfinal\fi
\begin{document}

\title{End-To-End Trainable Video Super-Resolution Based on a New Mechanism for Implicit Motion Estimation and Compensation}


\newcommand*\samethanks[1][\value{footnote}]{\footnotemark[#1]}

\author{Xiaohong Liu$^1$ \quad Lingshi Kong$^1$ \quad Yang Zhou$^2$\quad Jiying Zhao$^2$\quad Jun Chen$^1$\thanks{Corresponding Author} \\$^1$McMaster University \quad $^2$University of Ottawa\\
	{\textit {\{liux173, kongl3, chenjun\}@mcmaster.ca \quad \{yzhou152, jzhao\}@uottawa.ca}}}

\maketitle
\thispagestyle{empty}
\begin{abstract}
	Video super-resolution  aims at generating a high-resolution  video from its low-resolution counterpart. With the rapid rise of deep learning, many recently proposed video super-resolution methods  use convolutional neural networks in conjunction with explicit motion compensation to capitalize on statistical dependencies within and across low-resolution frames. Two common issues of such methods are noteworthy. Firstly, the quality of the final reconstructed HR video is often very  sensitive to the accuracy of motion estimation. Secondly, the warp grid needed for motion compensation, which is specified by the two flow maps delineating pixel displacements in horizontal and vertical directions, tends to introduce additional errors and jeopardize the temporal consistency across video frames. To address these issues, we propose a novel dynamic local filter network  to perform implicit  motion estimation and compensation by employing, via locally connected layers, sample-specific and position-specific dynamic local filters that are tailored to the target pixels. We also propose a global refinement network  based on ResBlock and autoencoder structures to exploit non-local correlations and enhance the spatial consistency of super-resolved frames. The experimental results demonstrate that the proposed method outperforms the state-of-the-art, and validate its strength in terms of local transformation handling, temporal consistency as well as edge sharpness.
\end{abstract}
\section{Introduction}
\begin{figure}[htp]
	\begin{center}
		\begin{minipage}[h]{0.325\linewidth}
			\begin{center}
				\includegraphics[width=\textwidth]{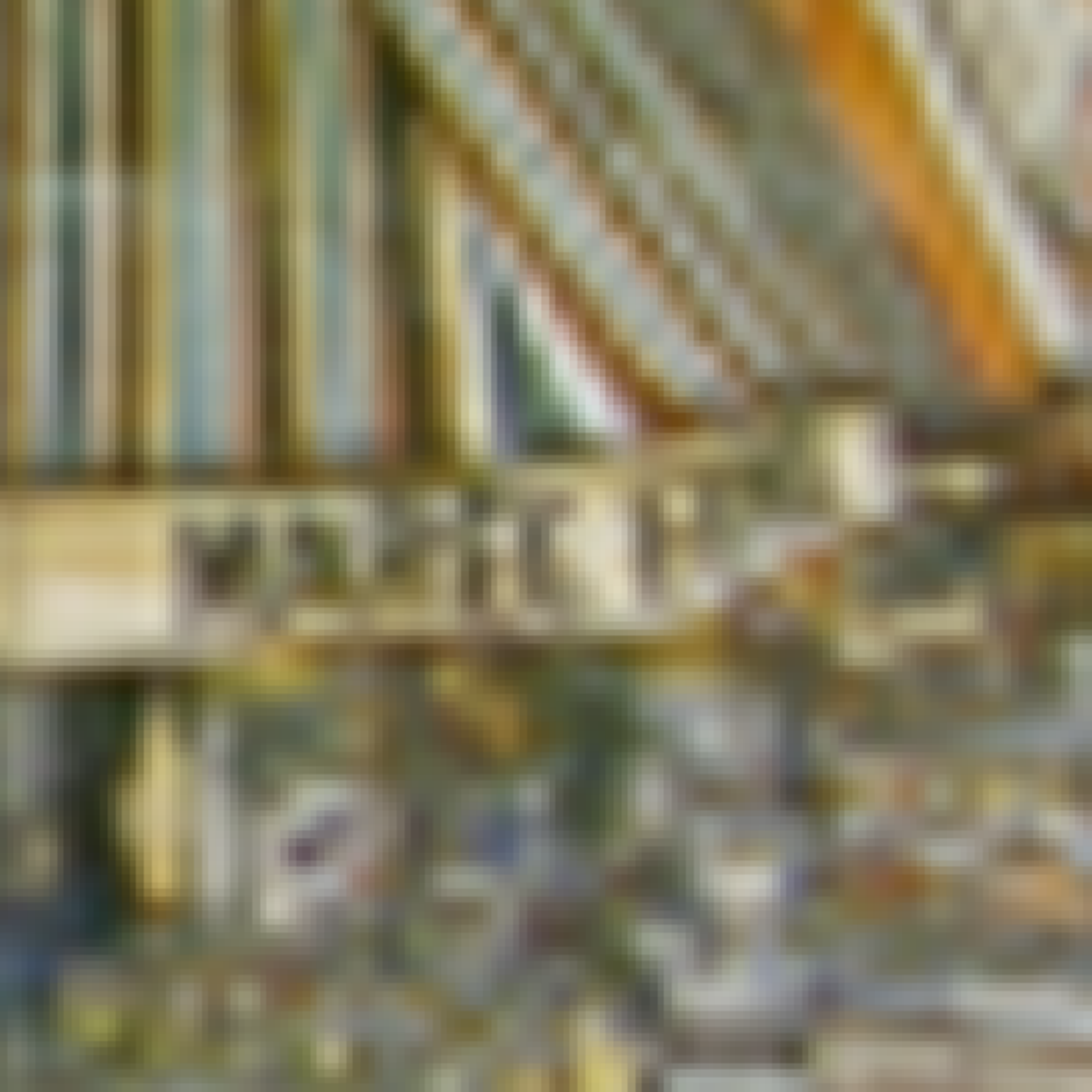}\\
				\scriptsize{(a) Bicubic}
			\end{center}
		\end{minipage}
		\begin{minipage}[h]{0.325\linewidth}
			\begin{center}
				\includegraphics[width=\textwidth]{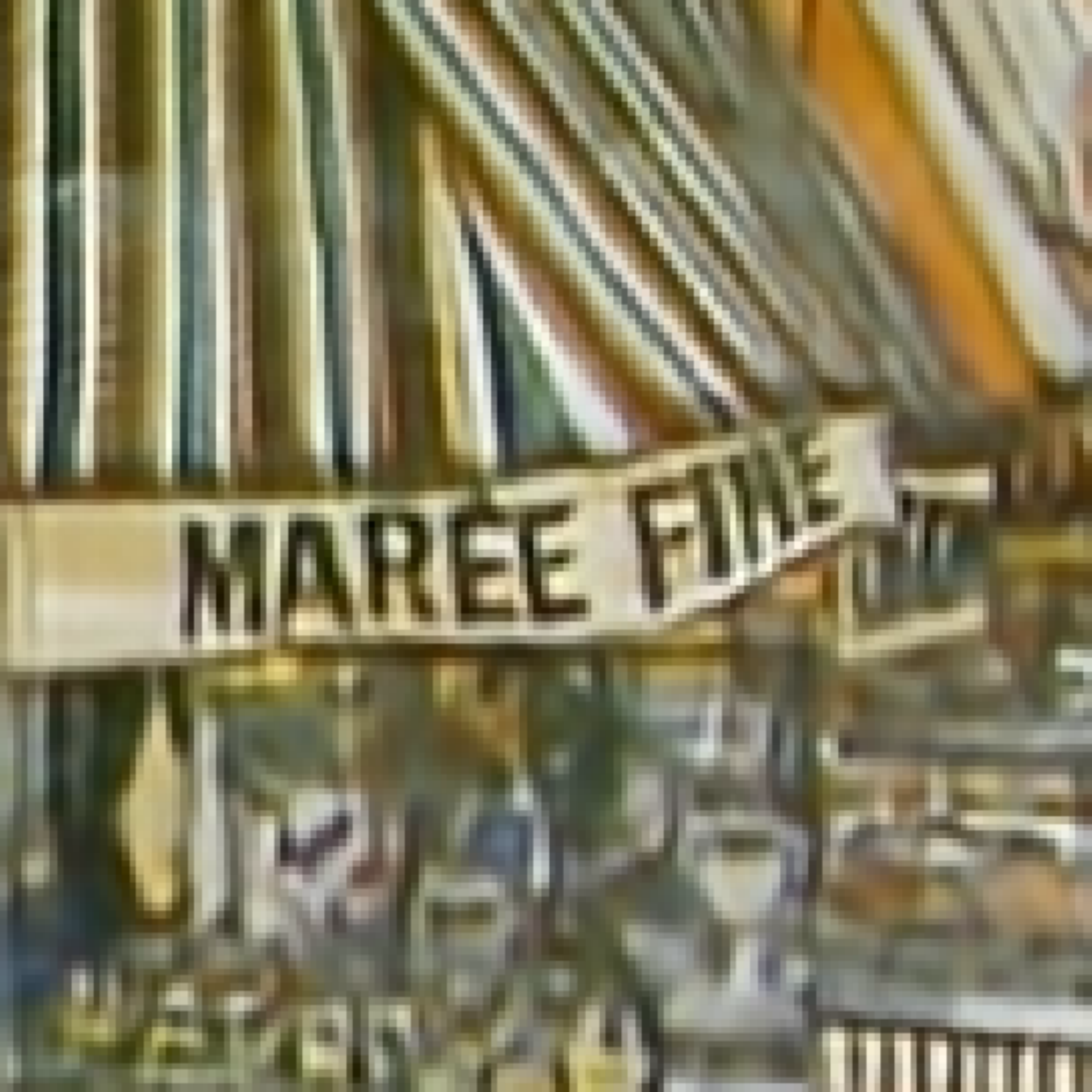}\\
				\scriptsize{(b) Our Result}
			\end{center}
		\end{minipage}
		\begin{minipage}[h]{0.325\linewidth}
			\begin{center}
				\includegraphics[width=\textwidth]{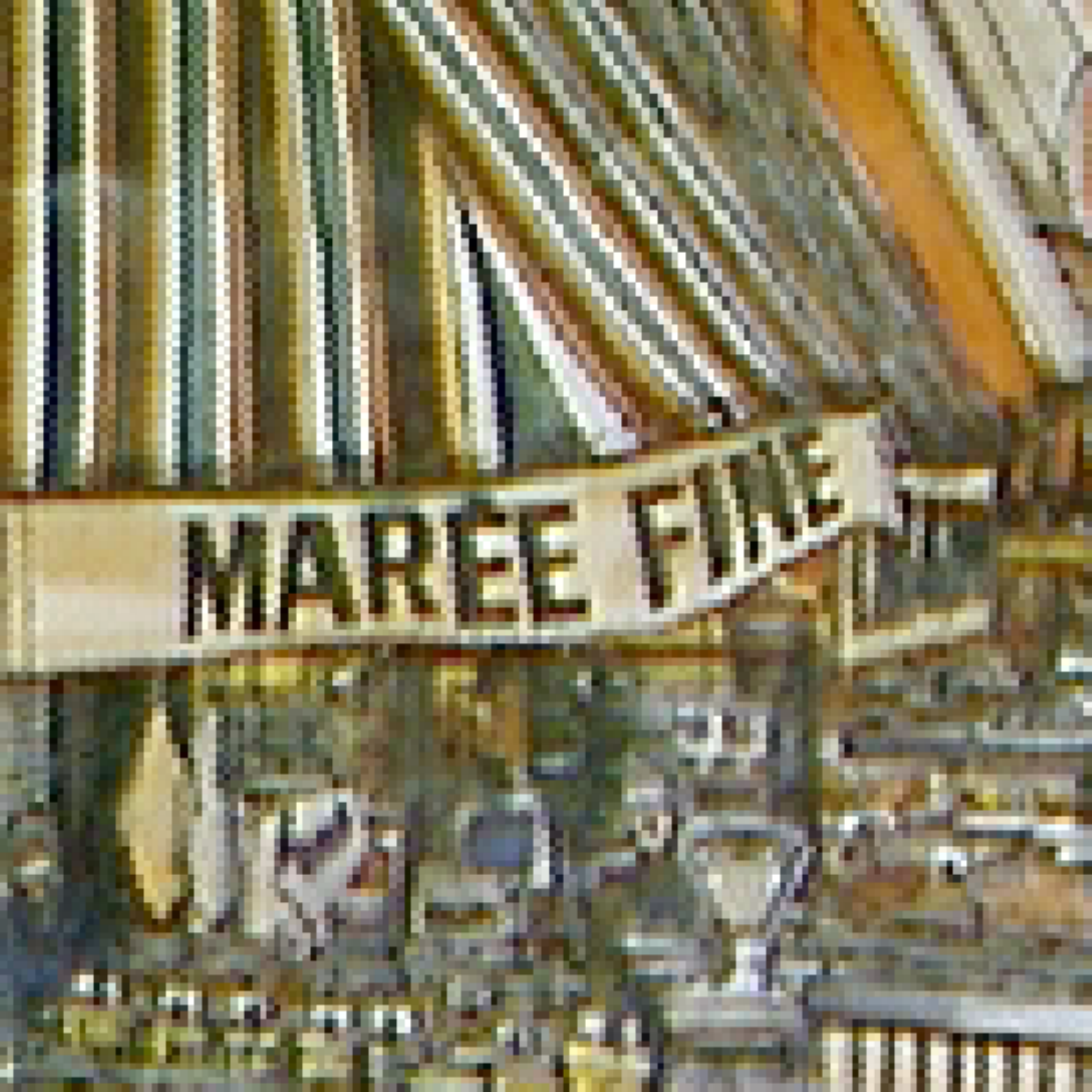}\\
				\scriptsize{(c) Ground Truth}
			\end{center}
		\end{minipage}
	\end{center}
	\caption{The comparison of the bicubic interpolation, our result and the ground truth with the scale ratio set to 4.}
	\label{fig:simple_comparison}
\end{figure}
Super-resolution (SR) is considered as a promising technique to produce high-resolution (HR) pictorial data using low-resolution (LR) sensors without resorting to hardware upgrades. Over the past few decades, it has received significant attention in a wide range of areas, including, among others, medical imaging \cite{MedicalImaging1,MedicalImaging2}, satellite imaging \cite{satellite1,satellite2,satellite3} and surveillance \cite{surveillance1,surveillance2,surveillance3}. Recently, it has also been used as a pre-processing step to facilitate various recognition tasks by enhancing the raw data \cite{recoginition1,recoginition2}. Super-resolution can be divided into two categories: single image super-resolution (SISR) and video super-resolution (VSR). 
SISR can be viewed as a certain sophisticated image interpolation operation, which attempts to supply the missing details by strategically exploiting the spatial patterns in LR inputs. In contrast, VSR takes advantage of both spatial and temporal relationships among consecutive frames to improve the quality of reconstructed videos. The traditional approaches to the VSR problem typically consist of three  sub-tasks: sub-pixel motion estimation, motion compensation and up-sampling \cite{farsiu2004fast,ma2015handling,tao2017detail,liu2018robusttip}. In general, motions across LR frames are estimated explicitly and the estimated LR displacements are employed to compensate sub-pixel motions by warping relevant LR frames to the target frame; the compensated LR frames are then fused to reconstruct the corresponding HR frame. The existing deep-learning-based VSR methods largely follow similar approaches \cite{kappeler2016video, caballero2017real, liu2017robust, sajjadi2018frame,zhou2018video}. One common limitation of such methods is that the motion compensation module is not trainable, i.e., it cannot be updated through the  training process. It is also worth noting that the super-resolved frames are very sensitive to the accuracy of the initial motion estimation, rendering the quality of SR outputs unstable.

In this paper, we propose a new approach to VSR using local dynamic filters via locally connected (LC) layers for implicit motion compensation and demonstrate its competitive advantages over the existing ones. The effectiveness of this LCVSR approach can be attributed to three major factors: \textit{1)} The overall system is end-to-end trainable and does not require any pre-training; the accuracy of motion estimation improves progressively through the training process. \textit{2)} Local motion estimation and compensation is performed implicitly by a novel dynamic local filter network (DLFN) with LC layers. There are at least two benefits of using the DLFN. Firstly, the implicit motion estimation, realized by sample-specific and position-specific dynamic local filters  generated on-the-fly according to the target pixels, can deal with complicated local transformations in video frames such as regional blurring, irregular local movement and photometric changes. Secondly, the simultaneous action of dynamic local filters on all input LR frames via LC layers helps to maintain the temporal consistency. \textit{3)} The spatial consistency of super-resolved outputs is enforced by a novel global refinement network (GRN) constructed using ResBlock and autoencoder structures. Since the implicit motion estimation performed by the DLFN is spatially localized, it may cause inconsistencies across neighboring areas. As such, the GRN  plays a critical role of restoring the spatial consistency. Moreover, the GRN has the capability of exploiting non-local correlations due to its constituent autoencoder structure, which makes up for the lack of global motion estimation in the DLFN. Fig.~\ref{fig:simple_comparison} shows the comparison of the bicubic interpolation (the LR input), our result (the HR output) and the ground truth for a sample video frame.
\section{Related Work}
Many SISR and VSR methods have been proposed over the past few decades. The traditional methods typically solve the SR problem, which is inherently under-determined, by formulating it as a certain regularized optimization problem \cite{farsiu2004fast, elad2001fast, kohler2016robust, liu2014bayesian, ma2015handling, liu2017robustsip, liu2018robusttip}. The recent years, however, have witnessed the increasing dominance of deep-learning-based SR methods. The work by Dong \etal \cite{dong2014learning} is among the earliest ones that brought convolutional neural networks (CNNs) to bear upon SISR. In their proposed SRCNN, a very shallow network is used to extract LR features, which are subsequently leveraged to generate HR images via non-linear mapping. To avoid time-consuming operations in the HR space, Shi \etal \cite{shi2016real} proposed an efficient sub-pixel convolution network (ESPCN) to extract and map features from the LR space to the HR space using convolutional layers instead of naive pre-defined interpolations such as bilinear or bicubic. Zhang \etal \cite{zhang2018residual} designed a residual dense block with direct connections for the purpose of a more thorough extraction of local features from LR images.

Compared with SISR, VSR is inherently more complex due to the additional challenge of harnessing the relevant information in the temporal domain. To cope with this challenge, Kappeler \etal \cite{kappeler2016video} proposed to employ the handcrafted optical flow method by Drulea and Nedevschi \cite{drulea2011total} to compensate motions across input frames and then feed the compensated frames into a pre-trained CNN to perform the SR operation. Huang \etal \cite{huang2018video} developed a new VSR method based on the so-called bidirectional recurrent convolutional network (BRCN). The BRCN is a variant of recurrent neural network (RNN) with commonly-used recurrent connections replaced by 
weight-sharing convolutional connections; as a consequence, it inherits the strength of RNN in terms of capturing long-term temporal dependencies and at the same time admits a more efficient implementation. Liao \etal \cite{liao2015video} introduced a SR draft-ensemble approach in which multiple SR drafts are generated using an optical flow method with different estimation settings and then synthesized by a carefully constructed CNN to produce the final HR output.  

To avoid inaccurate motion estimation caused by fixed temporal radius, Liu \etal \cite{liu2017robust} proposed a temporal adaptive neural network. This network has several SR inference branches, each with a different temporal radius; the final HR output is obtained by adaptively aggregating all SR inferences. They also introduced a spatial alignment network that can efficiently estimate motions between neighboring frames.
The VESPCN developed by Caballero \etal \cite{caballero2017real} combines spatio-temporal networks with an end-to-end trainable spatial transformer module to generate the super-resolved video. Based on the motion compensation module in the VESPCN \cite{caballero2017real}, Tao \etal \cite{tao2017detail} designed a sub-pixel motion compensation (SPMC) layer to compensate motion and up-sample video frames simultaneously. Moreover, they advocated the use of an encoder-decoder style structure with ConvLSTM \cite{xingjian2015convolutional} and skip-connections \cite{mao2016image} for effectively processing sequential videos and reducing the training time. 

To improve the temporal consistency of super-resolved videos, Sajjadi \etal \cite{sajjadi2018frame} proposed a frame recurrent VSR method, which enables the processing of the current frame to benefit from the inferred SR results for the previous frames. This method is more efficient than those treating the VSR problem as a sequence of multi-frame SR problems due to the recurrent nature of its operations.  

Jo \etal \cite{jo2018deep} developed a novel VSR method, known as VSRDUF, which works as follows: A deep neural network is employed to generate dynamic upsampling filters and frame residuals; certain provisional HR frames are constructed from their LR counterparts through dynamic upsampling filters, and the inferred residuals are then added to such frames to produce the final output. This work is most related to ours in the sense that motion compensation is only performed implicitly. However, it will be seen that the underlying mechanisms are fundamentally different. 
\begin{figure*}[t]
	\centering
	\includegraphics[width=0.9\linewidth]{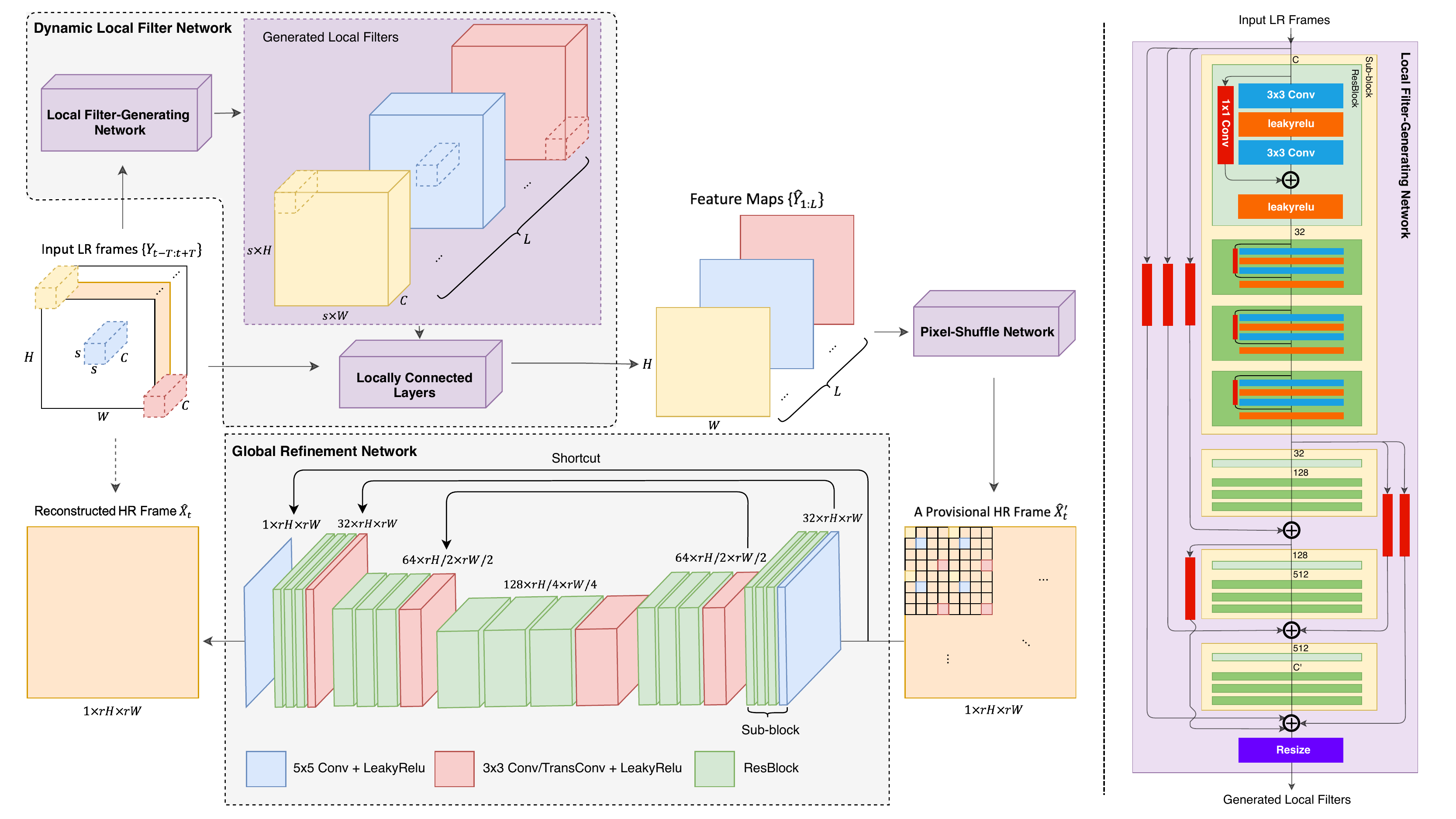}
	\caption{The left part shows the overall architecture of the proposed LCVSR system. The right part provides a detailed illustration of the LFGN.}
	\label{fig:VSR}
	\vspace{-2mm}
\end{figure*}

\section{Method}
In this section we give a detailed description of the proposed VSR method with an emphasis on its most prominent feature, namely, the use of local dynamic filters via LC layers to implicitly compensate motions across video frames.  The process starts with converting the input LR frames from RGB to YCbCr color space. Only the Y channels are fed into the proposed VSR system, which helps to reduce the computational complexity. The Cb and the Cr channels are upsampled via bicubic interpolation and merged with the super-resolved Y channels to generate HR frames in YCbCr, from which the final result in RGB is obtained. 
Let $Y_t \in\mathbb{R}^{H\times W}$  denote the $t$-th LR frame in the $Y$ channel degraded by blurring and down-sampling operations from the corresponding HR frame $X_t \in \mathbb{R}^{rH\times rW}$, where $r$ is the scale ratio. The given LR video sequence of $C$ consecutive frames centered at $Y_t$ is denoted as $\{Y_{t-T : t+T}\}$, where $T$ is the temporal radius and $C=2T+1$. The corresponding HR video sequence is $\{X_{t-T : t+T}\}$. We use $F_{LCVSR}$ as the functional representation of the proposed LCVSR system with the end-to-end relation
\begin{equation}
\hat{X}_t = F_{LCVSR}(\{Y_{t-T:t+T}\}; \theta_{LCVSR}),
\end{equation}
where $\hat{X}_t$ is the reconstructed HR frame and $\theta_{LCVSR}$ denotes the ensemble of the system parameters. Regardless of the batch size, the shape of the input tensor is set to be $C\times H\times W$ while that of the output is set to be $1 \times rH \times rW$. The proposed LCVSR system, as shown in Fig.~\ref{fig:VSR}, consists of three modules, which are respectively the dynamic local filter network (DLFN), the pixel-shuffle network \cite{shi2016real} and the global refinement network (GRN). 
The input LR frames $\{Y_{t-T:t+T}\}$ are first fed into the DLFN, which has  two sub-modules: local filter-generating network (LFGN) and LC layers. The LFGN produces sample-specific and position-specific dynamic local filters filters  on-the-fly according to the spatio-temporal relationship among the inputs. 
These dynamic local filters, each of size $s\times s\times C$ (with $s=2d+1$, where $d$ is the spatial radius), then act on the LR frames via LC layers to generate the feature maps $\{\hat{Y}_{1:L}\}$ (we set $L=r^2$ in this work). 
These feature maps are forwarded to the pixel-shuffle network to construct a provisional HR frame $\hat{X}'_t$, which is subsequently fed into the GRN to enhance the spatial consistency and cope with global transformations. 
The output of the GRN is the reconstructed HR frame $\hat{X}_t$. 
\subsection{Dynamic Local Filter Network} \label{3.1}
The existing VSR methods typically compensate motions across video frames by explicitly estimating pixel displacements in horizontal and vertical directions. This error-prone estimation step may potentially jeopardize the quality of SR results. Therefore, it is of considerable interest to develop deep-learning-based techniques for implicit motion estimation and compensation. One possible approach is to use the conventional CNNs with weight-sharing filters, which have been shown to achieve outstanding performance in image classification and segmentation tasks\cite{he2017mask, segnet2017}. However,  motion, blur and photometric changes encountered in the VSR problem are usually sample-specific and position-specific, in other words, each pixel in a video frame may exhibit a unique degradation pattern, which cannot be effectively exploited by the weight-sharing filters. For this reason, we propose a DLFN with LC layers that can perform local operations tailored to the spatio-temporal characteristics of the target pixels. Specifically, a sample-specific and position-specific dynamic local filter is generated for each pixel in the input LR frames; these dynamic local filters then collectively act on the input frames via LC layers to generate feature maps by compensating motions and other transforms in an implicit manner. 


Let $\Theta_l\in \mathbb{R}^{sH\times sW \times C}$ denote the $l$th set of (unbiased) local filters. Each local filter in $\Theta_l$  (say, $\Theta_{i,j,l}\coloneqq\{\Theta^{(m,n,k)}_{i,j,l}:m,n=1,2,3; k=1,\cdots,C\}$) is associated with a specific pixel (say, the $(i,j)$-pixel) in the $t$th LR frame.
The $l$th feature map $\hat{Y}_{l}$ is obtained by applying $\Theta_{l}$ on the input LR frames $\{Y_{t-T:t+T}\}$. More precisely, we have
\begin{equation}
\hat{Y}_l^{(i,j)} = \sum_{m=i-d}^{i+d}\sum_{n=j-d}^{j+d}\sum_{k=t-T}^{t+T}\Theta_{i,j,l}^{(m-i+d+1,n-j+d+1,k-t+T+1)}\cdot Y_{k}^{(m,n)},
\end{equation}
where   $\hat{Y}_l^{(i,j)}$ represents the value of the $(i, j)$-pixel in the $l$th feature map, and $Y_{k}^{(m,n)}$ denotes the $(m,n)$-pixel  in the $k$th LR frame. It is worth pointing out that both $\Theta_{l}$ and $\hat{Y}_{l}$ depend on $t$ and should actually be written as $\Theta_{t,l}$ and $\hat{Y}_{t,l}$ respectively; here we suppress the subscript $t$ for notational simplicity.

To generate dynamic local filters, we build a novel LFGN based on ResBlocks \cite{he2016deep}. Its input-output relationship can be expressed as
\begin{equation}
\Theta = F_{LFGN}(\{Y_{t-T:t+T}\}; \theta_{LFGN}),
\end{equation}
where $F_{LFGN}$ is the functional representation of the LFGN and $\theta_{LFGN}$ denotes the ensemble of its parameters. Note that the output $\Theta \coloneqq \{\Theta_{1:L}\} \in \mathbb{R}^{C \times sH\times sW \times L}$ is a 4-D tensor (which consists of all dynamic local filters) whereas the input of the LFGN is a 3-D tensor of shape $C\times H\times W$. To generate $\Theta$ based on $\{Y_{t-T:t+T}\}$, we employ modified ResBlocks in concatenation to progressively increase the depth of the input tensor from $C$ to $C'$, where $C'= C \times s^2\times L$, then resize the resulting 3-D tensor of shape $C' \times H\times W$ to a 4-D tensor of shape ${C \times sH\times sW \times L}$. The LFGN consists of one Resize module and four sub-blocks, each of which is built using ResBlocks. We find that using more ResBlocks in each sub-block leads to better performance. However, to strike a balance between system performance and computational complexity, in each sub-block we deploy one ResBlock for depth enlargement and three ResBlocks with no shape change. Besides, grouped convolutions \cite{krizhevsky2012imagenet} are also utilized. The four sub-blocks are densely connected by shortcuts to facilitate information exchange among them. If the tensors at the two sides of a shortcut have different shapes, a $1 \times 1$ convolution is performed to make the shape compatible; otherwise, we directly connect the two sides without modification. Each ResBlock consists of two $3 \times 3$ convolutional layers, two LeakyReLU layers \cite{xu2015empirical} and an inner shortcut. The output of the DLFN consists of $L$ feature maps, each of which is generated by exploiting, via implicit motion compensation, the relevant spatio-temporal information in all LR frames. These feature maps are then fed into the pixel-shuffle network to construct a provisional HR frame $\hat{X}'_t$.


 It is worth emphasizing that dynamic local filters are intermediate computational results produced within the proposed system and should not be viewed as the parameters of the system itself. They are generated 
 by the LFGN based on the input LR frames, then act back on the input frames, via LC layers, to perform pixel-level fine-grained motion estimation and compensation. More generally, this is an effective mechanism for leveraging the learning capability of a deep neural network (say, the LFGN in the current setting) to realize
 dynamic localized functionalities. See Sections 
 \ref{sec:visualization} and \ref{experiment_network_size} for some supporting experimental results.

\subsection{Global Refinement Network} \label{3.2}
Since the proposed DLFN performs localized motion estimation and compensation across LR frames,
 it can potentially cause inconsistencies among neighboring areas. To address this issue, we propose a GRN (see Fig.~\ref{fig:VSR}) employing ResBlock and autoencoder structures to improve the spatial consistency of super-resolved frames. The autoencoder structure enlarges the receptive field so that the GRN also has the ability to deal with global transformations, which makes up for the lack of global motion estimation in the DLFN. The GRN mainly consists of five sub-blocks connected by shortcuts. Each sub-block contains a convolutional layer or a transposed convolutional layer with LeakyReLu as the activation function, followed by three ResBlocks that are structurally the same as those in the DLFN. The encoder, formed by the second and third sub-blocks, reduces the spatial dimension but increases the depth dimension to enlarge the receptive field progressively. In contrast, the decoder, formed by the last two sub-blocks, reduces the depth dimension but increases the spatial dimension to perform global refinement. 
Finally, a $5 \times 5 $ convolutional layer activated by LeakyRelu produces the reconstructed  HR frame. The input-output relationship of the proposed GRN is given by
\begin{equation}
\hat{X}_t = F_{GRN}(\hat{X}'_t; \theta_{GRN}),
\end{equation}
where $F_{GRN}$ is the functional representation of the GRN and  $\theta_{GRN}$ denotes the ensemble of its parameters. 
\subsection{Data Preparation} \label{3.3}
Deep-learning-based VSR methods rely heavily on the quality and the quantity of the training datasets. Unfortunately, so far there is no standard training dataset for VSR.  To build our own, we totally collect 100k  ground-truth sequences, each with 7 consecutive frames of size $252 \times 444$, where 70k sequences are selected from the Vimeo-90k dataset recently built by Xue \etal \cite{xue17toflow} and the rest 30k sequences are extracted from several videos provided by Harmonic\footnote {https://www.harmonicinc.com/free-4k-demo-footage/}; as a comparison, the current state-of-the-art VSRDUF uses 160k sequences for training. We adopt the Vid4 dataset \cite{liu2014bayesian} and the SPMCS dataset \cite{tao2017detail} for testing. Our input LR frames are generated from the ground-truth sequences via Gaussian blur and downsampling. For the Gaussian blur, we set the standard deviation to be 1 and the kernel size to be $3 \times 3$. As to the downsampling operation, we choose the scale ratio $r=3,4$ (considered to be the most challenging cases in the VSR task).

\subsection{Implementation} \label{3.4}
The proposed LCVSR system is end-to-end trainable and no pre-training is needed for  sub-networks. Our training is carried out on a PC with two NVIDIA GeForce GTX 1080 Ti, but only one GPU is used for testing. We adopt Xavier initialization \cite{glorot2010understanding} and set the mini-batch size to be 12.
The $\mathcal{L}_2$ loss function is used to calculate the reconstruction error  as follows:
\begin{equation}
\mathcal{L}_2(X_t, \hat{X}_t) = \left \|X_t-\hat{X}_t\right \|^2_2.
\end{equation}
We train the proposed system for about 0.8 million iterations using the Adam optimizer \cite{kingma2014adam} with $\beta_1=0.9$, $\beta_2=0.999$. The learning rate is set to $10^{-4}$ at the beginning and decays to $10^{-5}$ after 0.7 million iterations. Our source code will be made publicly available. 
\section{Experimental Results}
In our experiments, we set $C=7$, $T=3$, $s=3$ and $d=1$. As such, one super-resolved frame is generated based on 7 consecutive LR frames with the middle one as the reference, and the size of generated dynamic local filters is  $3\times 3\times 7$. We use PSNR and SSIM for quantitative assessment of the SR results. All PSNR values are calculated based on the Y channel using the ITU-R BT.601 standard to make fair comparisons \cite{caballero2017real}. In addition to the aforementioned quantitative performance metrics, we also consider qualitative measures such as edge sharpness and temporal consistency. The following existing VSR methods are chosen as benchmarks: 
\textit{Bayesian} \cite{liu2014bayesian}, \textit{VSRNet} \cite{kappeler2016video}, \textit{VESPCN} \cite{caballero2017real}, \textit{$B_{1,2,3}+T$} \cite{liu2017robust}, \textit{SPMC} \cite{tao2017detail}, \textit{FRVSR} \cite{sajjadi2018frame} and \textit{VSRDUF} \cite{jo2018deep}. For the \textit{VSRDUF}, both its basic version with 16 layers (\textit{DUF-16L}) and the enhanced version with 52 layers (\textit{DUF-52L}) are used for comparisons. The quantitative experimental results of these benchmarks are obtained using the provided source codes (if available) or cited from the original papers.



\begin{figure*}[t]
	\centering
	\includegraphics[width=0.8\linewidth]{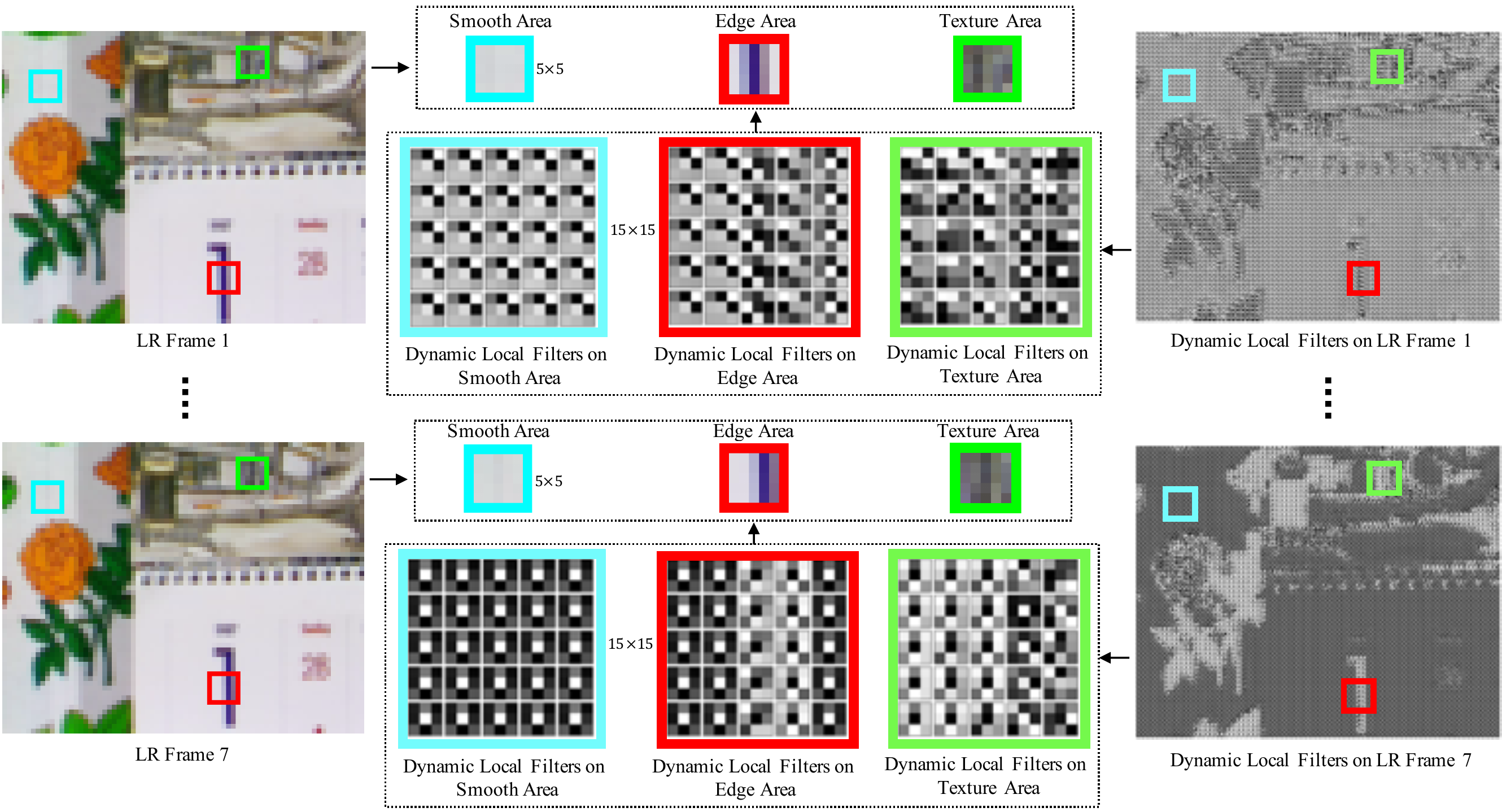}
	\caption{Dynamic local filters applied on smooth, edge and texture areas respectively.}
	\label{fig:DLF}
	\vspace{-4mm}
\end{figure*}
\begin{figure}[h]
	\centering
	\includegraphics[width=0.86\linewidth]{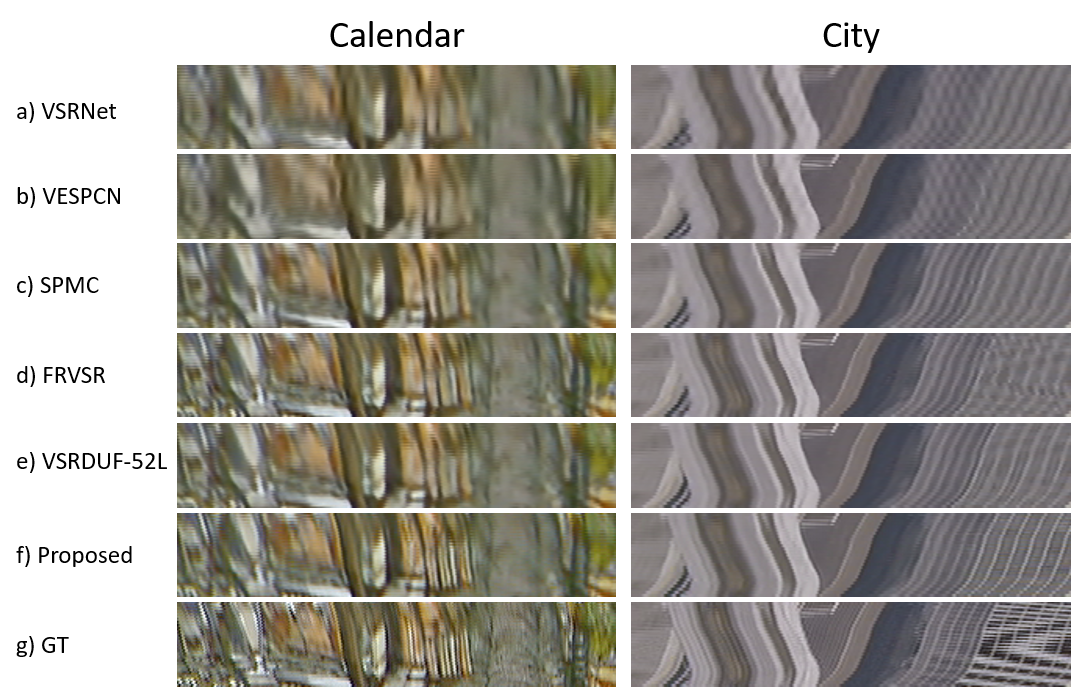}
	\caption{Visual comparison of the temporal profiles generated by the proposed VSR method and some existing ones for super-resolved \textit{Calendar} and \textit{City} frames in Vid4.}
	\label{fig:temporal}
	\vspace{-3mm}
\end{figure}

\subsection{Visualization of Dynamical Local Filters}\label{sec:visualization}
Although the performance of the proposed LCVSR system benefit from many contributing factors, arguably the most crucial one is the use of dynamic local filters, via LC layers, for implicit motion estimation and compensation. To gain a better understanding, it is instructive to distinguish generated filters from learned filters \cite{jia2016dynamic}. The learned filters such as those in the LFGN update themselves only during the training process and become static afterwards. On the contrary, the generated filters are adaptive in the sense that they are not fully specified until the input is given. The dynamic local filters in the proposed system belong to this category. They are computed based on the input LR frames and used as the kernel weights in LC layers; moreover, to reduce the computational complexity, they are applied to the LR space rather than the HR space. Fig.~\ref{fig:DLF} illustrates the dynamic local filters that are used to generate the first feature map $\hat{Y}_1$, shown as the yellow cube in Fig.~\ref{fig:VSR}. It also shows some sample patches of size $5\times5$ extracted from  smooth, edge and texture areas in the first and the last LR frames together with their corresponding dynamic local filters, which are of size $15\times15$ (since $s=3$). It can be seen from Fig.~\ref{fig:DLF} that the dynamic local filters are spatially content-adaptive within each frame and temporally distinct (even when the associated pixels are of similar nature) across different frames. This provides supporting evidence for their ability to adapt according to the spatio-temporal characteristics of the target pixels and perform implicit motion estimation and compensation. It can also been seen from 
Fig.~\ref{fig:DLF} that, in a given frame, the dynamic local filters for pixels with a homogeneous neighborhood tend to have similar patterns, which helps to retain intra-frame spatial consistency. For example, the local filters associated with the pixels in the smooth area and those in the edge area (except the exact edge pixels) are quite alike. In contrast, each pixel in the texture area has a distinct dynamic local filter, which is essential since this area is very sensitive to motions. 
\begin{figure*}[t]
	\begin{center}
		\begin{subfigure}[b]{0.14\textwidth}
			\includegraphics[width=\textwidth]{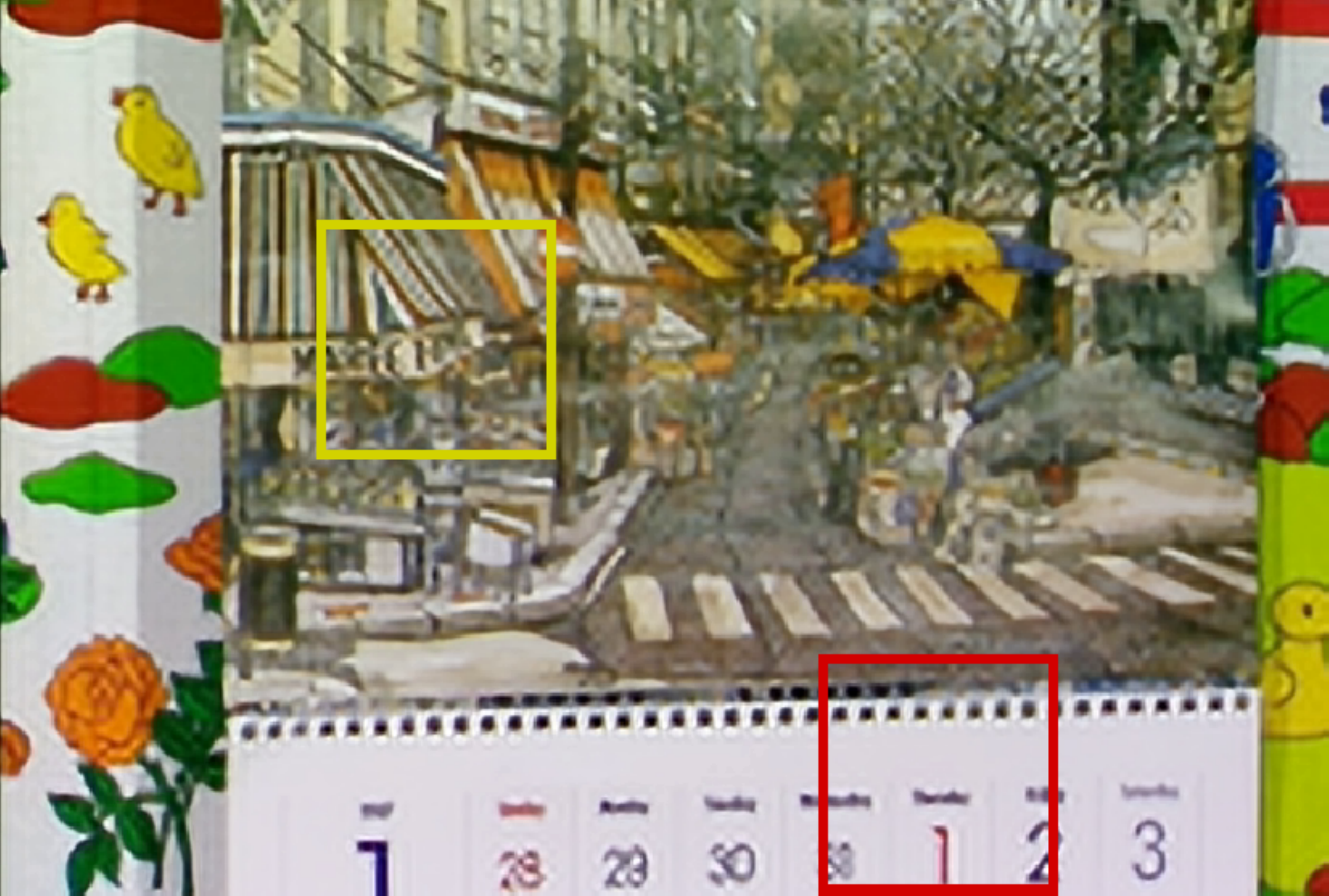}
			\label{fig:Calendar1}
		\end{subfigure}
		\hspace*{-0.4em}
		\begin{subfigure}[b]{0.14\textwidth}
			\includegraphics[width=\textwidth]{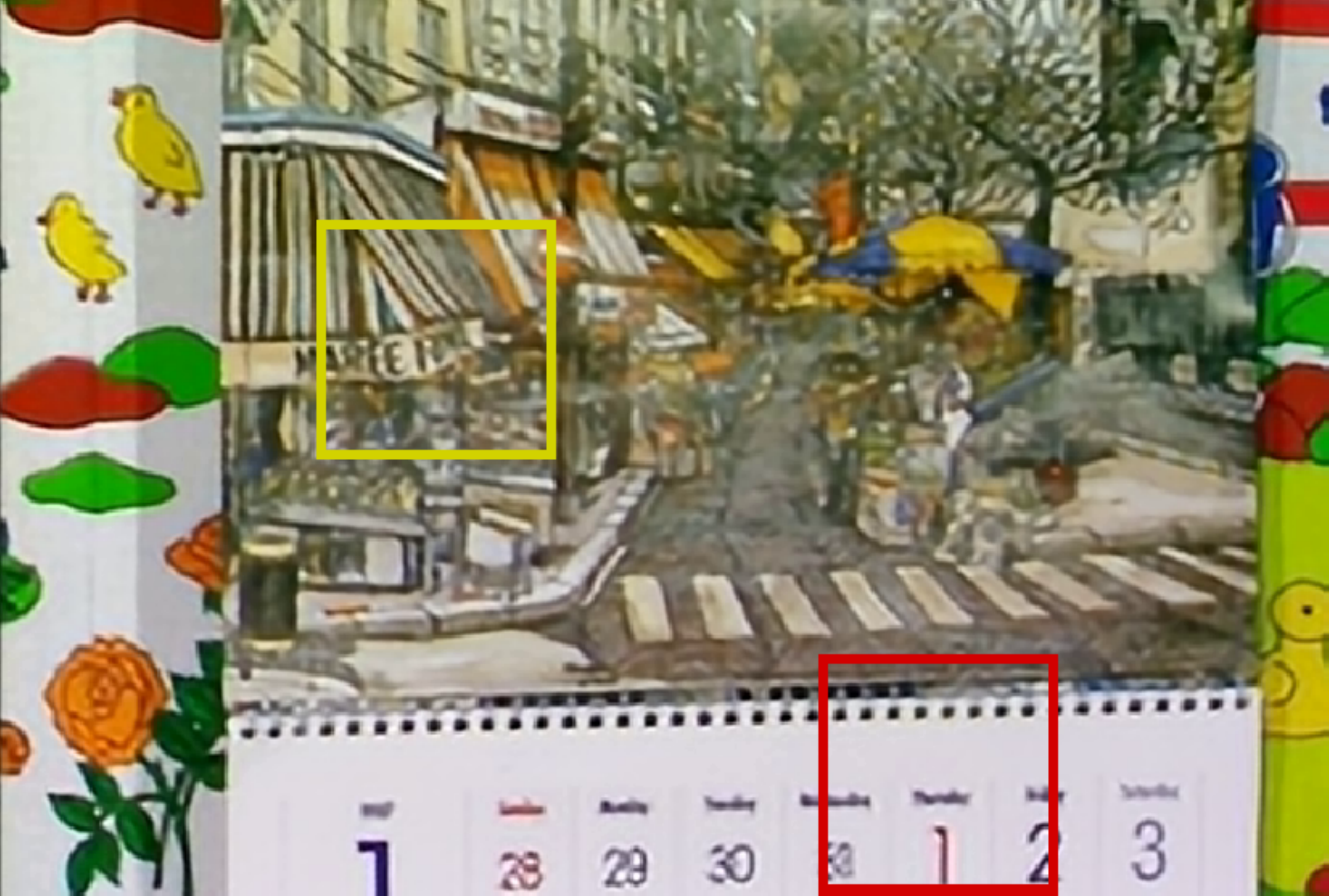}
			\label{fig:Calendar2}
		\end{subfigure}
		\hspace*{-0.4em}
		\begin{subfigure}[b]{0.14\textwidth}
			\includegraphics[width=\textwidth]{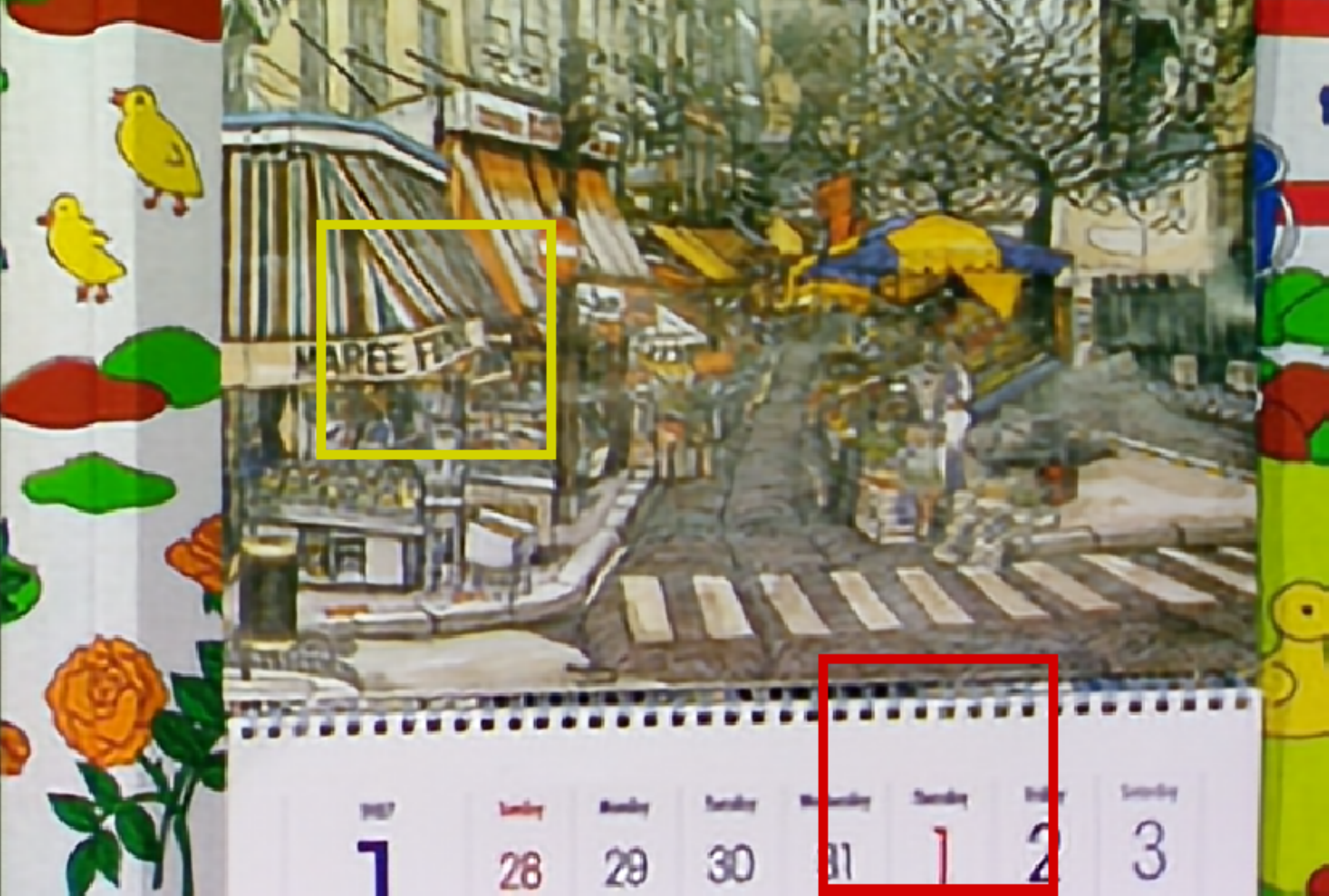}
			\label{fig:Calendar3}
		\end{subfigure}
		\hspace*{-0.4em}
		\begin{subfigure}[b]{0.14\textwidth}
			\includegraphics[width=\textwidth]{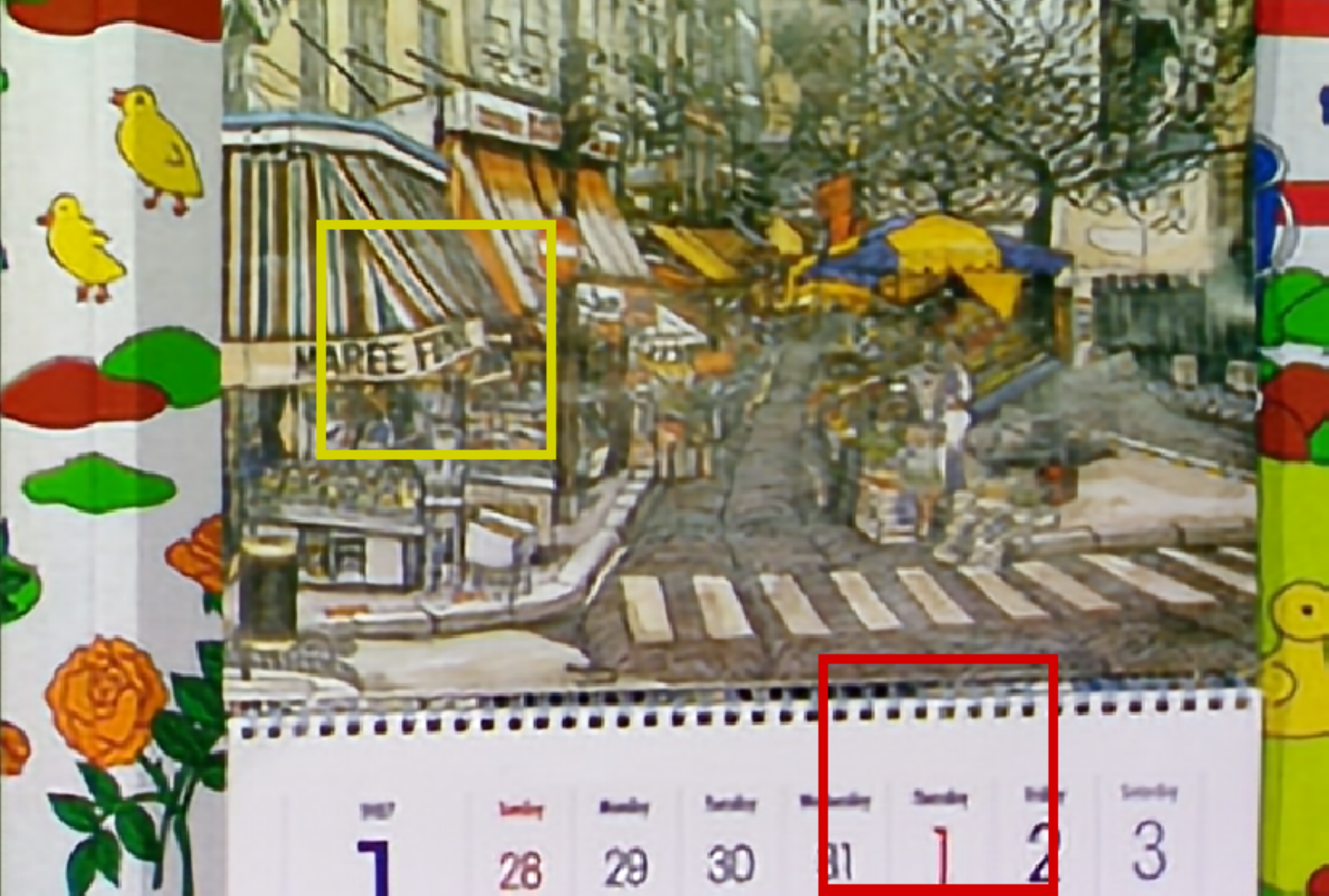}
			\label{fig:Calendar4}
		\end{subfigure}
		\hspace*{-0.4em}
		\begin{subfigure}[b]{0.14\textwidth}
			\includegraphics[width=\textwidth]{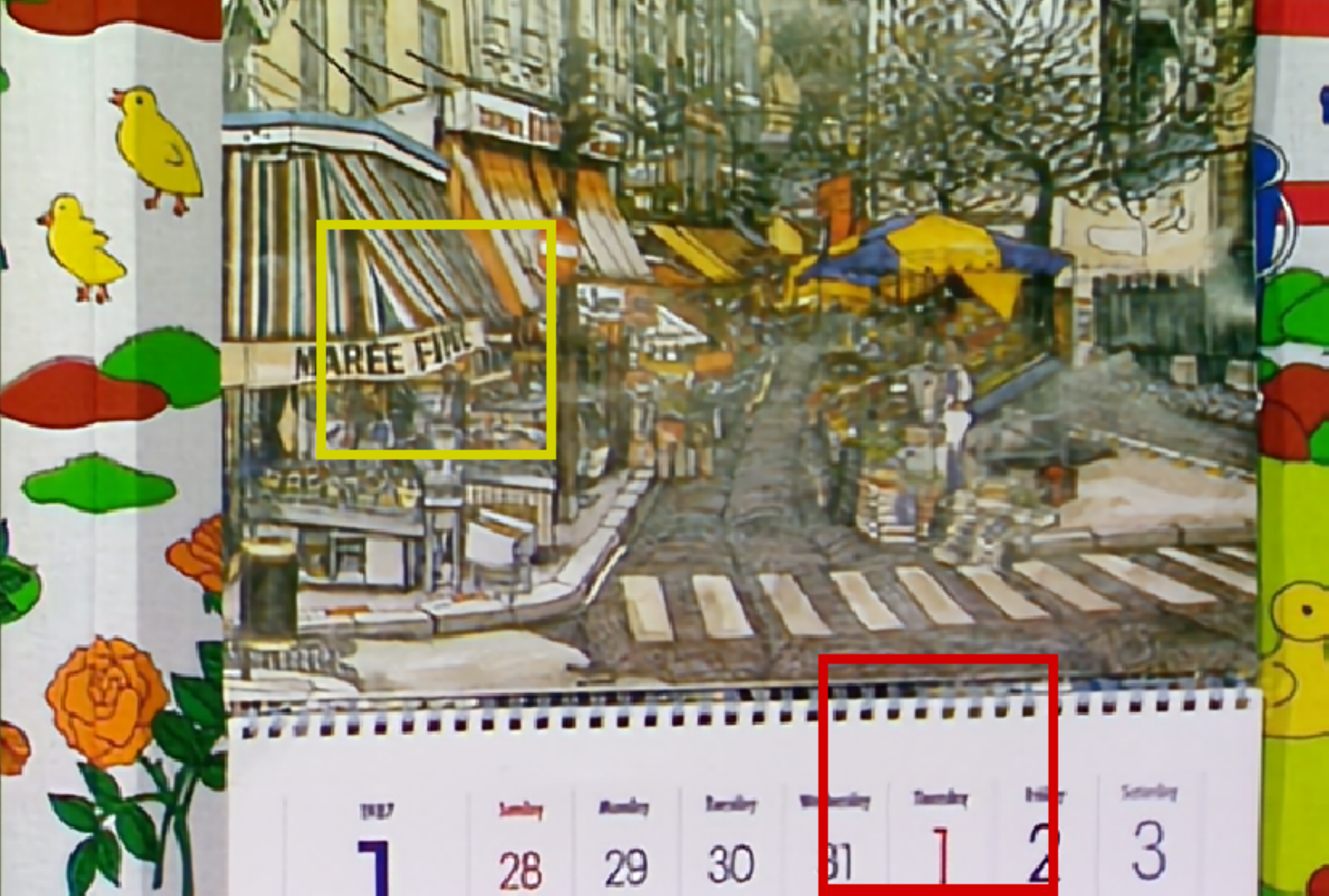}
			\label{fig:Calendar5}
		\end{subfigure}
		\hspace*{-0.4em}
		\begin{subfigure}[b]{0.14\textwidth}
			\includegraphics[width=\textwidth]{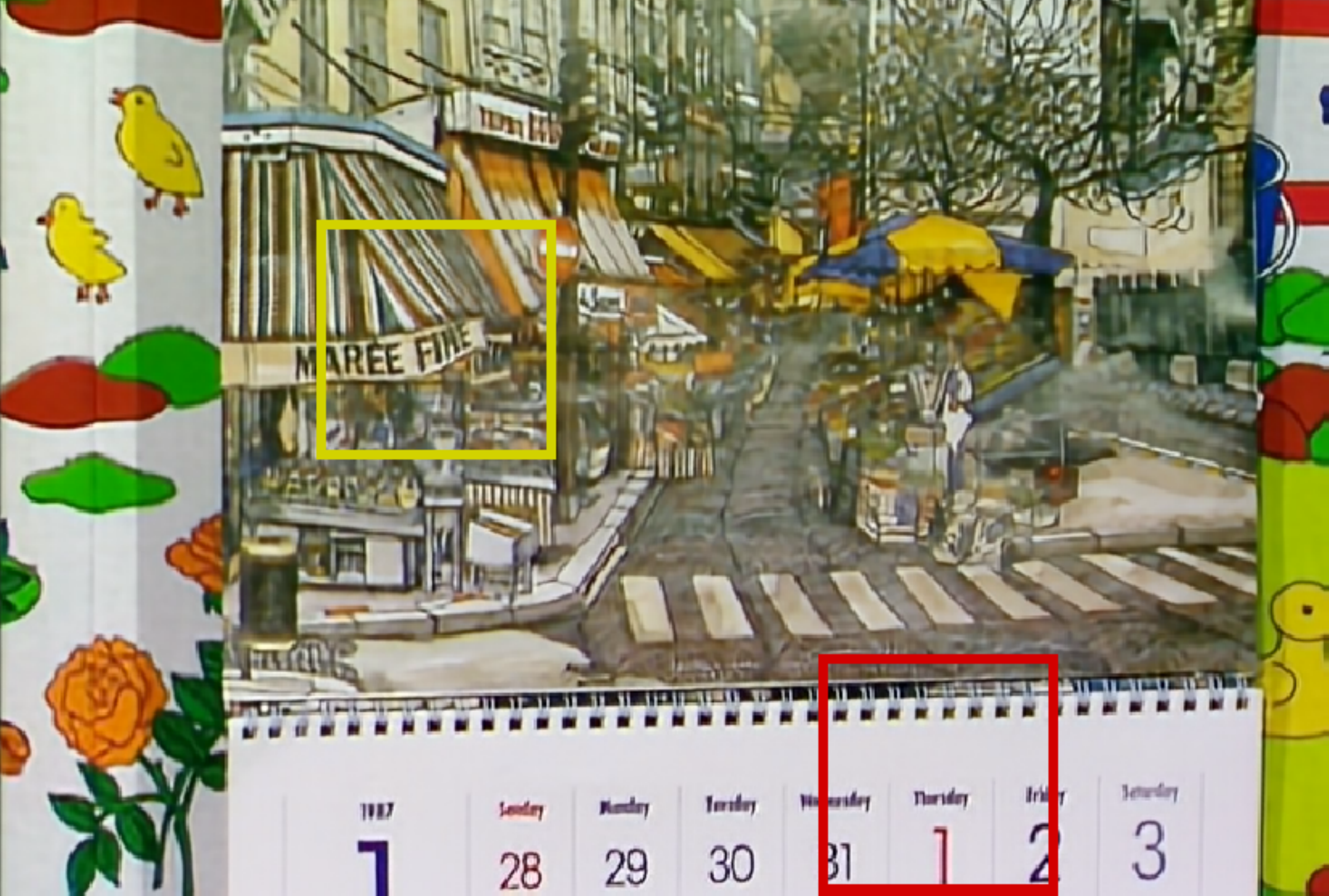}
			\label{fig:Calendar6}
		\end{subfigure}
		\hspace*{-0.4em}
		\begin{subfigure}[b]{0.14\textwidth}
			\includegraphics[width=\textwidth]{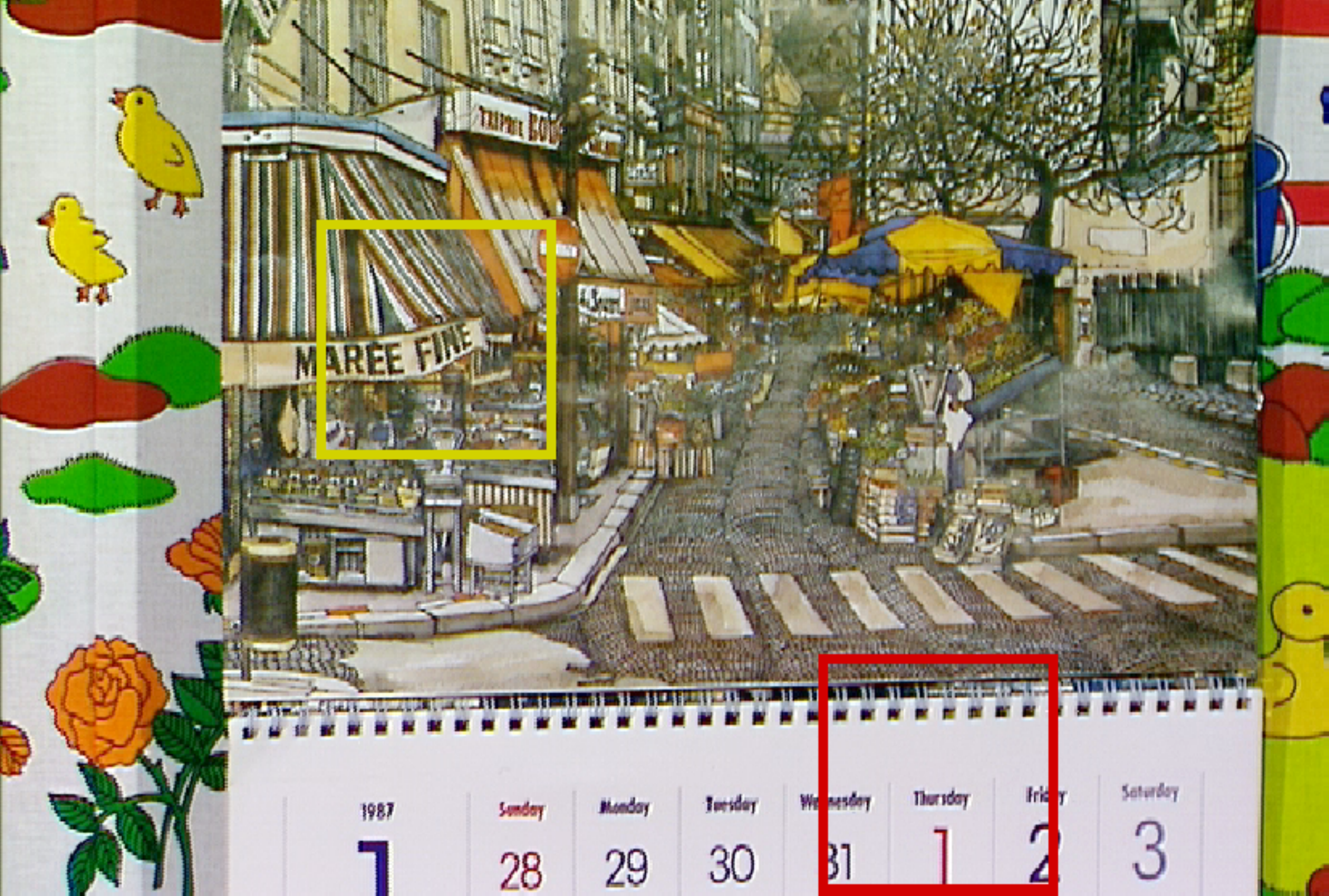}
			\label{fig:Calendar7}
		\end{subfigure}
		\\
		\vspace*{-1.2em}	
		\begin{subfigure}[b]{0.14\textwidth}
			\includegraphics[width=\textwidth]{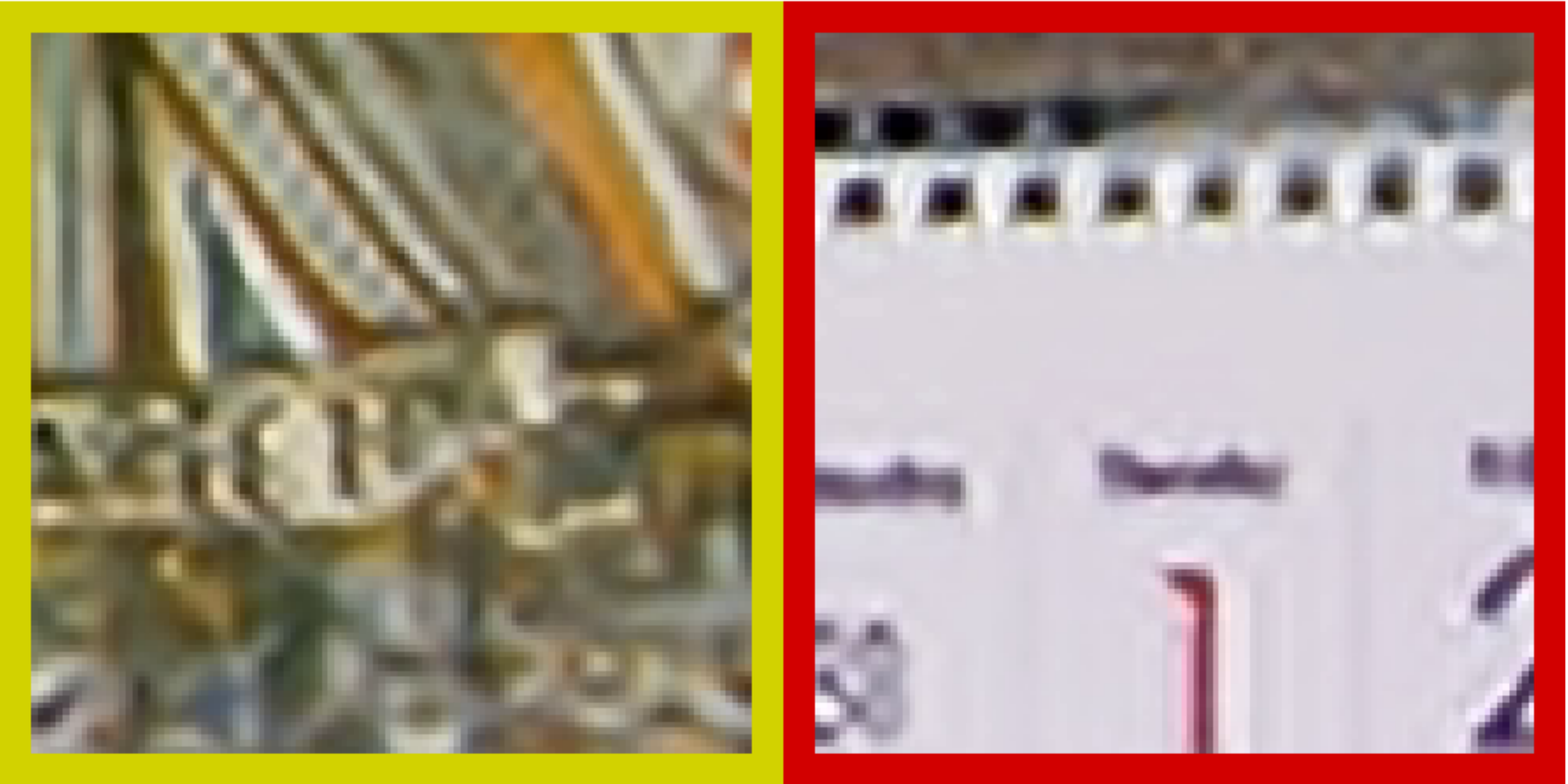}
			\label{fig:Calendar_part1}
		\end{subfigure}
		\hspace*{-0.4em}
		\begin{subfigure}[b]{0.14\textwidth}
			\includegraphics[width=\textwidth]{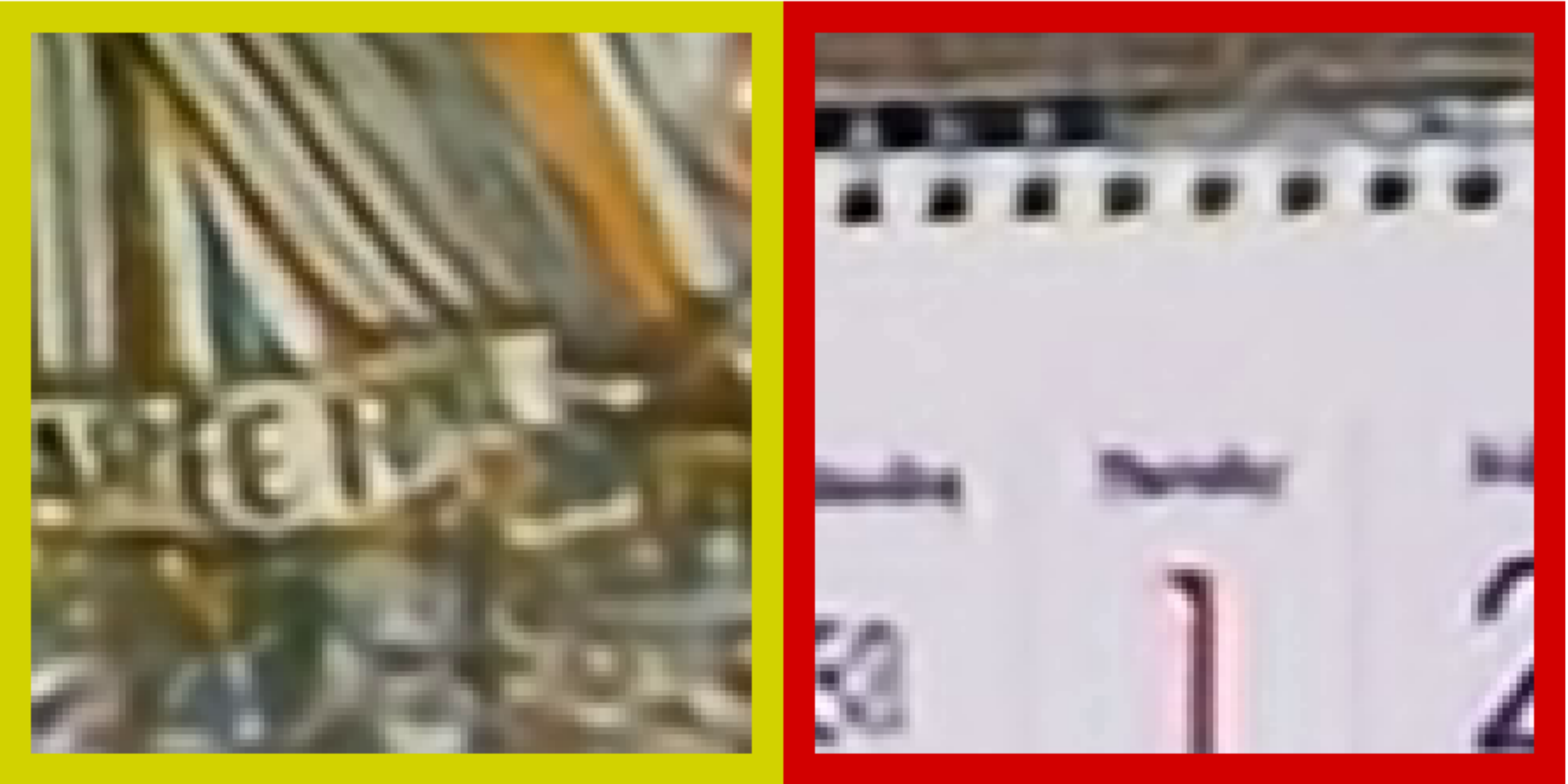}
			\label{fig:Calendar_part2}
		\end{subfigure}
		\hspace*{-0.4em}
		\begin{subfigure}[b]{0.14\textwidth}
			\includegraphics[width=\textwidth]{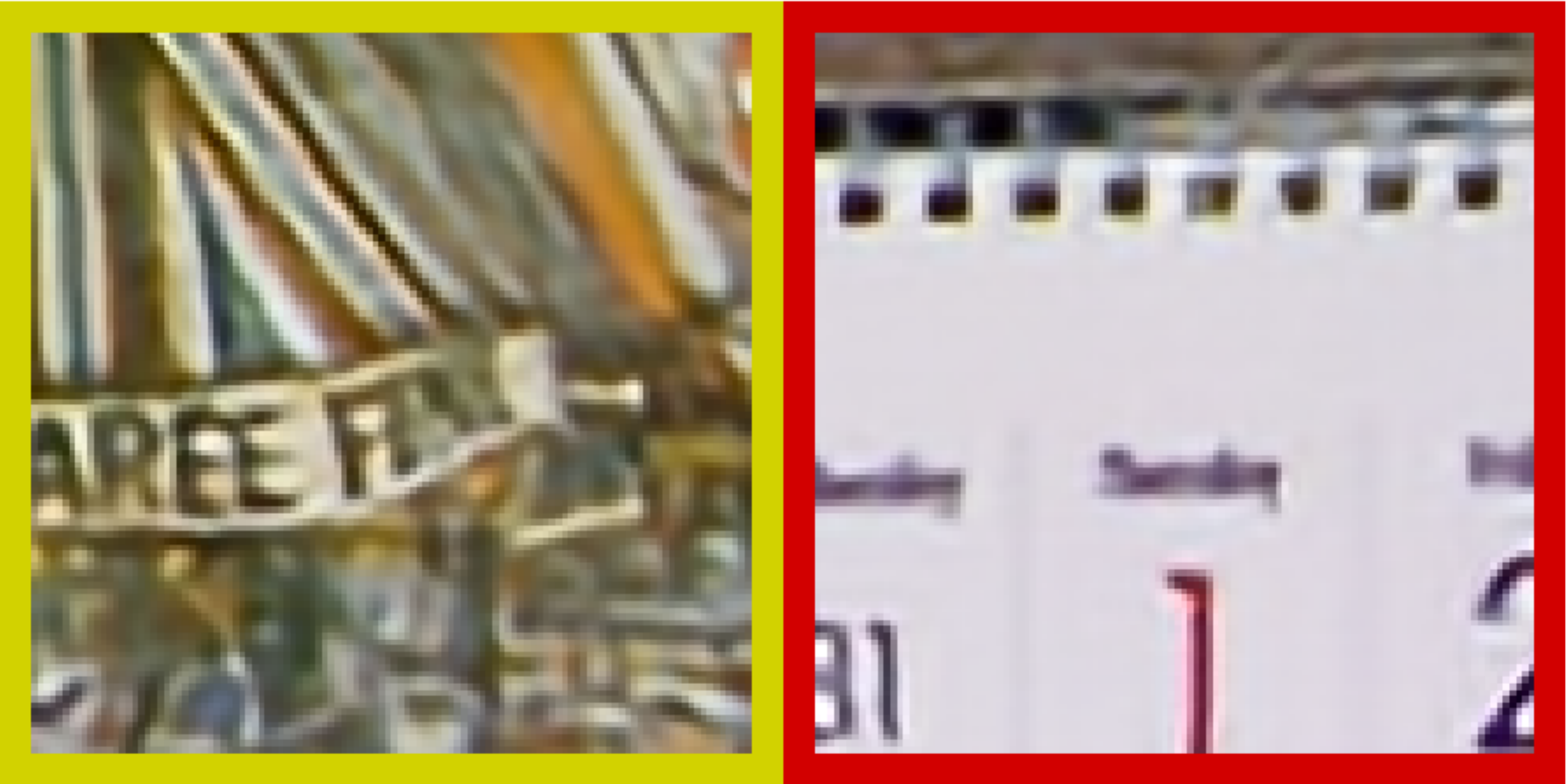}
			\label{fig:Calendar_part3}
		\end{subfigure}
		\hspace*{-0.4em}
		\begin{subfigure}[b]{0.14\textwidth}
			\includegraphics[width=\textwidth]{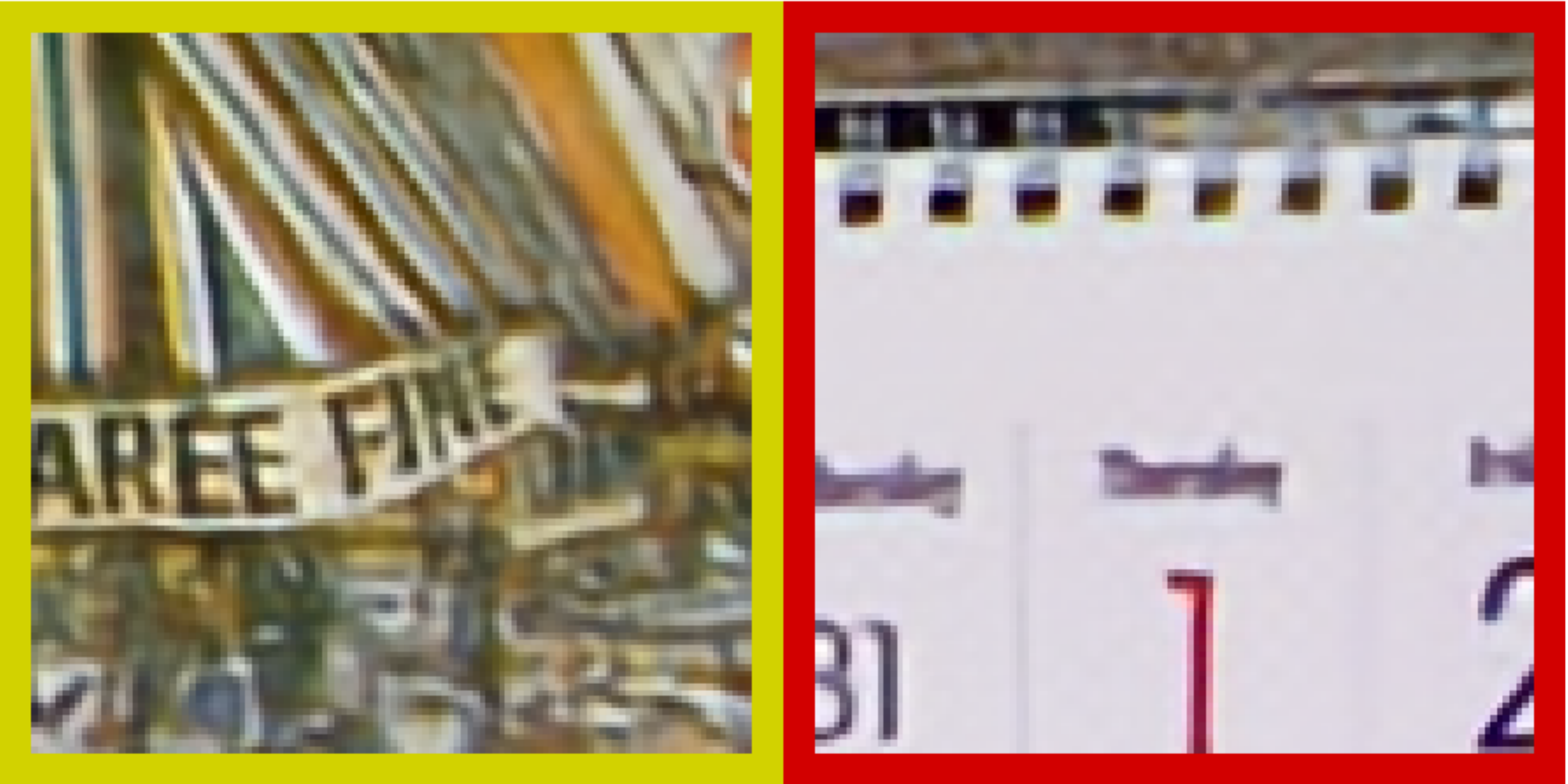}
			\label{fig:Calendar_part4}
		\end{subfigure}
		\hspace*{-0.4em}
		\begin{subfigure}[b]{0.14\textwidth}
			\includegraphics[width=\textwidth]{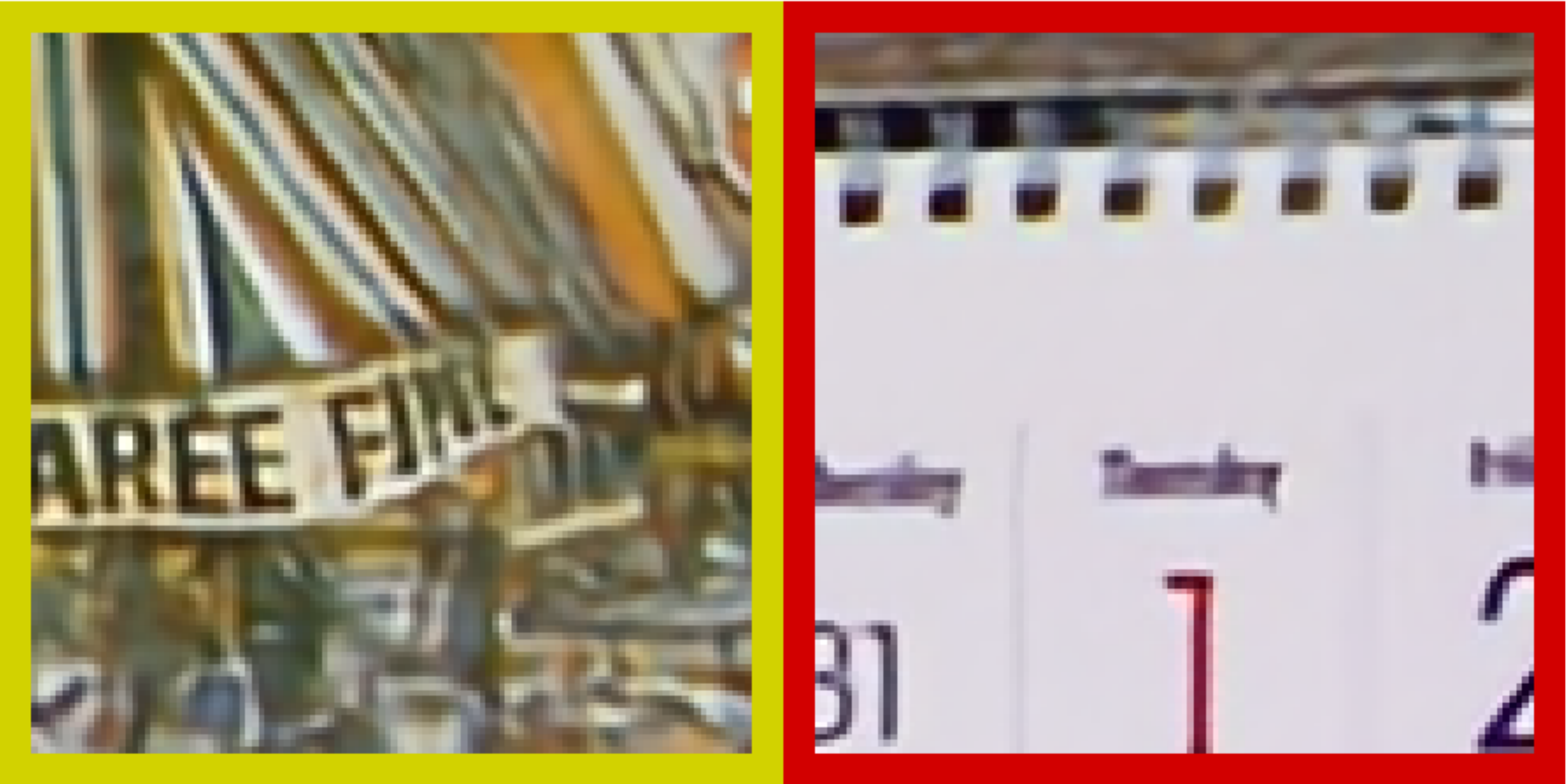}
			\label{fig:Calendar_part5}
		\end{subfigure}
		\hspace*{-0.4em}
		\begin{subfigure}[b]{0.14\textwidth}
			\includegraphics[width=\textwidth]{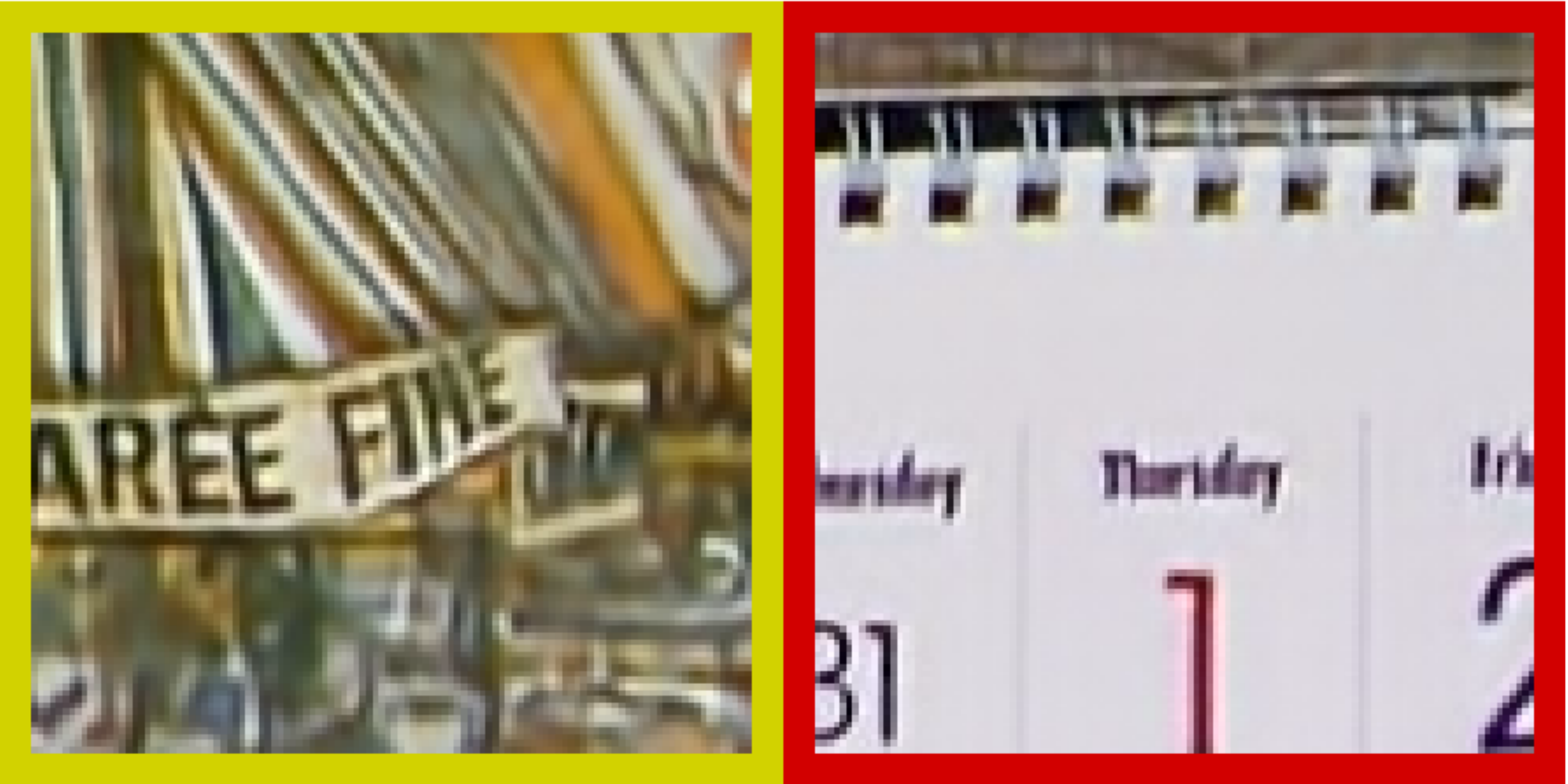}
			\label{fig:Calendar_part6}
		\end{subfigure}
		\hspace*{-0.4em}
		\begin{subfigure}[b]{0.14\textwidth}
			\includegraphics[width=\textwidth]{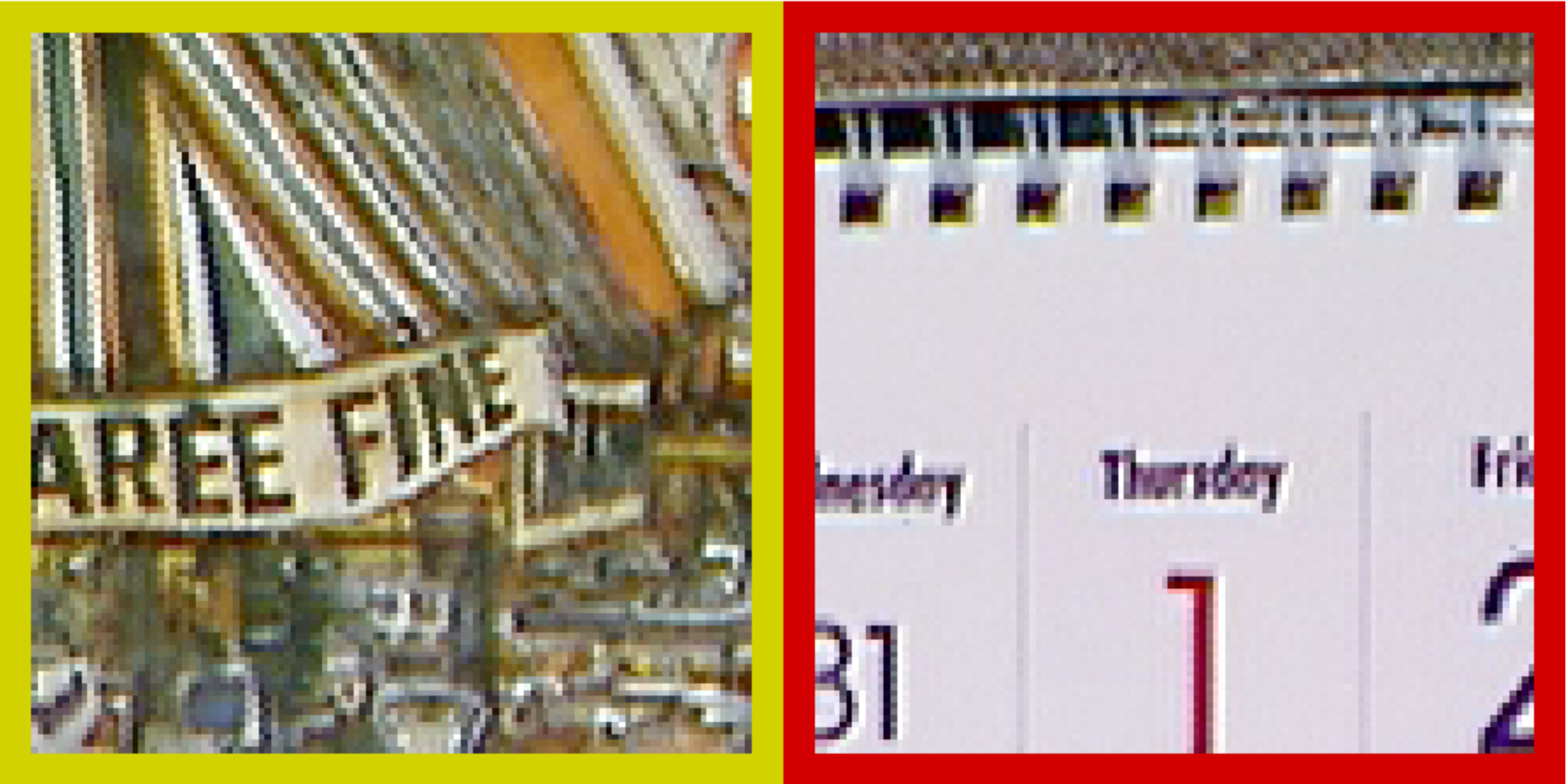}
			\label{fig:Calendar_part7}
		\end{subfigure}
		\\
		\vspace*{-1em}	
		\begin{subfigure}[b]{0.14\textwidth}
			\includegraphics[width=\textwidth]{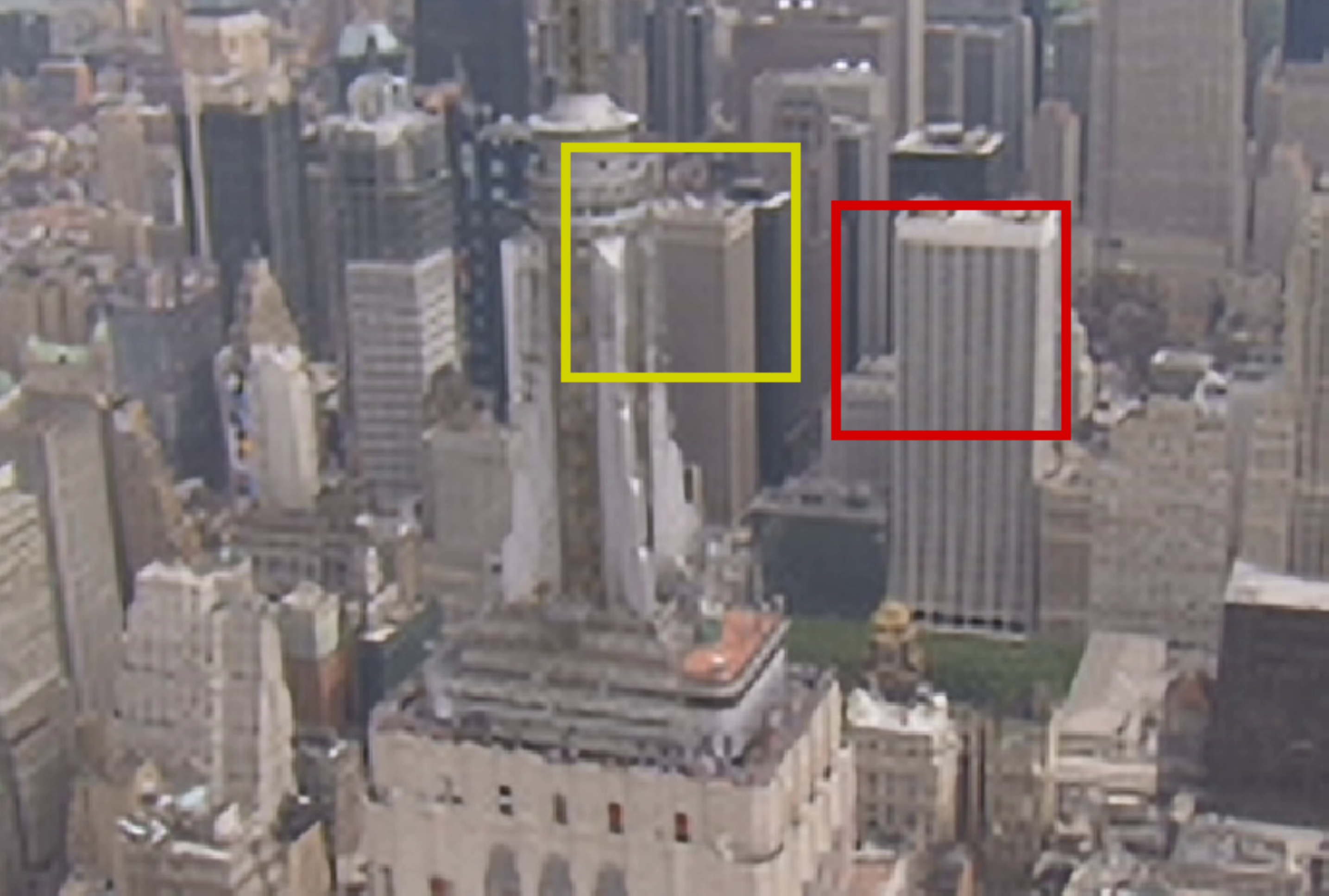}
			\label{fig:City1}
		\end{subfigure}
		\hspace*{-0.4em}
		\begin{subfigure}[b]{0.14\textwidth}
			\includegraphics[width=\textwidth]{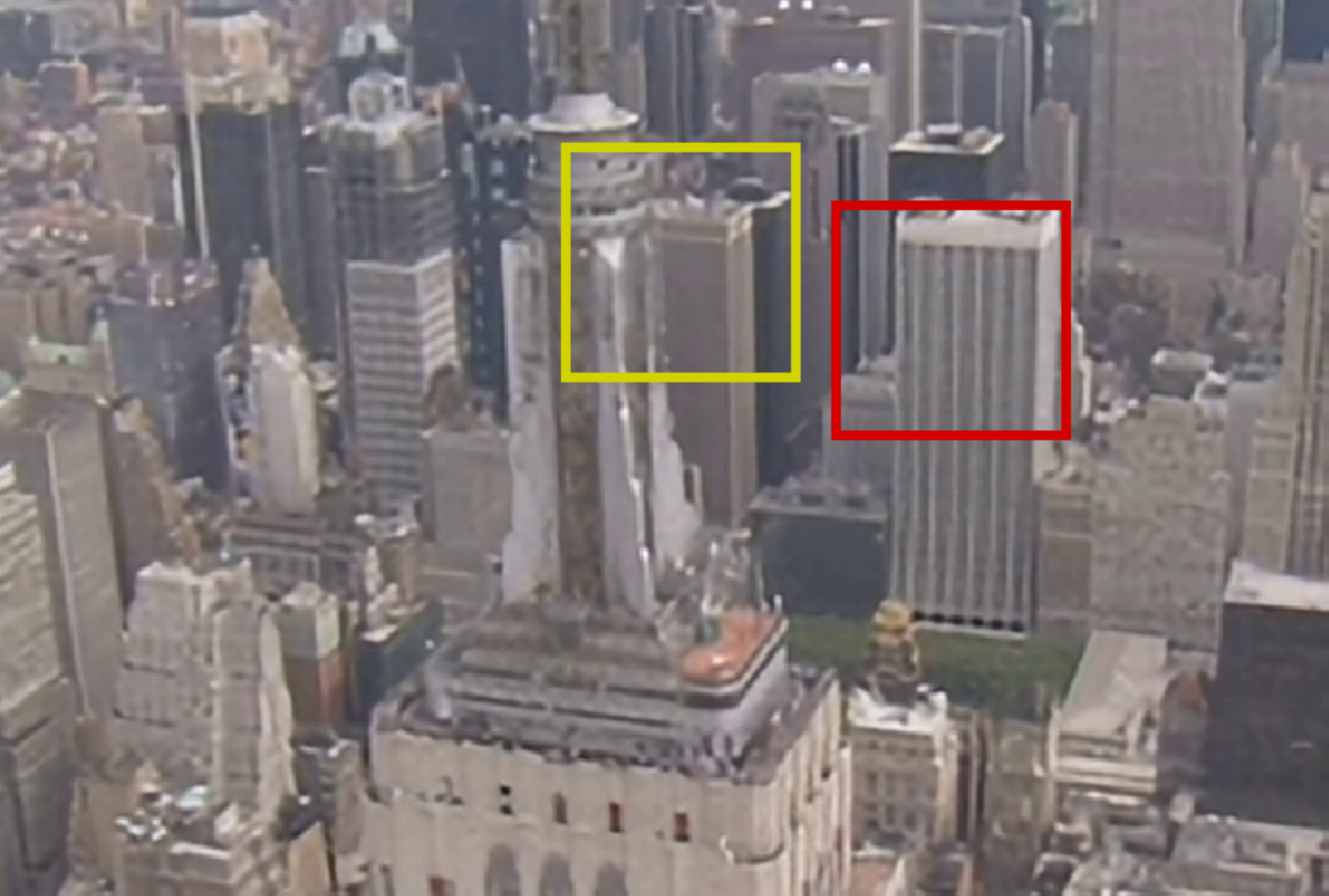}
			\label{fig:City2}
		\end{subfigure}
		\hspace*{-0.4em}
		\begin{subfigure}[b]{0.14\textwidth}
			\includegraphics[width=\textwidth]{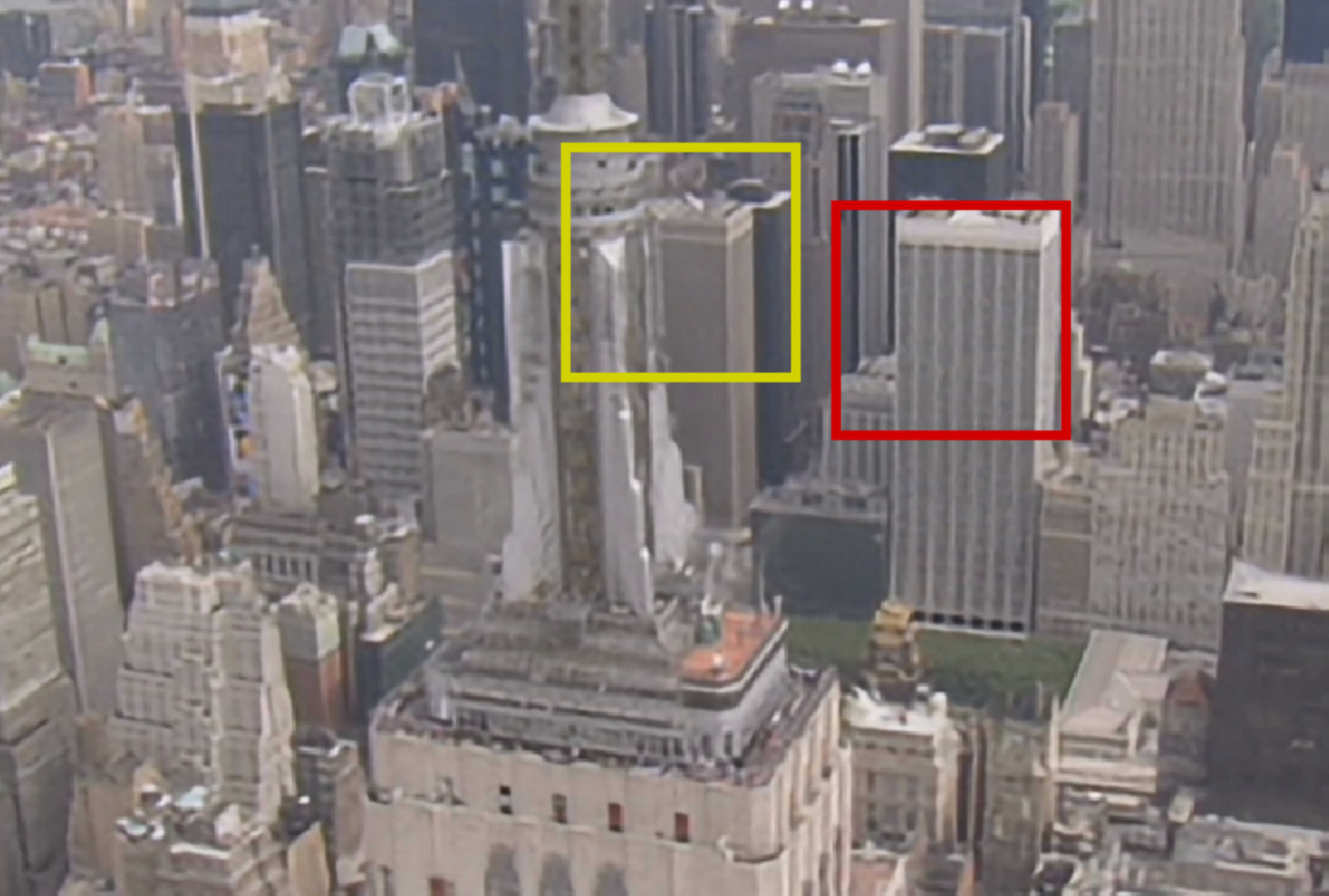}
			\label{fig:City3}
		\end{subfigure}
		\hspace*{-0.4em}
		\begin{subfigure}[b]{0.14\textwidth}
			\includegraphics[width=\textwidth]{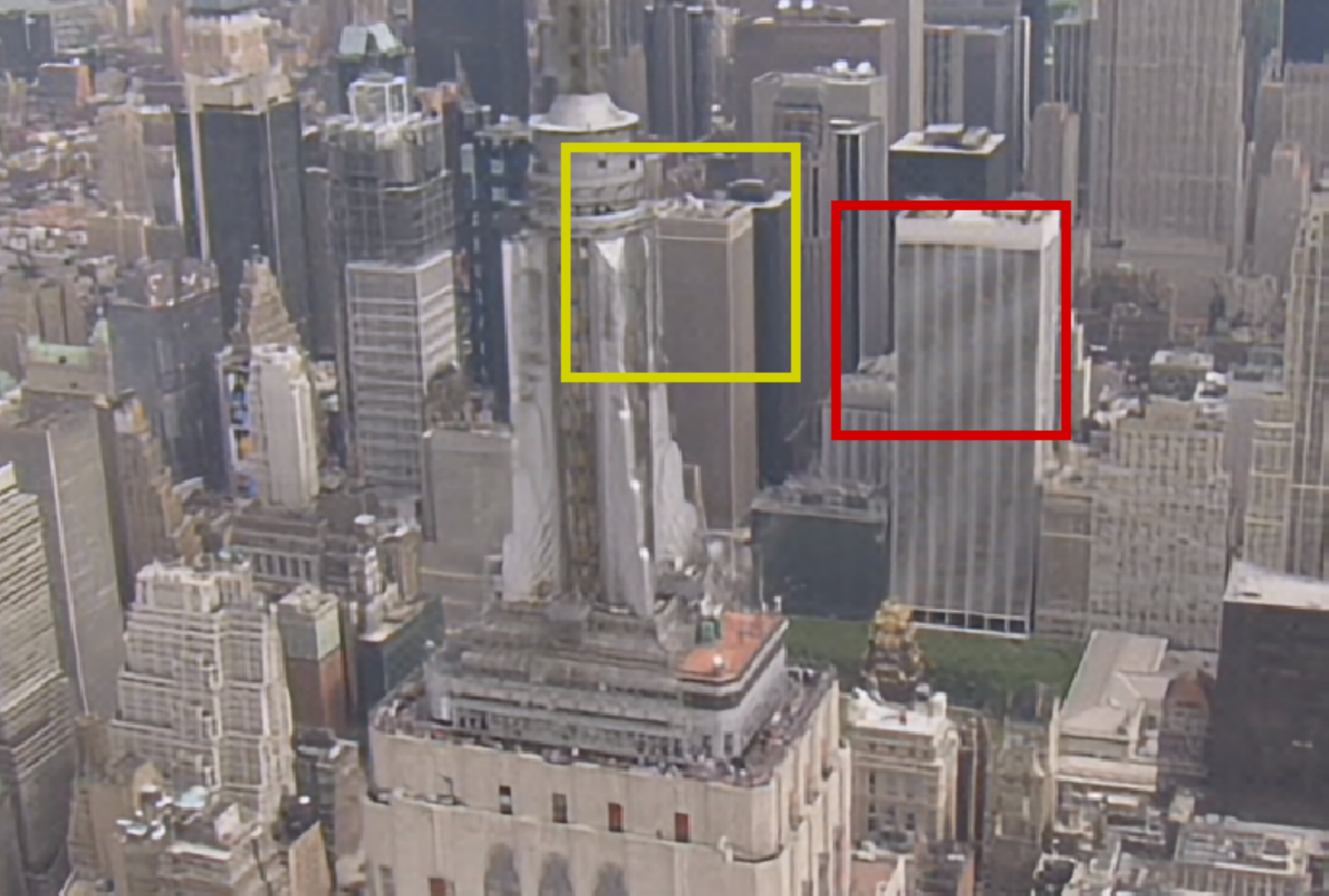}
			\label{fig:City4}
		\end{subfigure}
		\hspace*{-0.4em}
		\begin{subfigure}[b]{0.14\textwidth}
			\includegraphics[width=\textwidth]{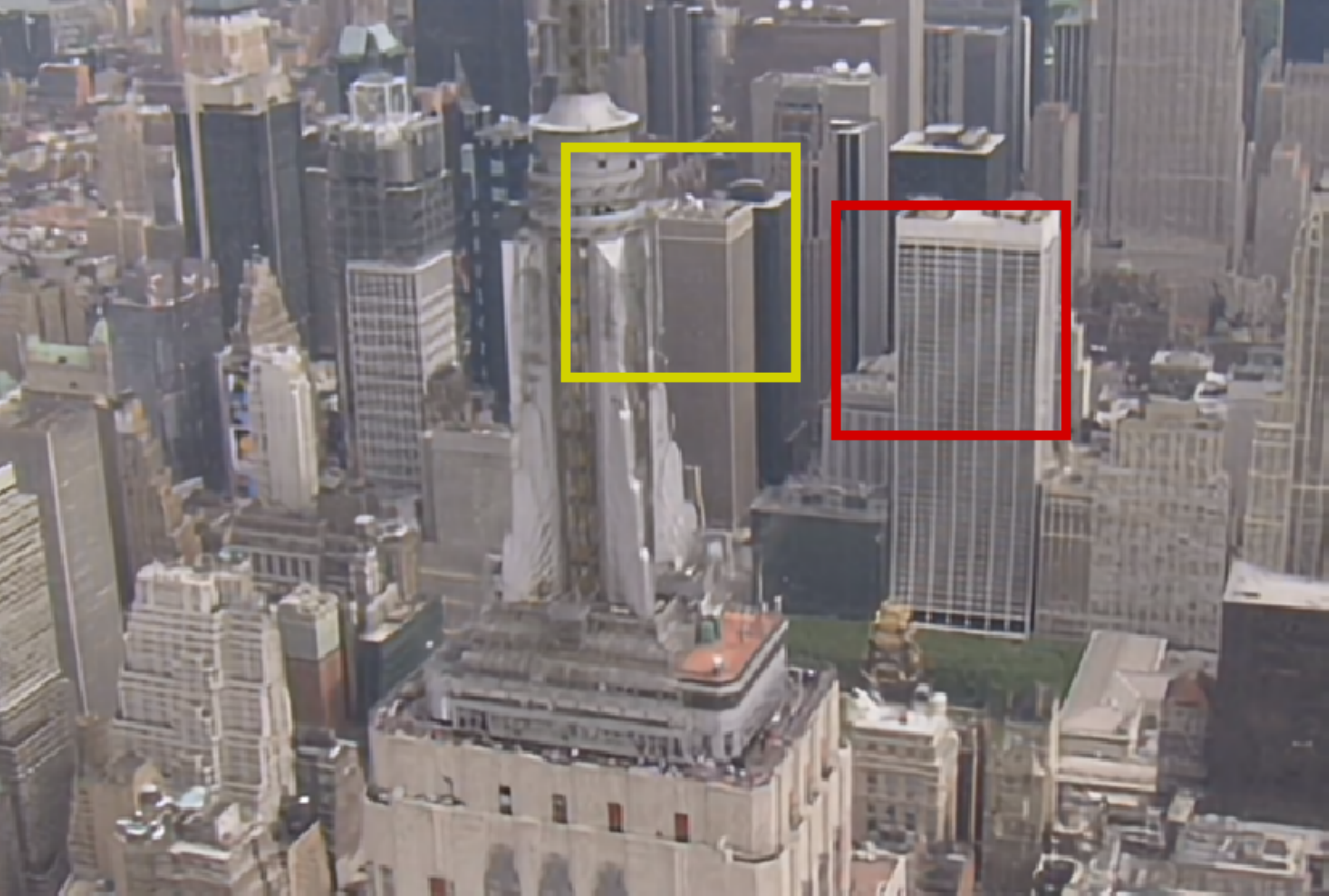}
			\label{fig:City5}
		\end{subfigure}
		\hspace*{-0.4em}
		\begin{subfigure}[b]{0.14\textwidth}
			\includegraphics[width=\textwidth]{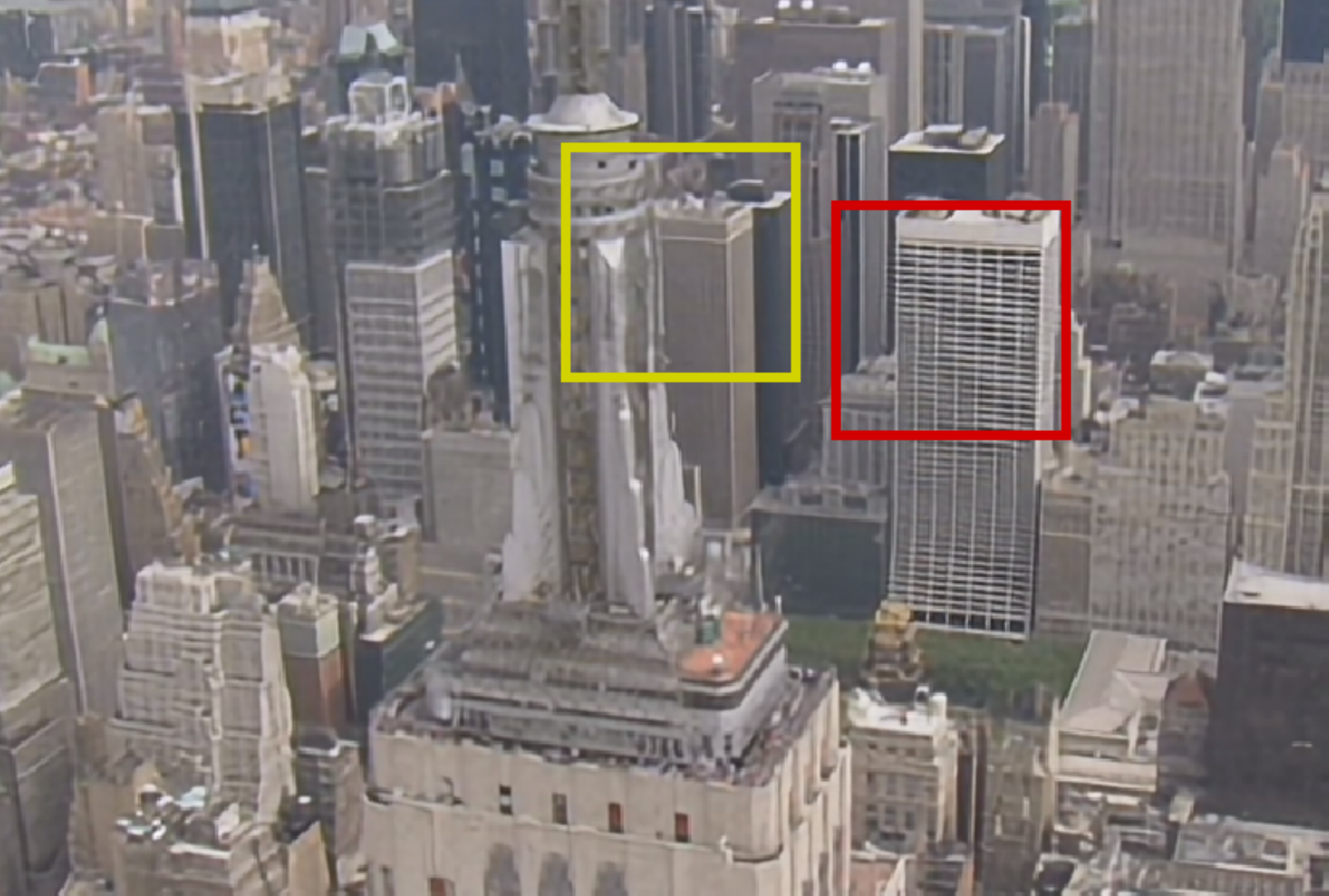}
			\label{fig:City6}
		\end{subfigure}
		\hspace*{-0.4em}
		\begin{subfigure}[b]{0.14\textwidth}
			\includegraphics[width=\textwidth]{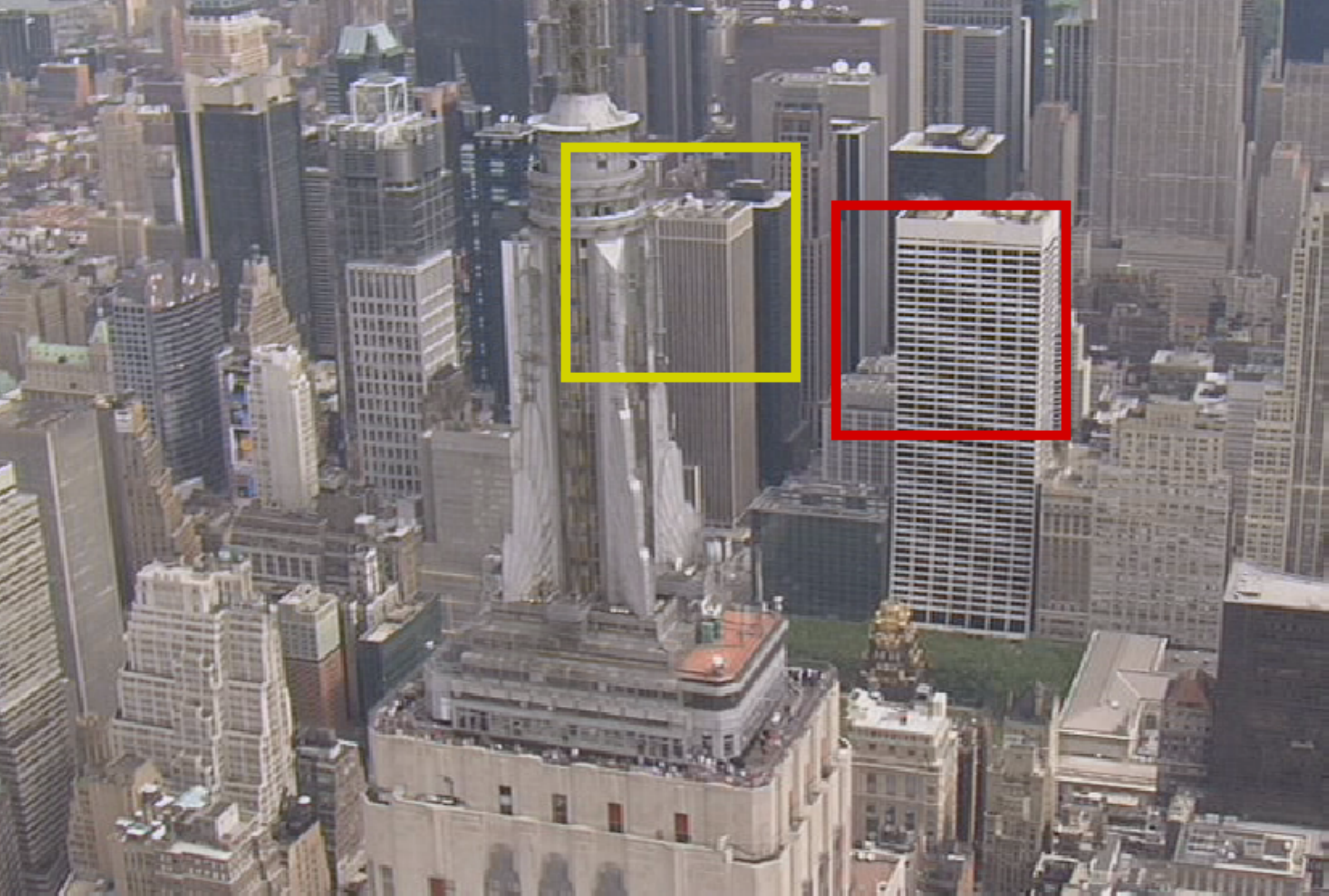}
			\label{fig:City7}
		\end{subfigure}
		\\
		\vspace*{-1.2em}	
		\begin{subfigure}[b]{0.14\textwidth}
			\includegraphics[width=\textwidth]{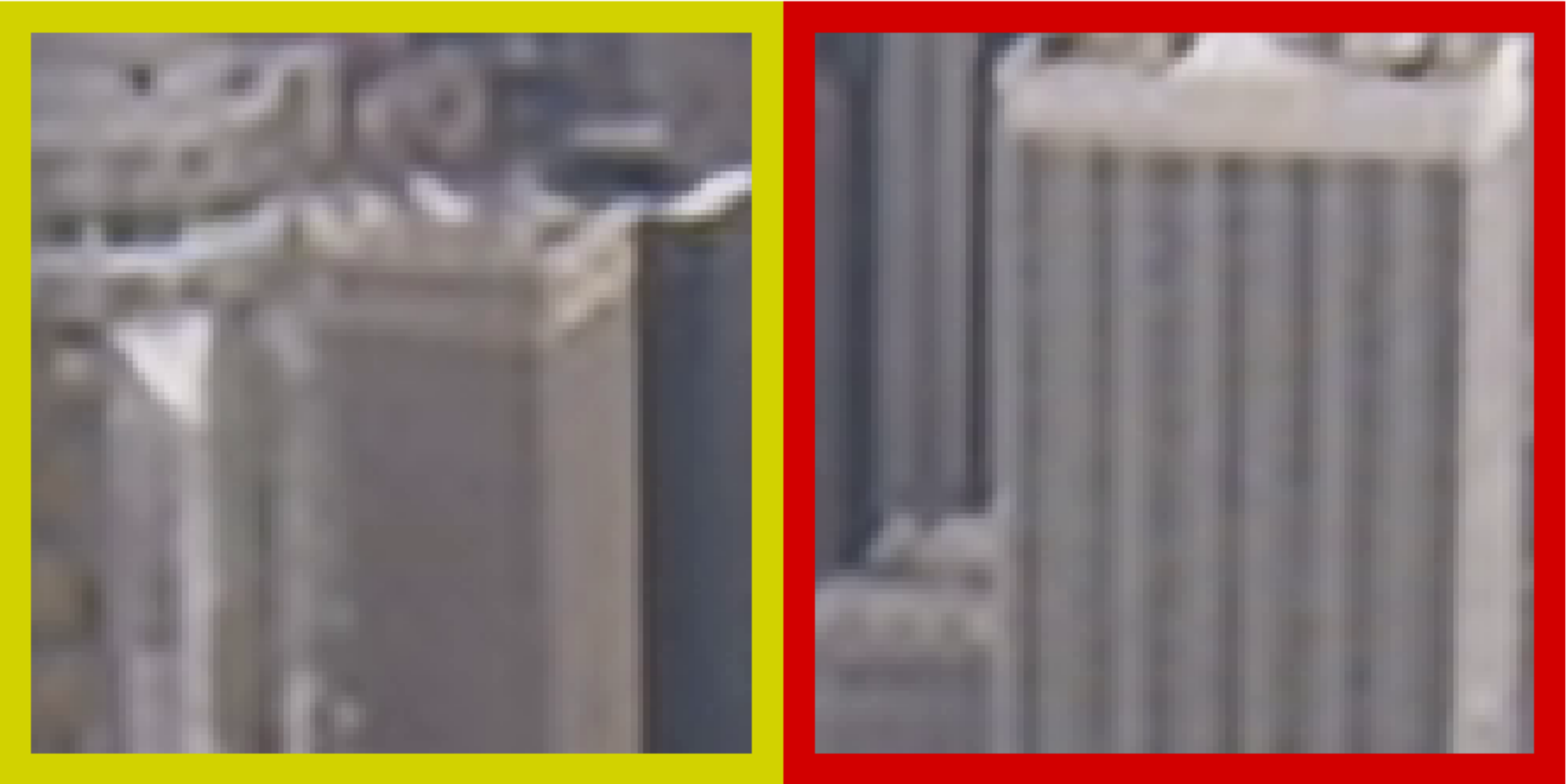}
			\label{fig:City_part1}
		\end{subfigure}
		\hspace*{-0.4em}
		\begin{subfigure}[b]{0.14\textwidth}
			\includegraphics[width=\textwidth]{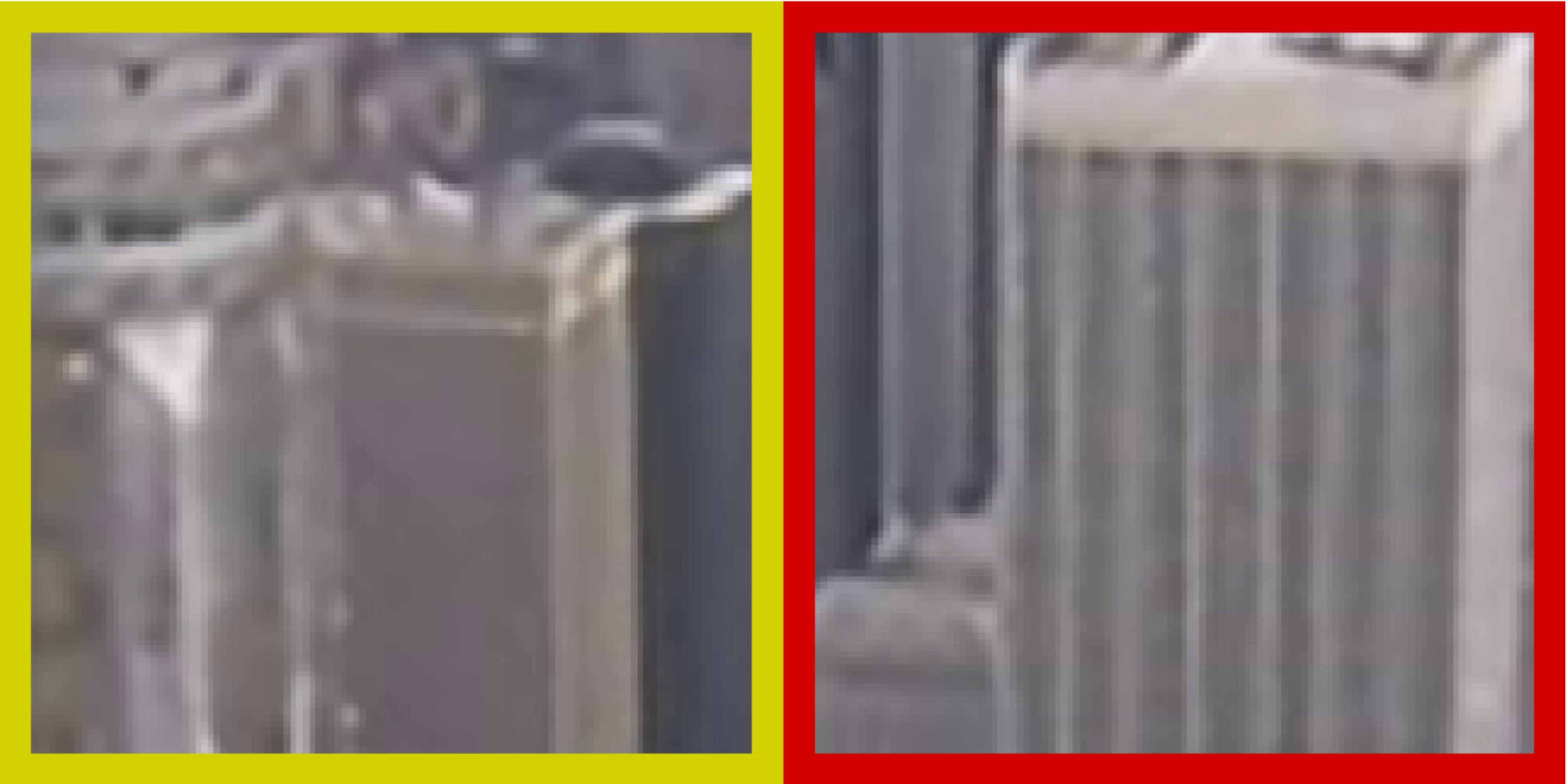}
			\label{fig:City_part2}
		\end{subfigure}
		\hspace*{-0.4em}
		\begin{subfigure}[b]{0.14\textwidth}
			\includegraphics[width=\textwidth]{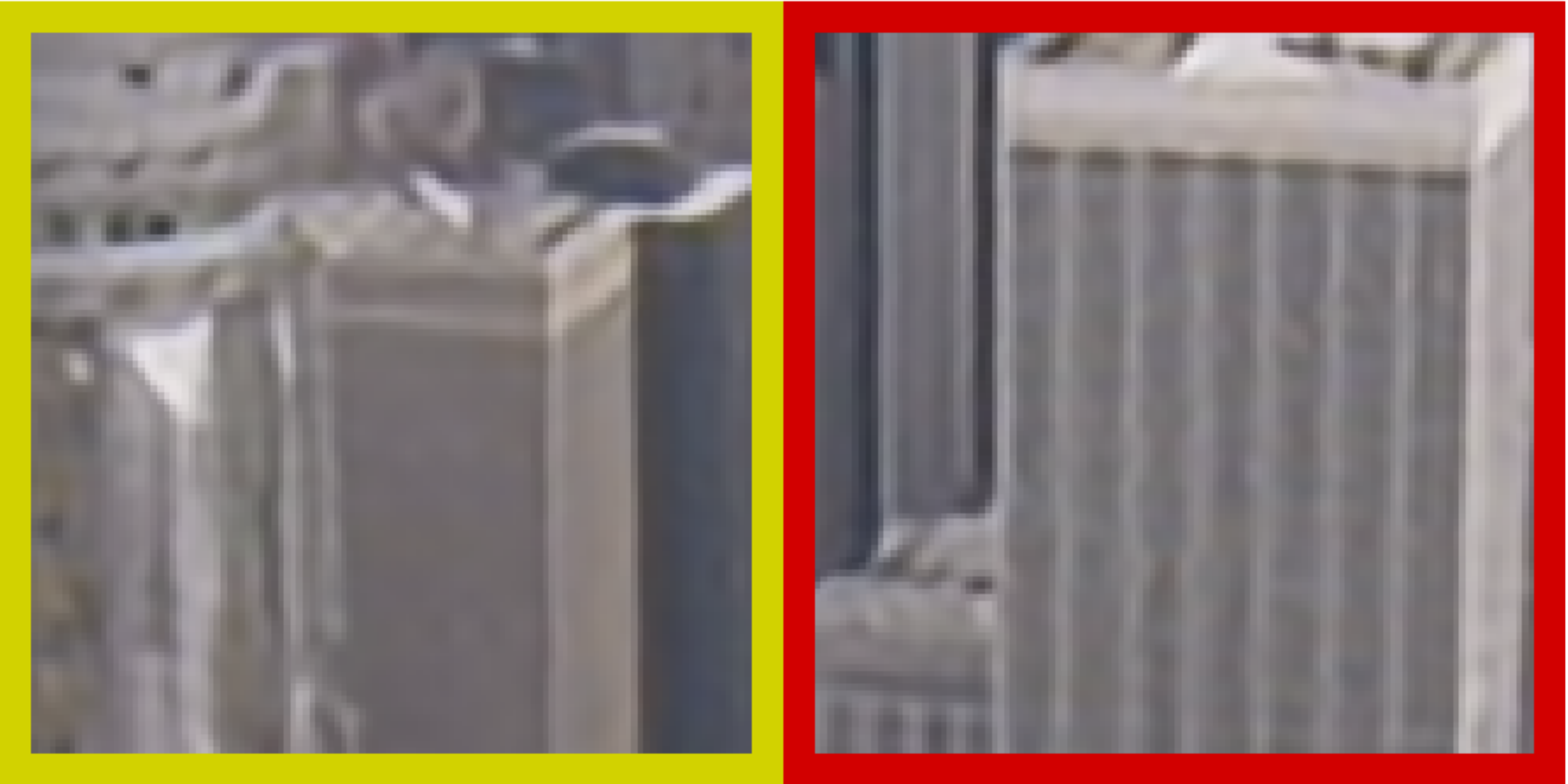}
			\label{fig:City_part3}
		\end{subfigure}
		\hspace*{-0.4em}
		\begin{subfigure}[b]{0.14\textwidth}
			\includegraphics[width=\textwidth]{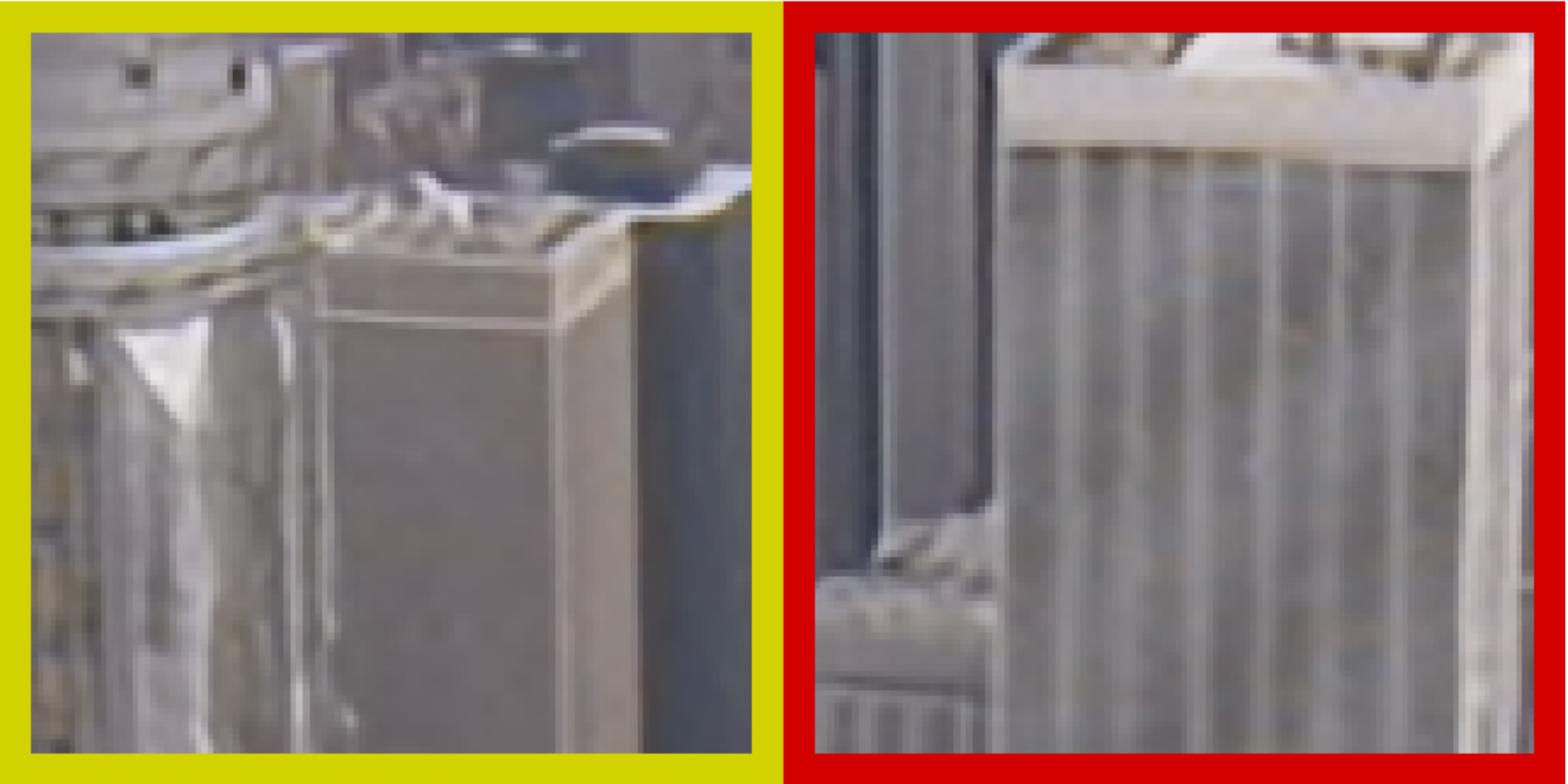}
			\label{fig:City_part4}
		\end{subfigure}
		\hspace*{-0.4em}
		\begin{subfigure}[b]{0.14\textwidth}
			\includegraphics[width=\textwidth]{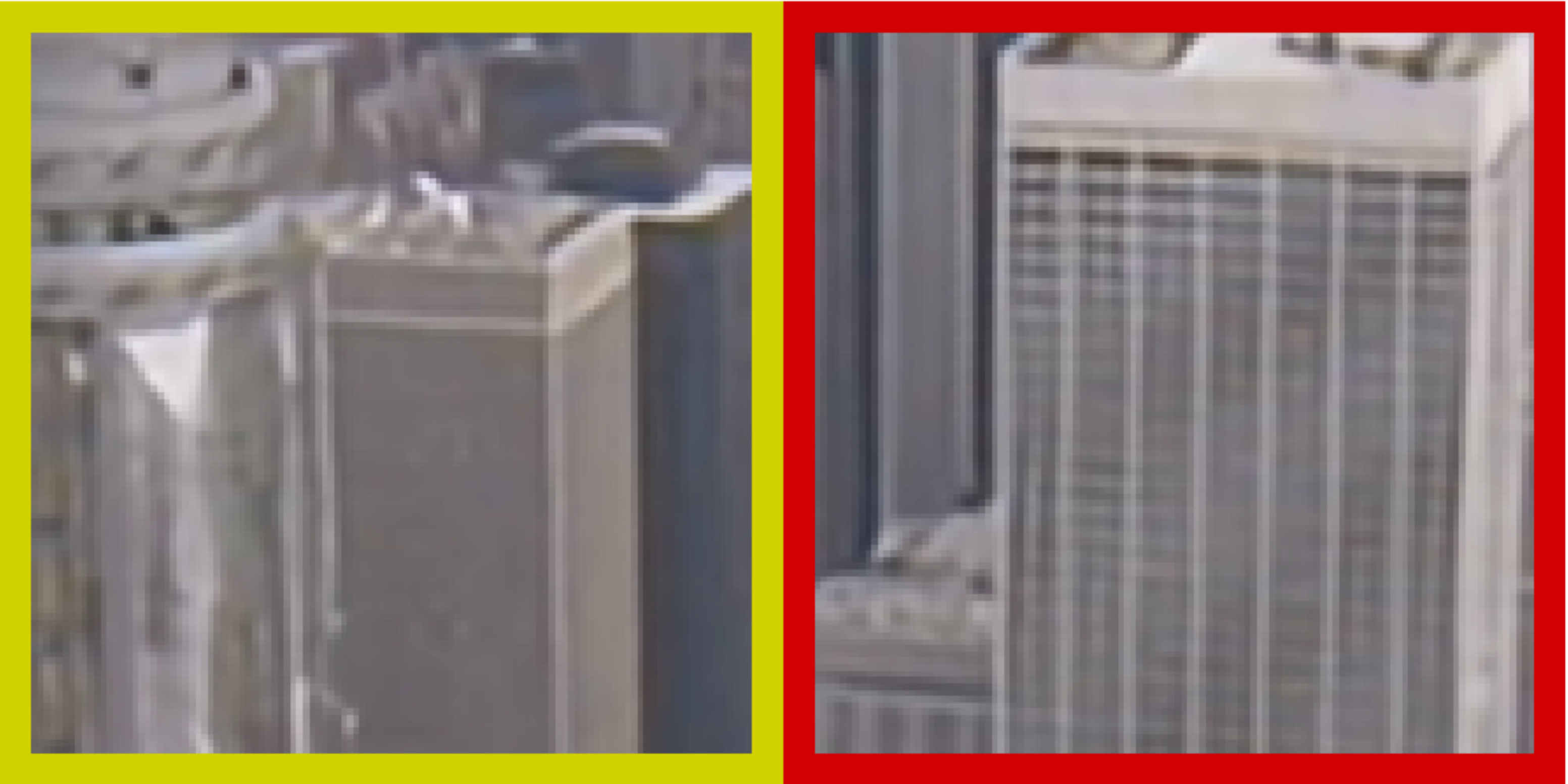}
			\label{fig:City_part5}
		\end{subfigure}
		\hspace*{-0.4em}
		\begin{subfigure}[b]{0.14\textwidth}
			\includegraphics[width=\textwidth]{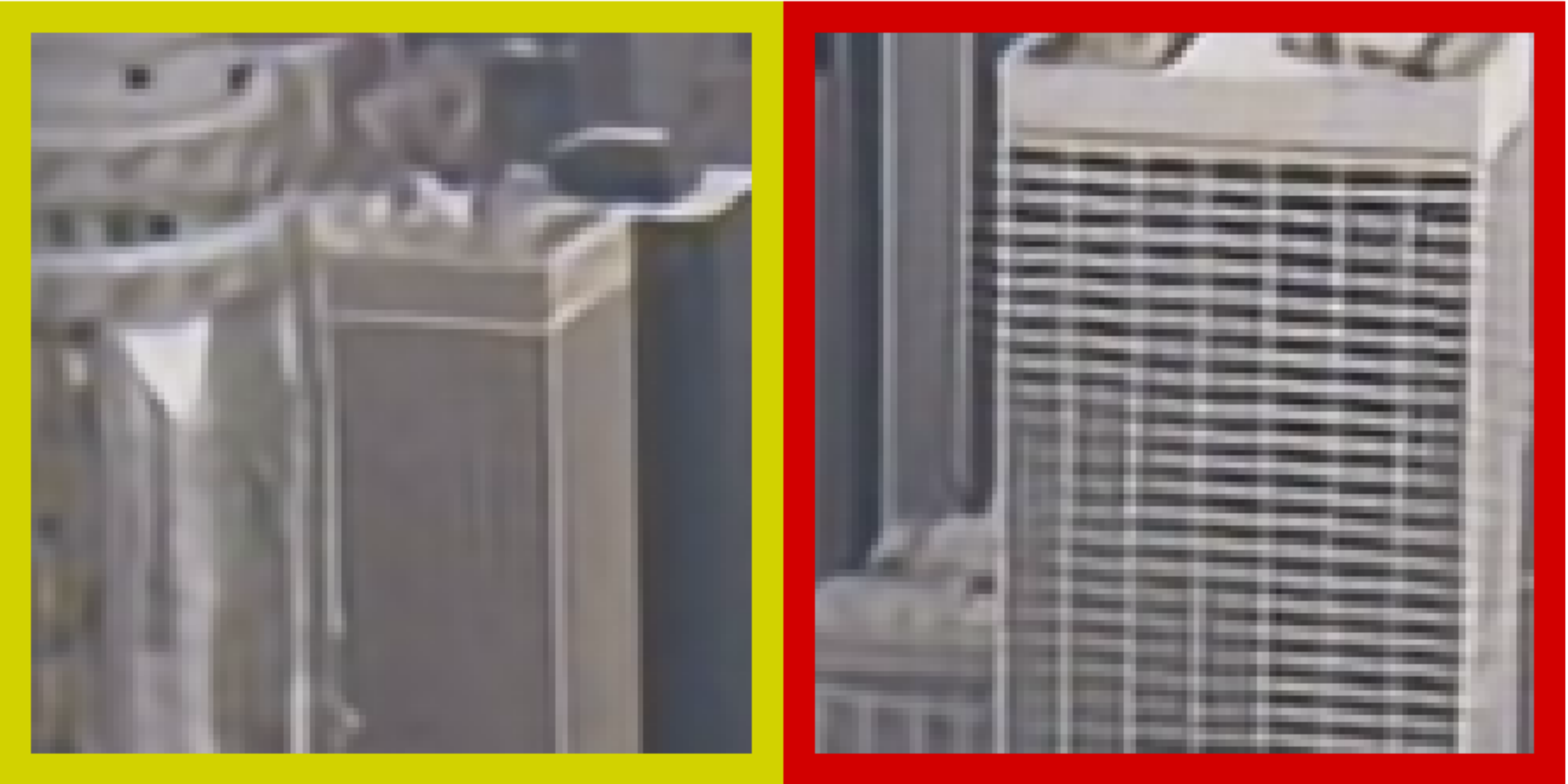}
			\label{fig:City_part6}
		\end{subfigure}
		\hspace*{-0.4em}
		\begin{subfigure}[b]{0.14\textwidth}
			\includegraphics[width=\textwidth]{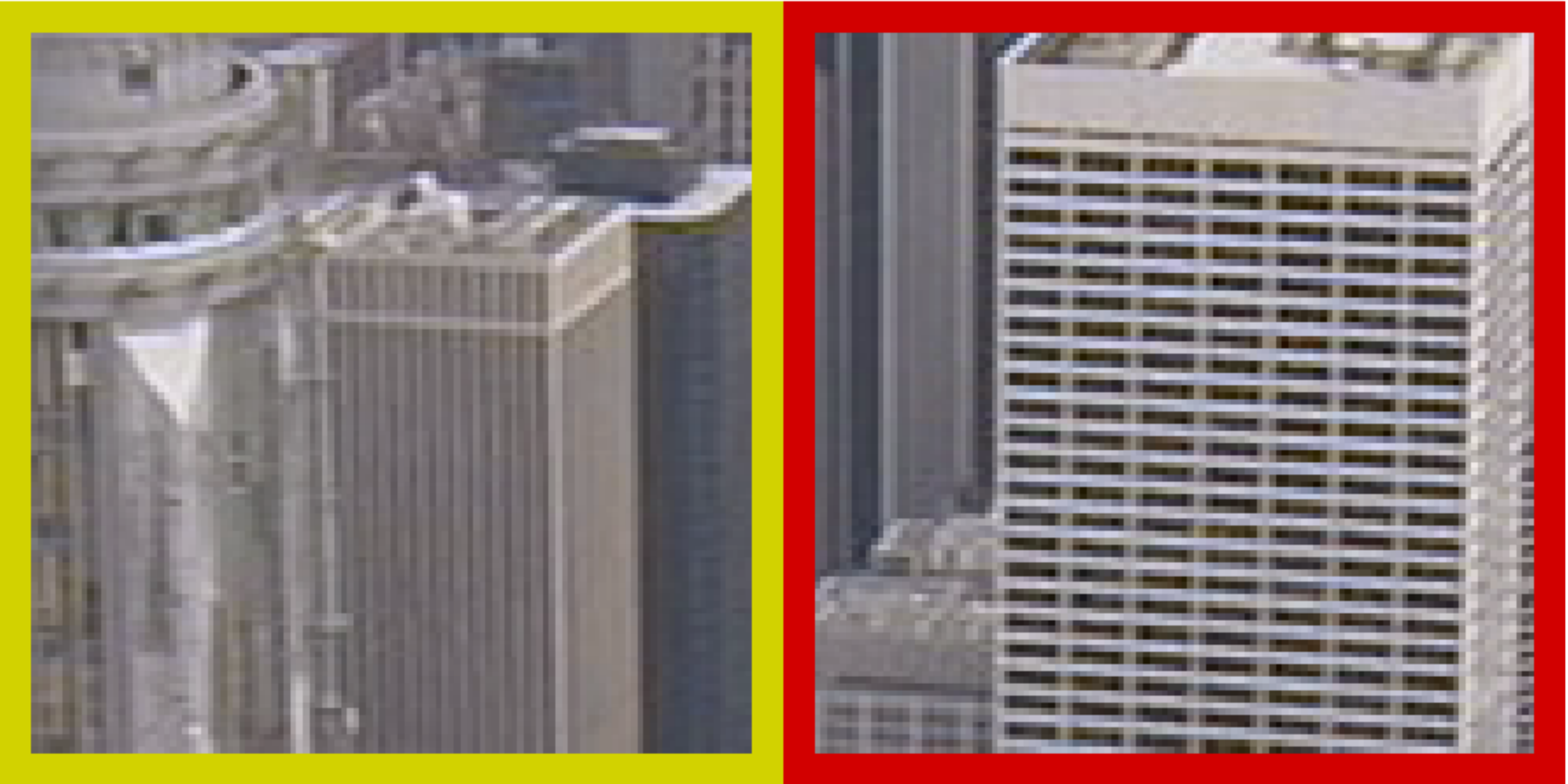}
			\label{fig:City_part7}
		\end{subfigure}
		\\
		\vspace*{-1em}	
		\begin{subfigure}[b]{0.14\textwidth}
			\includegraphics[width=\textwidth]{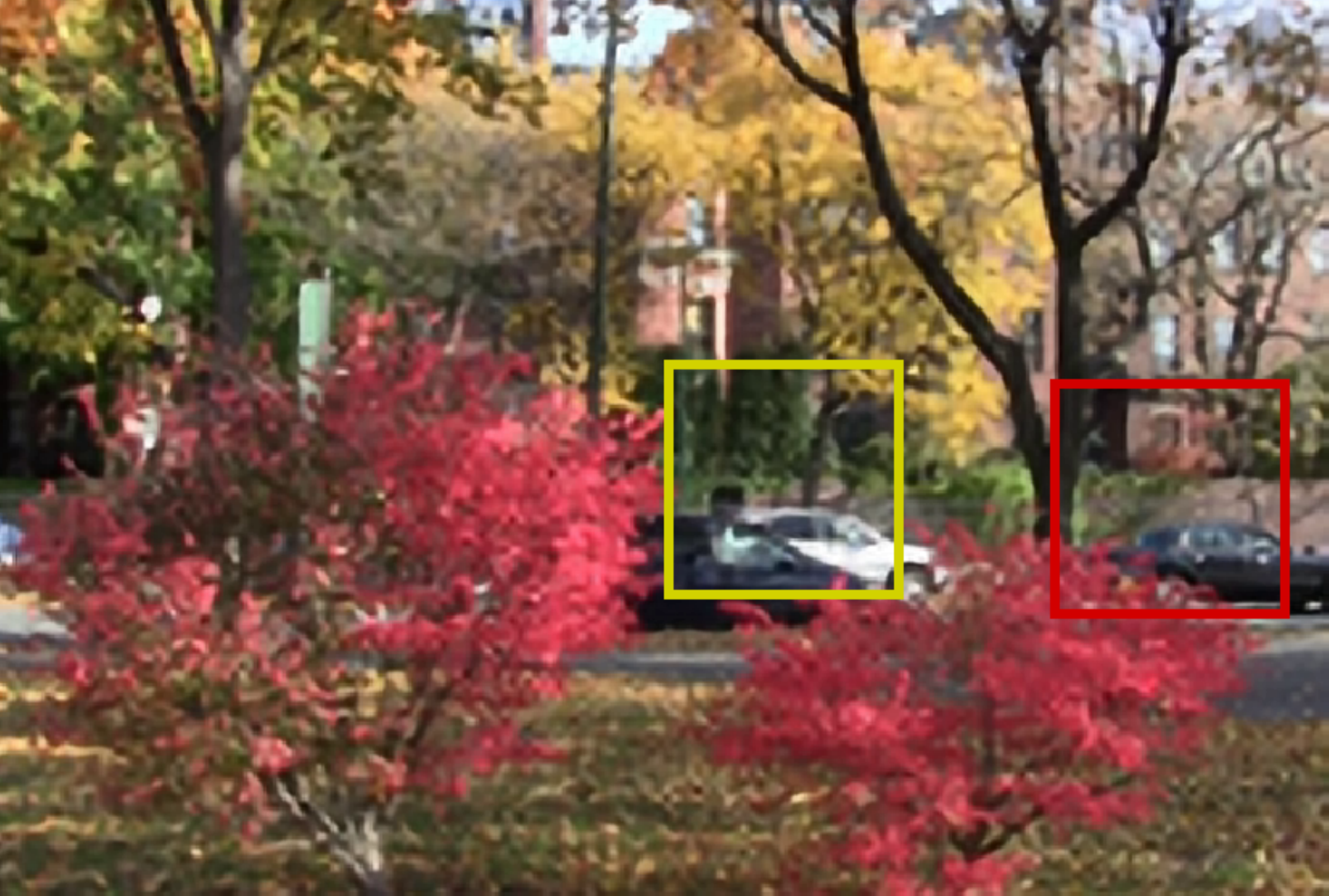}
			\label{fig:Foliage1}
		\end{subfigure}
		\hspace*{-0.4em}
		\begin{subfigure}[b]{0.14\textwidth}
			\includegraphics[width=\textwidth]{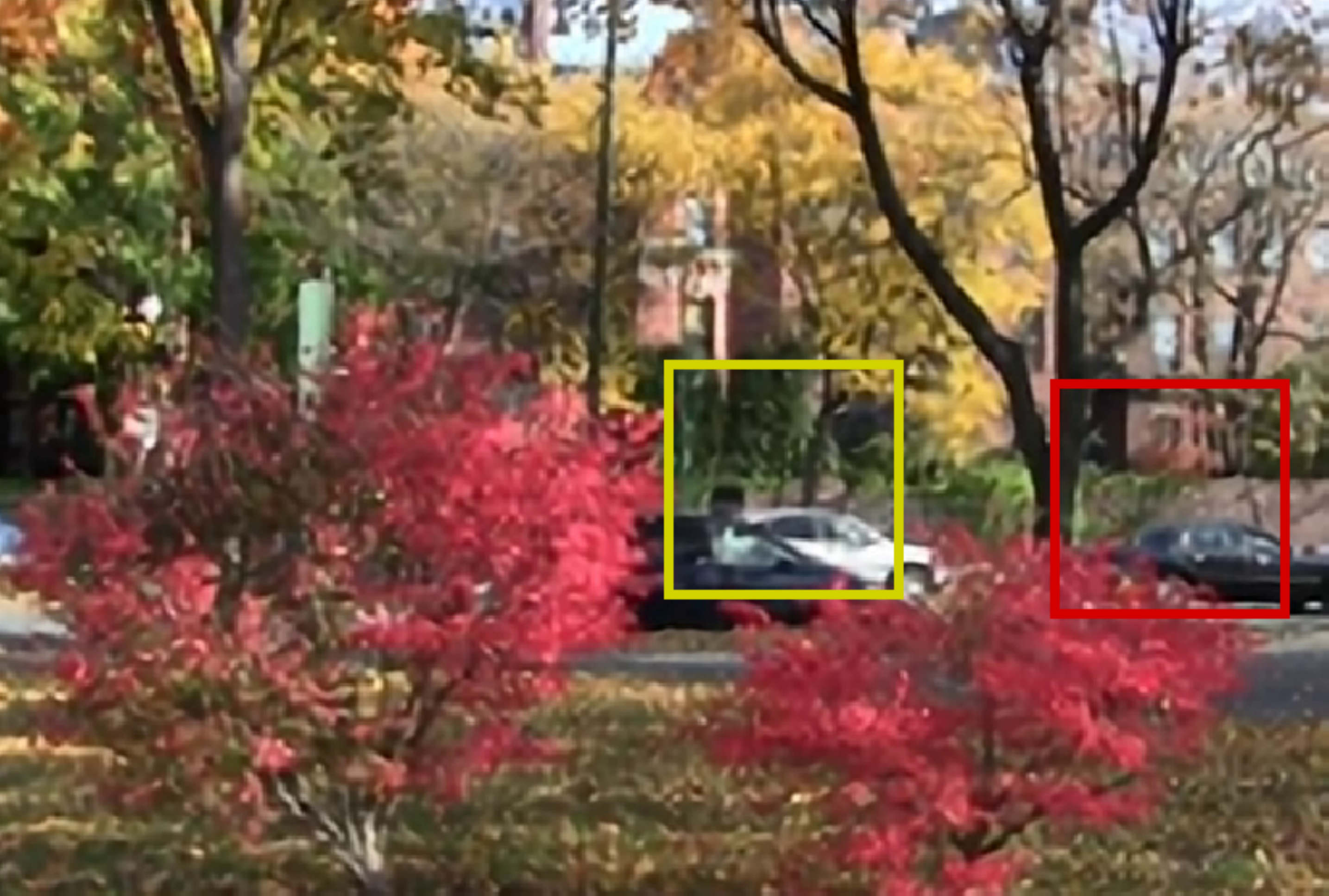}
			\label{fig:Foliage2}
		\end{subfigure}
		\hspace*{-0.4em}
		\begin{subfigure}[b]{0.14\textwidth}
			\includegraphics[width=\textwidth]{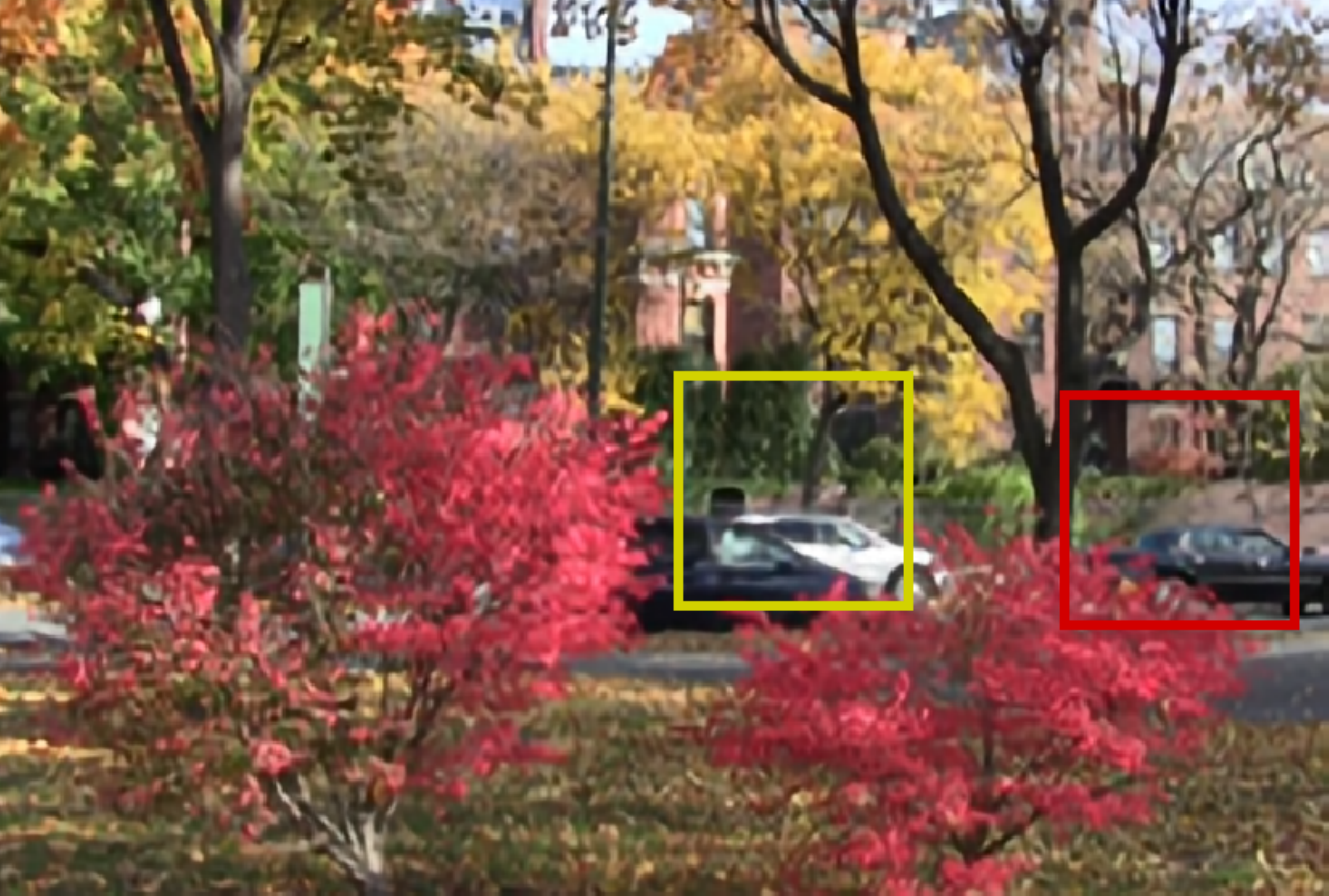}
			\label{fig:Foliage3}
		\end{subfigure}
		\hspace*{-0.4em}
		\begin{subfigure}[b]{0.14\textwidth}
			\includegraphics[width=\textwidth]{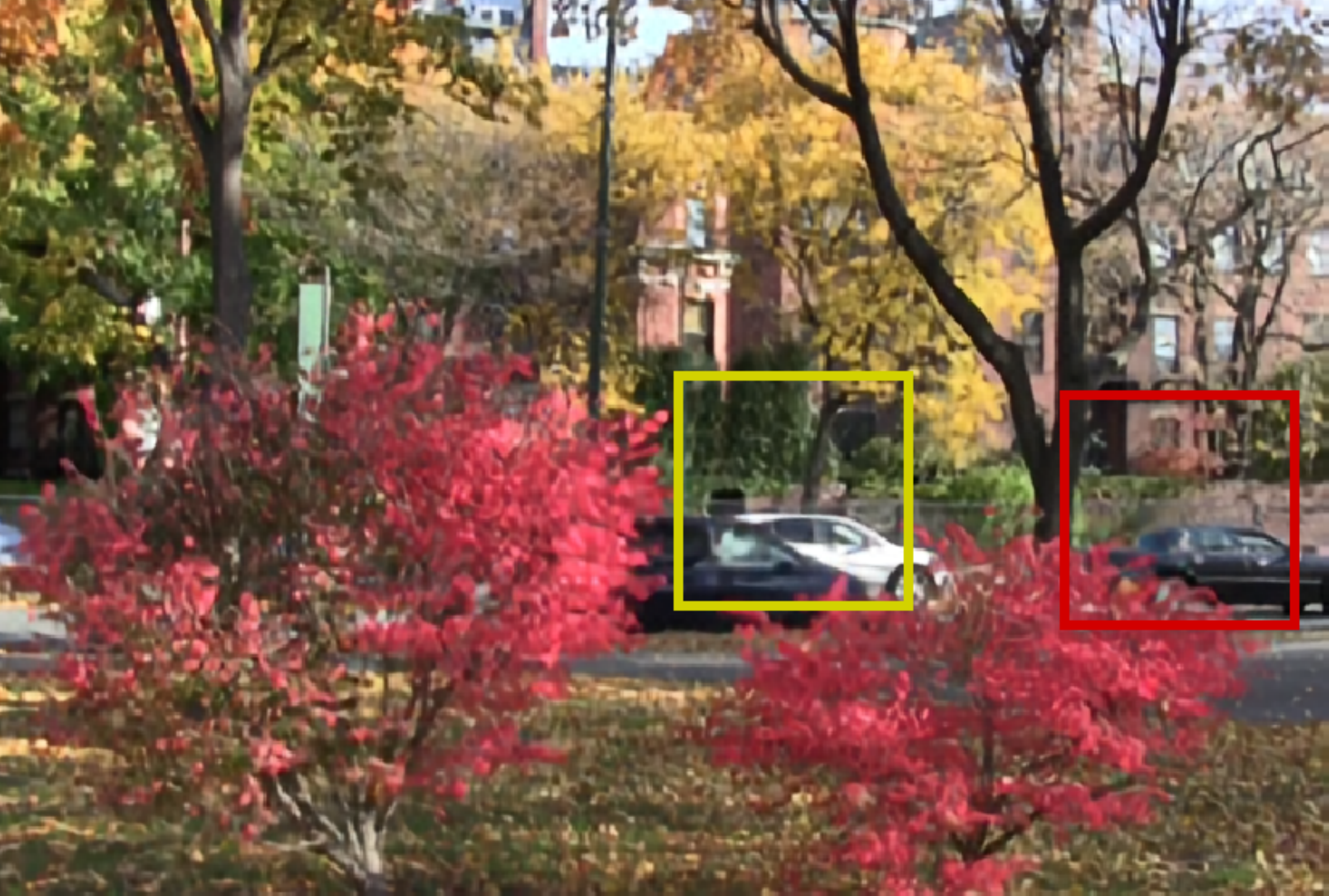}
			\label{fig:Foliage4}
		\end{subfigure}
		\hspace*{-0.4em}
		\begin{subfigure}[b]{0.14\textwidth}
			\includegraphics[width=\textwidth]{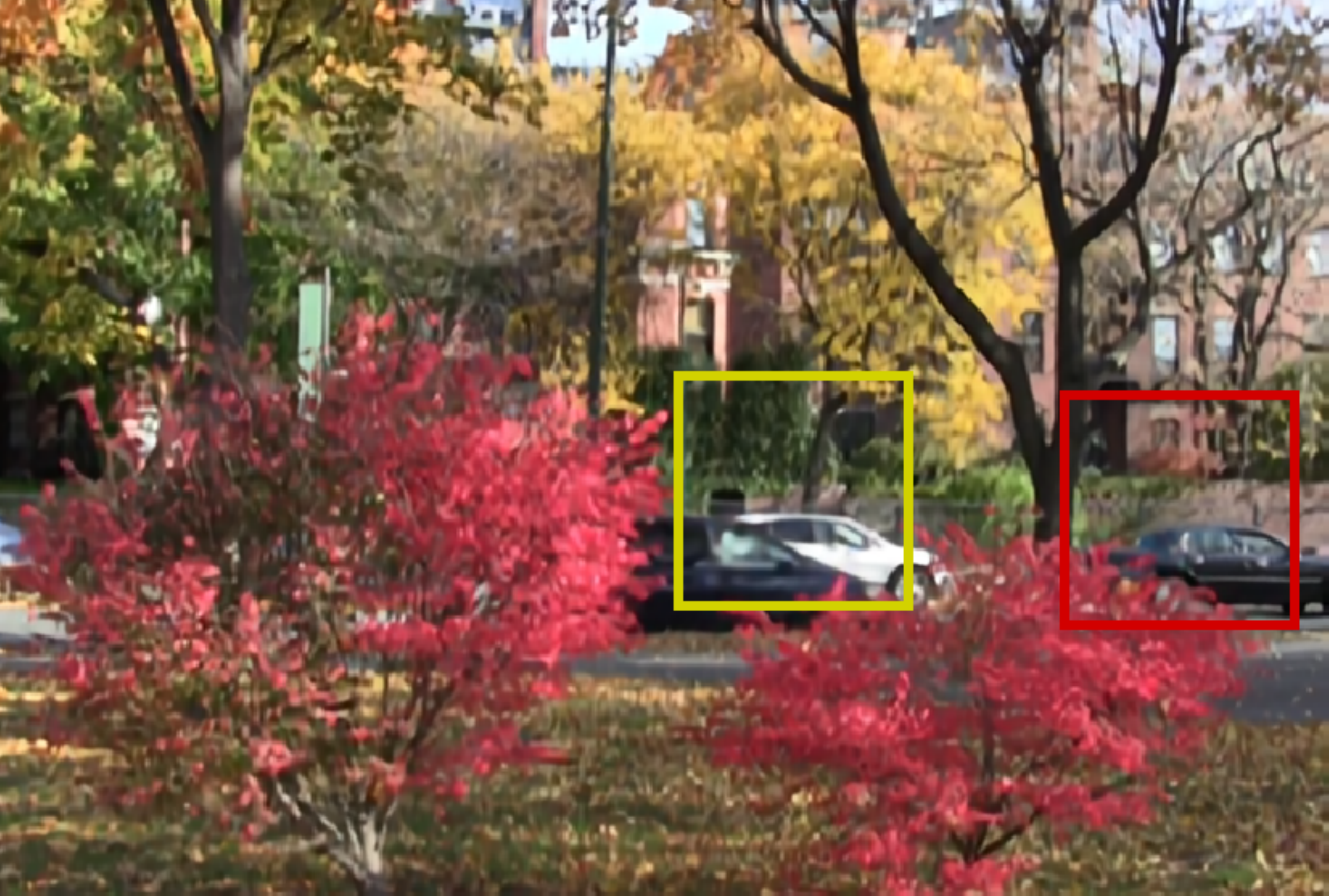}
			\label{fig:Foliage5}
		\end{subfigure}
		\hspace*{-0.4em}
		\begin{subfigure}[b]{0.14\textwidth}
			\includegraphics[width=\textwidth]{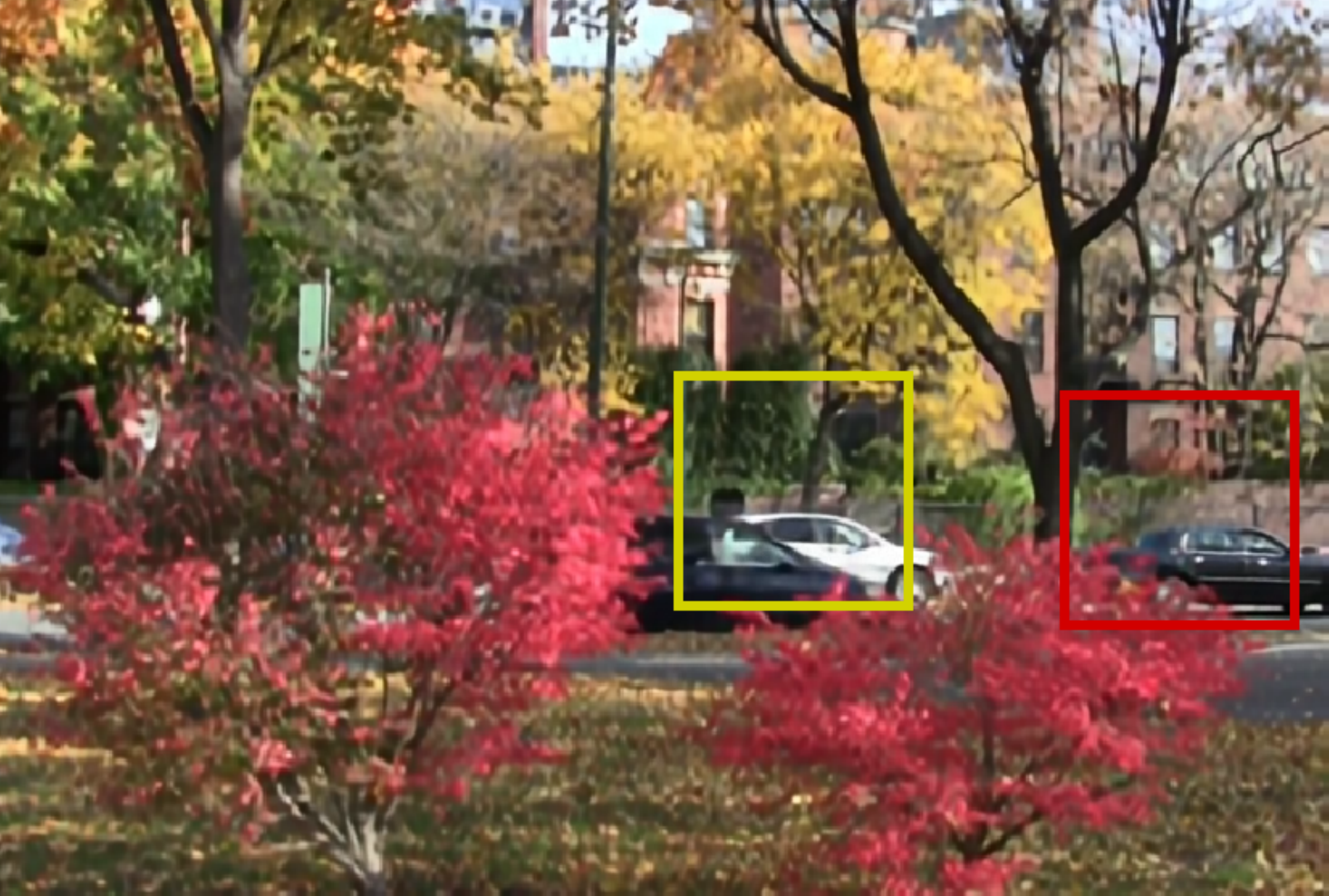}
			\label{fig:Foliage6}
		\end{subfigure}
		\hspace*{-0.4em}
		\begin{subfigure}[b]{0.14\textwidth}
			\includegraphics[width=\textwidth]{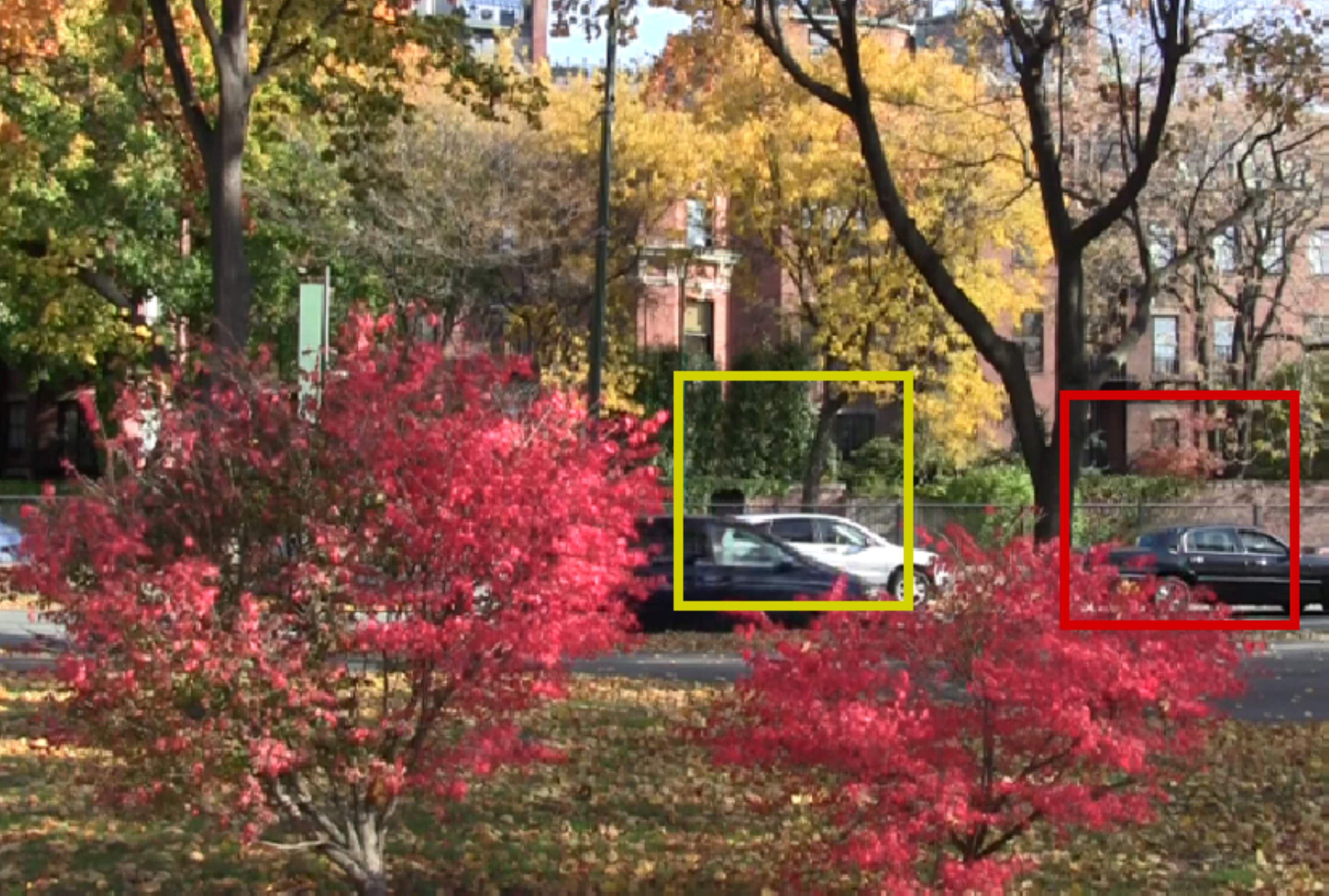}
			\label{fig:Foliage7}
		\end{subfigure}
		\\
		\vspace*{-1.2em}	
		\begin{subfigure}[b]{0.14\textwidth}
			\includegraphics[width=\textwidth]{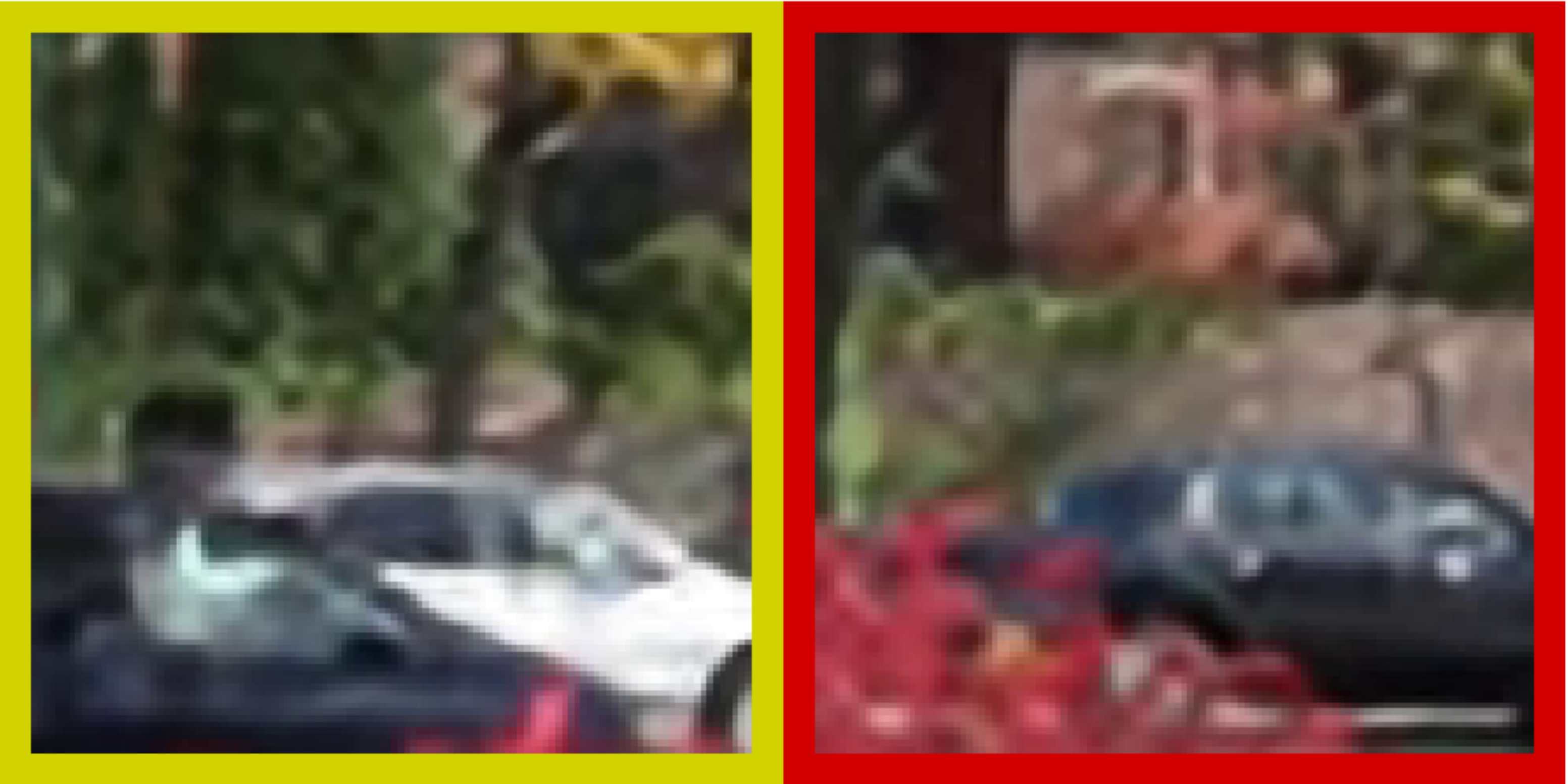}
			\label{fig:Foliage_part1}
		\end{subfigure}
		\hspace*{-0.4em}
		\begin{subfigure}[b]{0.14\textwidth}
			\includegraphics[width=\textwidth]{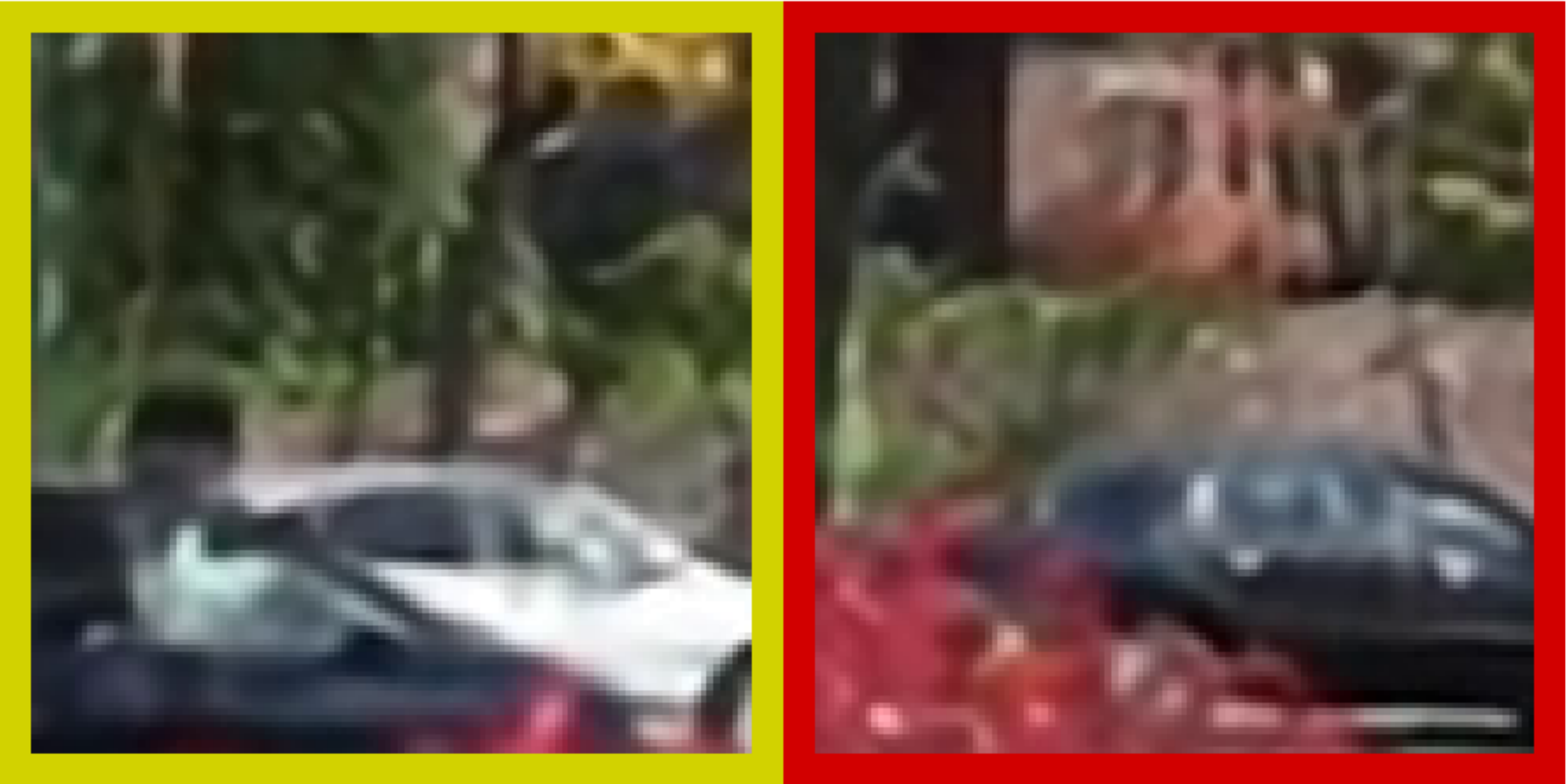}
			\label{fig:Foliage_part2}
		\end{subfigure}
		\hspace*{-0.4em}
		\begin{subfigure}[b]{0.14\textwidth}
			\includegraphics[width=\textwidth]{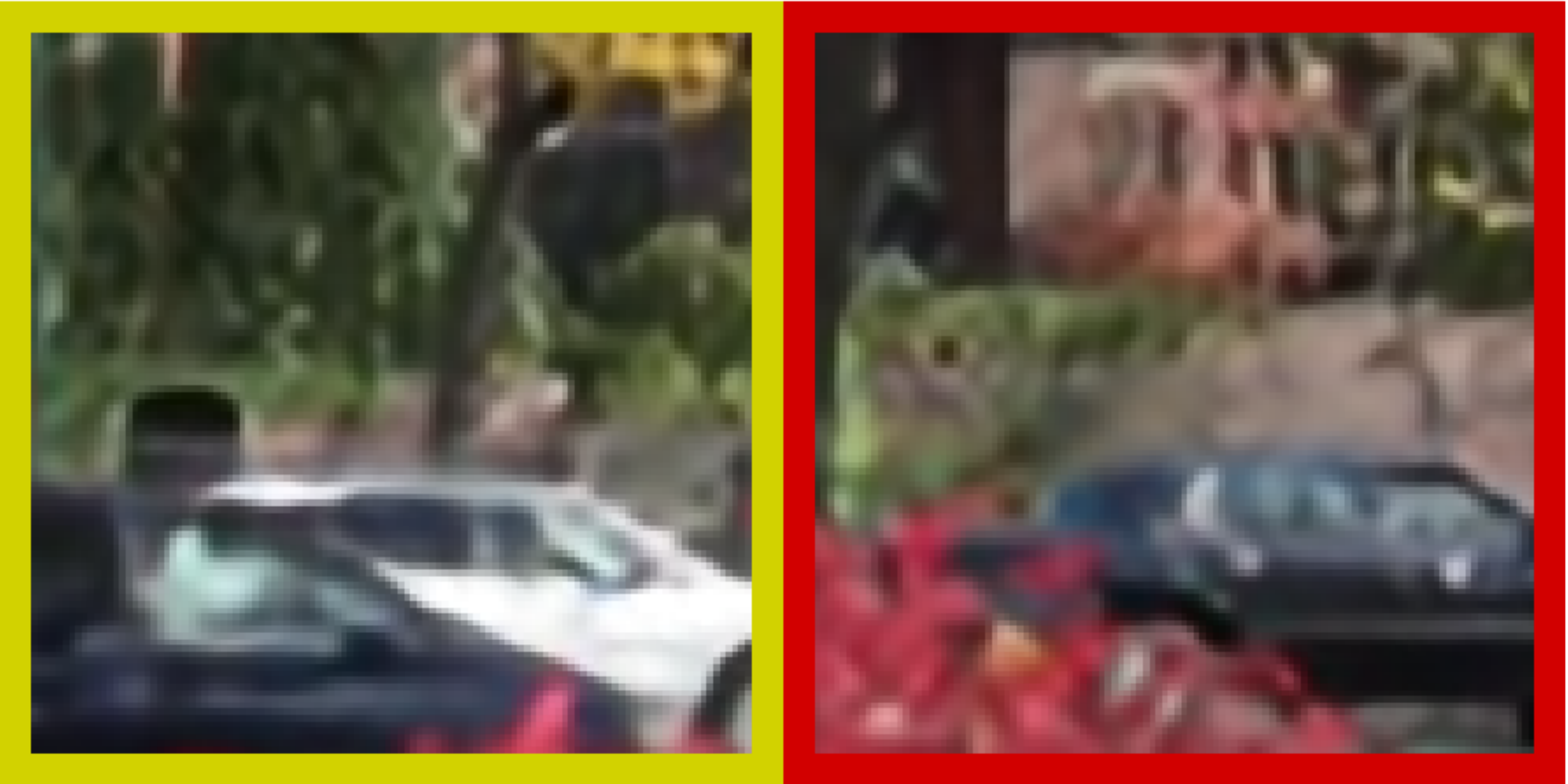}
			\label{fig:Foliage_part3}
		\end{subfigure}
		\hspace*{-0.4em}
		\begin{subfigure}[b]{0.14\textwidth}
			\includegraphics[width=\textwidth]{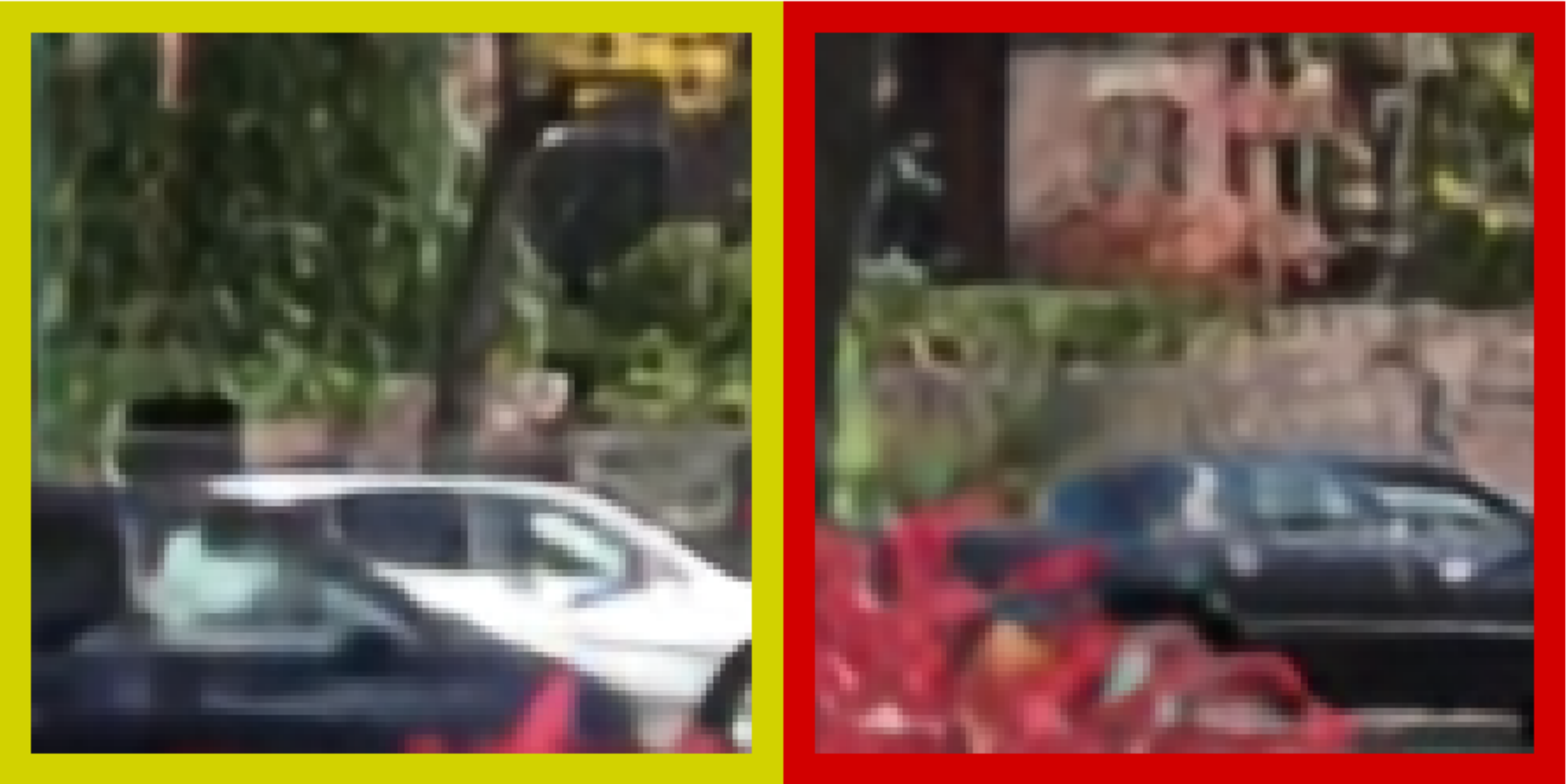}
			\label{fig:Foliage_part4}
		\end{subfigure}
		\hspace*{-0.4em}
		\begin{subfigure}[b]{0.14\textwidth}
			\includegraphics[width=\textwidth]{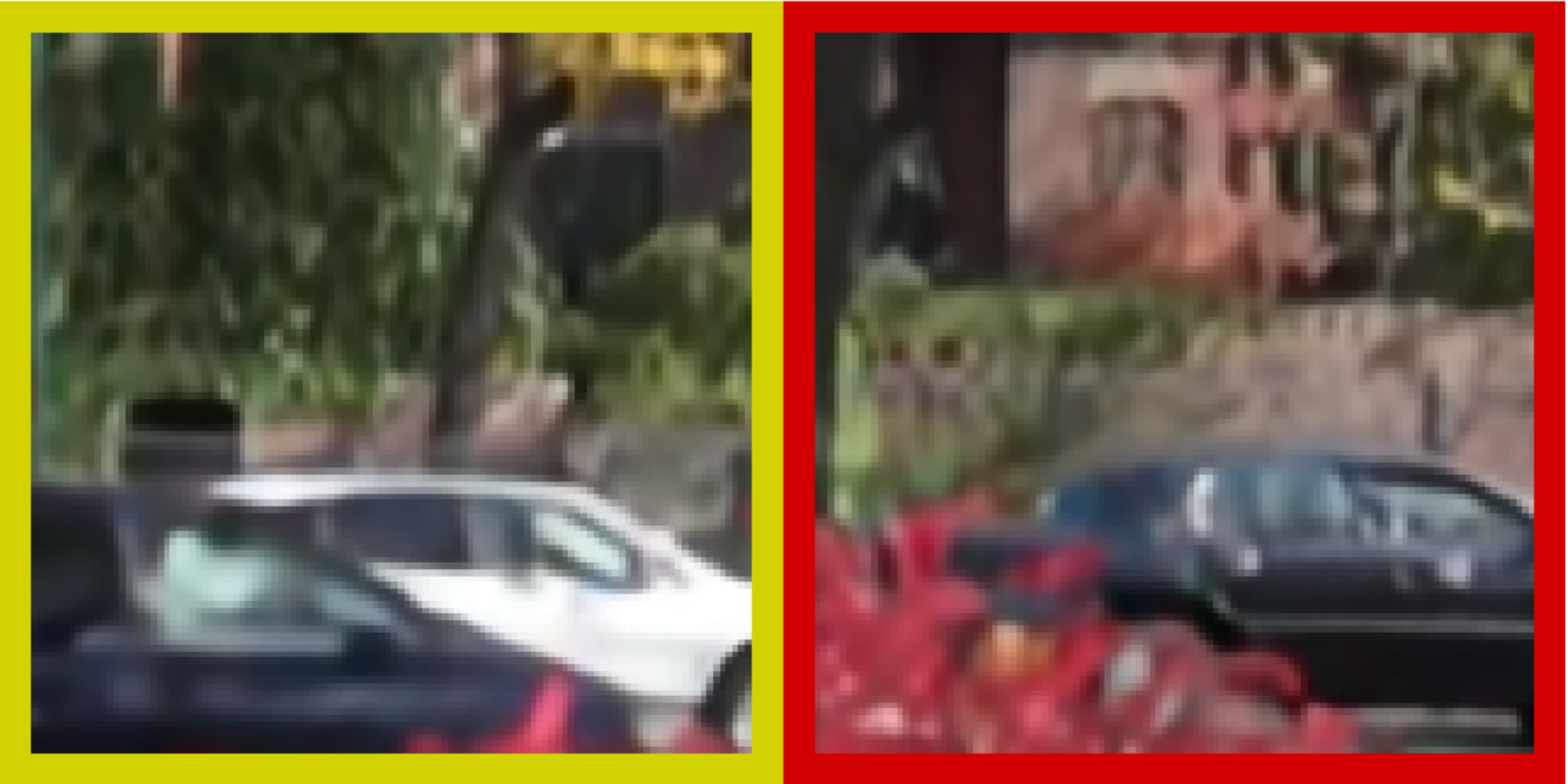}
			\label{fig:Foliage_part5}
		\end{subfigure}
		\hspace*{-0.4em}
		\begin{subfigure}[b]{0.14\textwidth}
			\includegraphics[width=\textwidth]{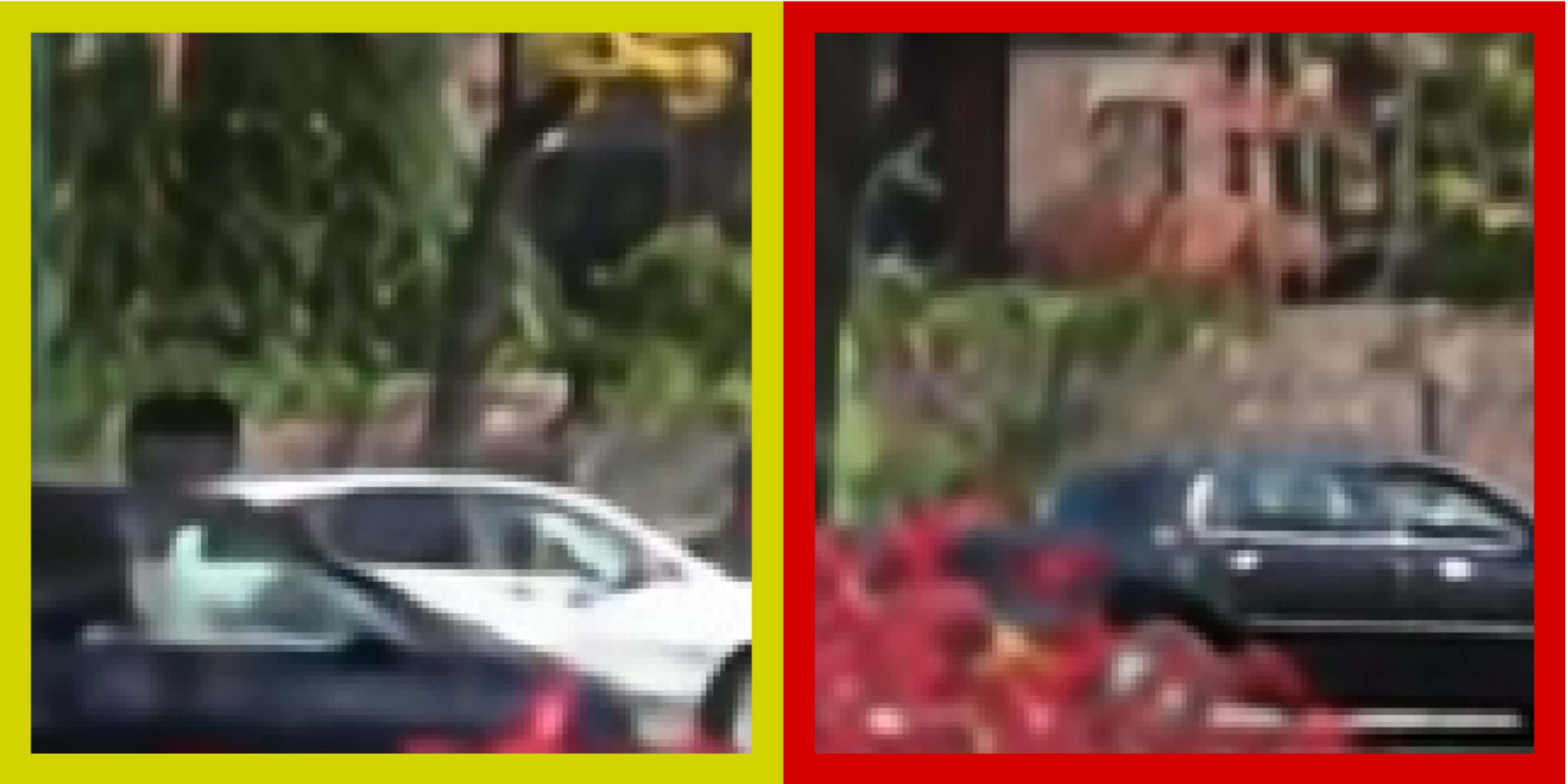}
			\label{fig:Foliage_part6}
		\end{subfigure}
		\hspace*{-0.4em}
		\begin{subfigure}[b]{0.14\textwidth}
			\includegraphics[width=\textwidth]{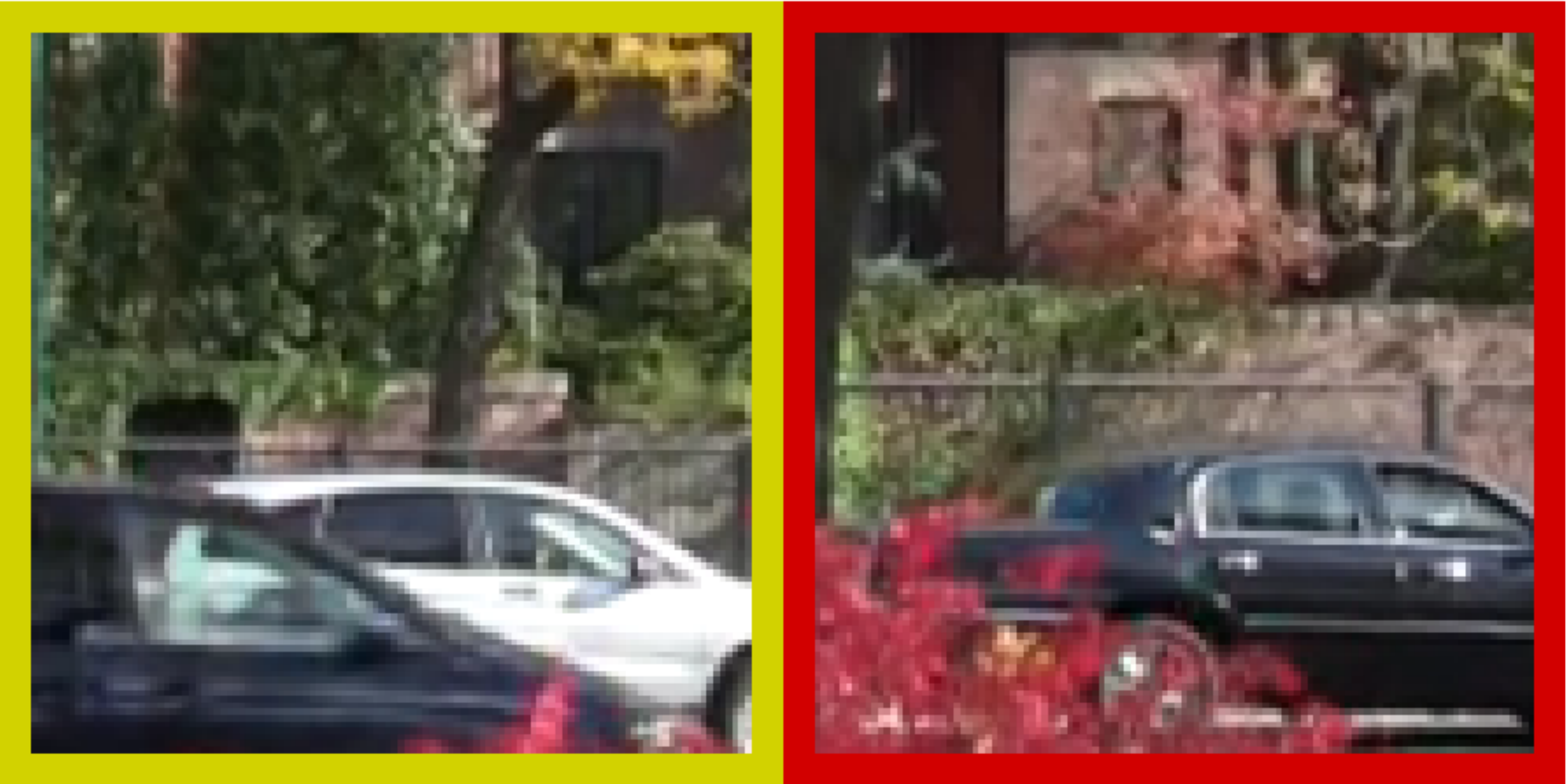}
			\label{fig:Foliage_part7}
		\end{subfigure}
		\\
		\vspace*{-1em}
		\begin{subfigure}[b]{0.14\textwidth}
			\includegraphics[width=\textwidth]{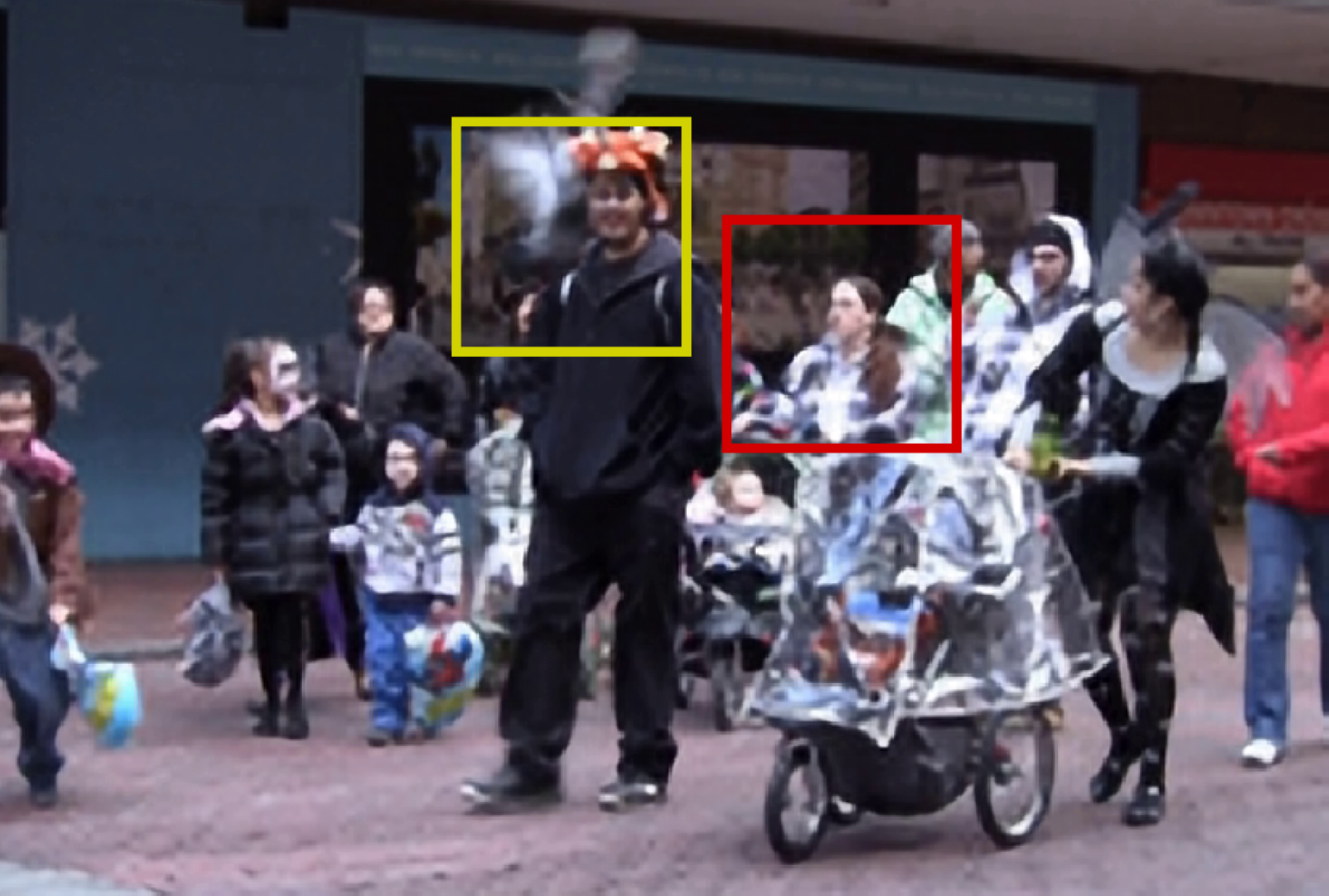}
			\label{fig:Walk1}
		\end{subfigure}
		\hspace*{-0.4em}
		\begin{subfigure}[b]{0.14\textwidth}
			\includegraphics[width=\textwidth]{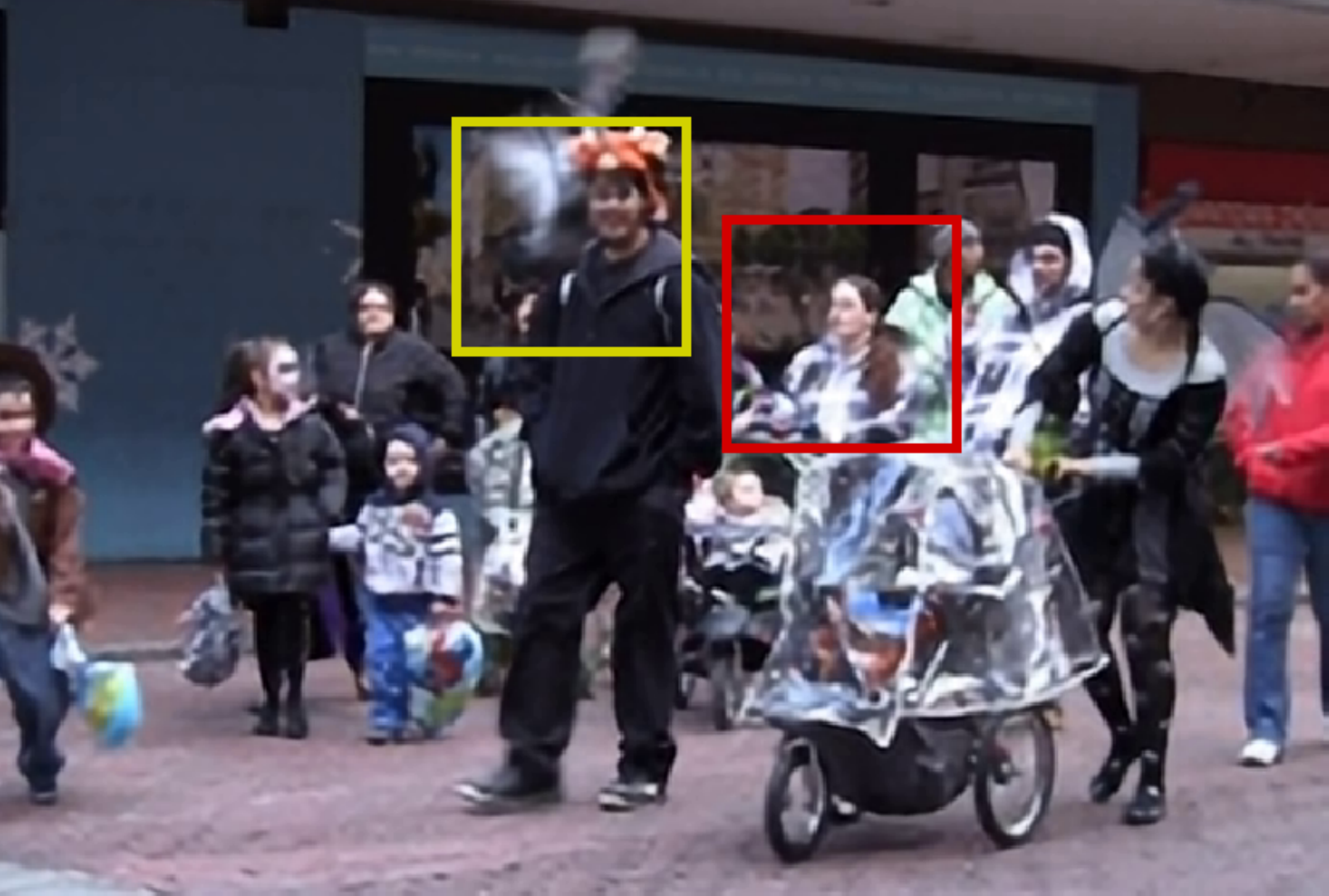}
			\label{fig:Walk2}
		\end{subfigure}
		\hspace*{-0.4em}
		\begin{subfigure}[b]{0.14\textwidth}
			\includegraphics[width=\textwidth]{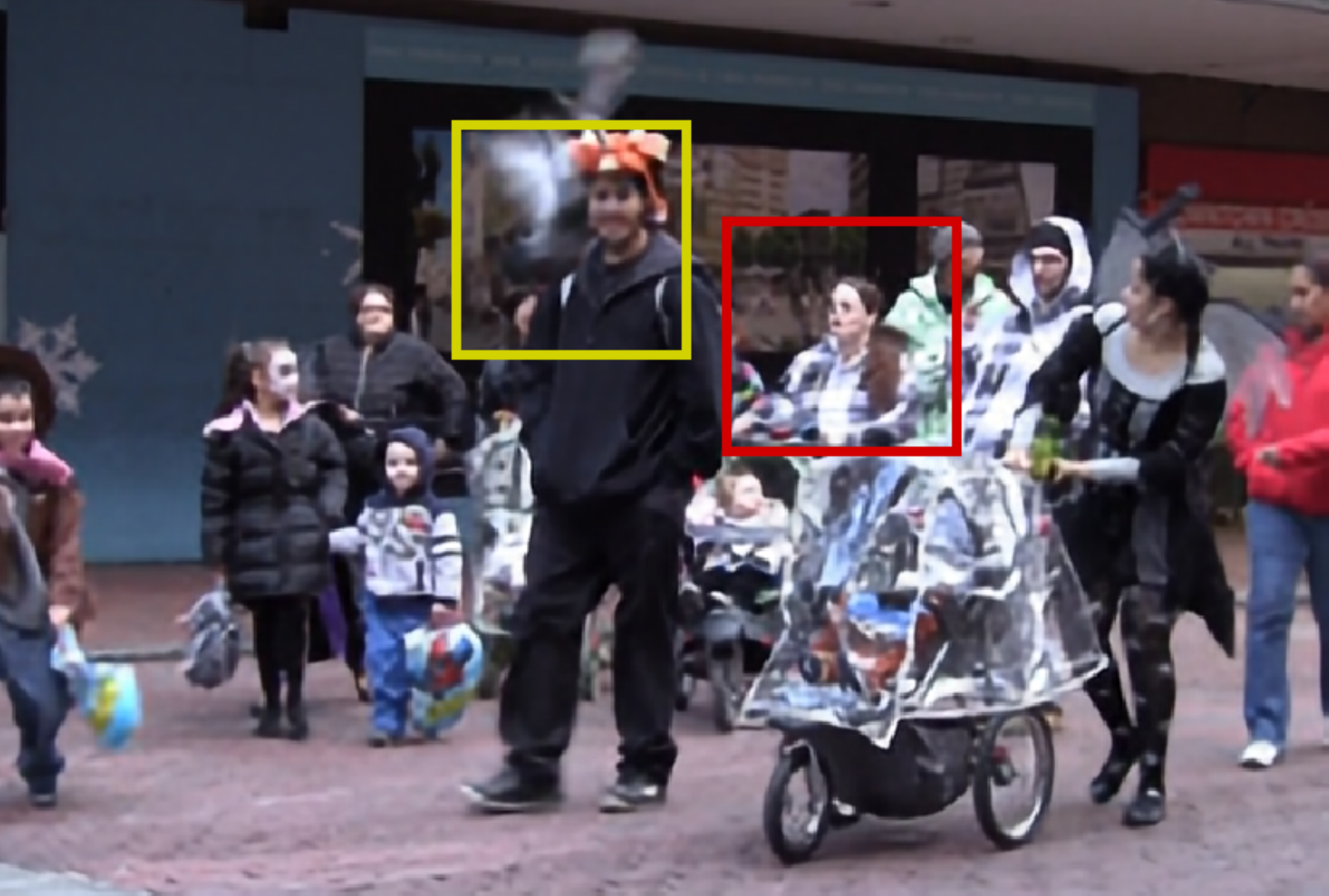}
			\label{fig:Walk3}
		\end{subfigure}
		\hspace*{-0.4em}
		\begin{subfigure}[b]{0.14\textwidth}
			\includegraphics[width=\textwidth]{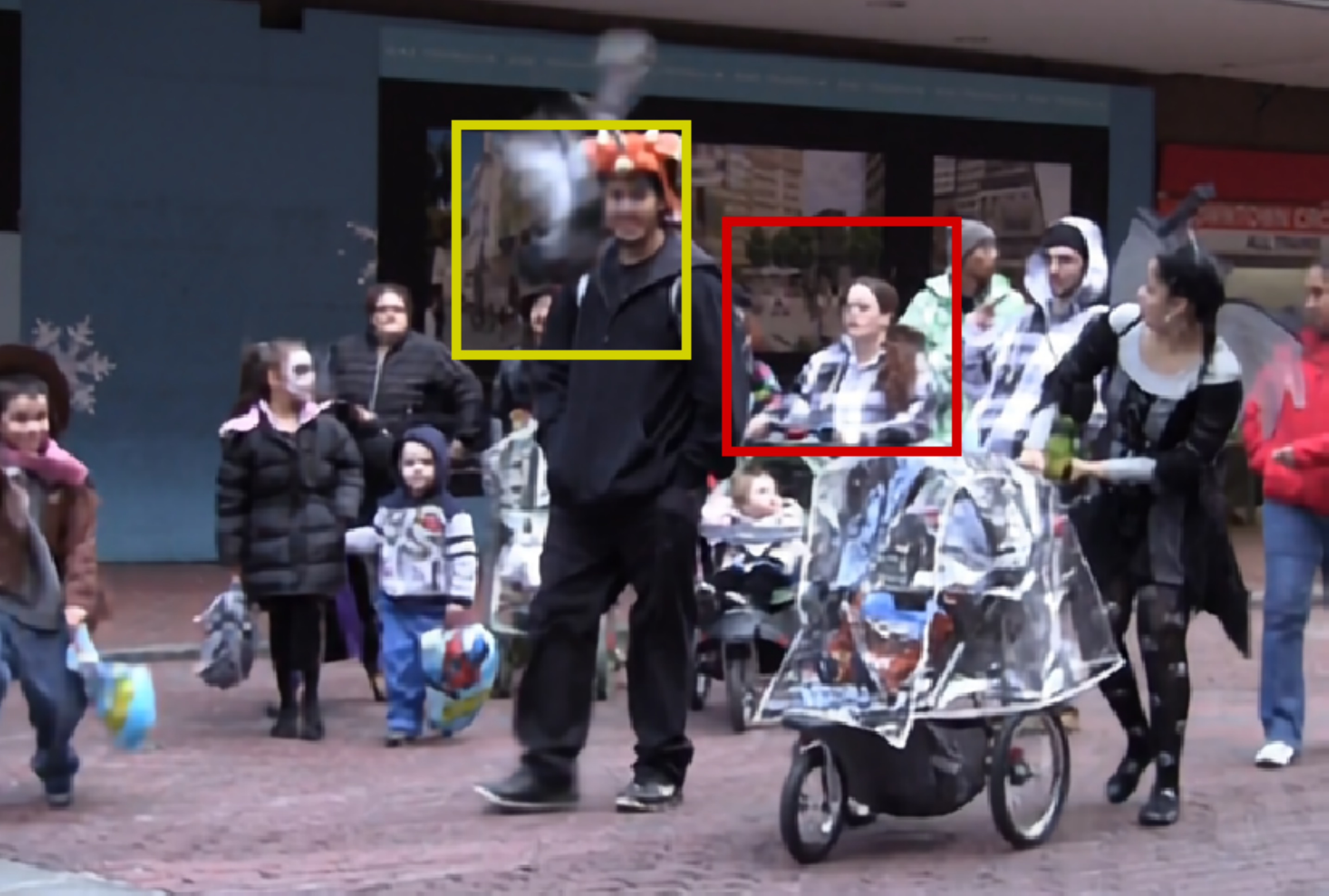}
			\label{fig:Walk4}
		\end{subfigure}
		\hspace*{-0.4em}
		\begin{subfigure}[b]{0.14\textwidth}
			\includegraphics[width=\textwidth]{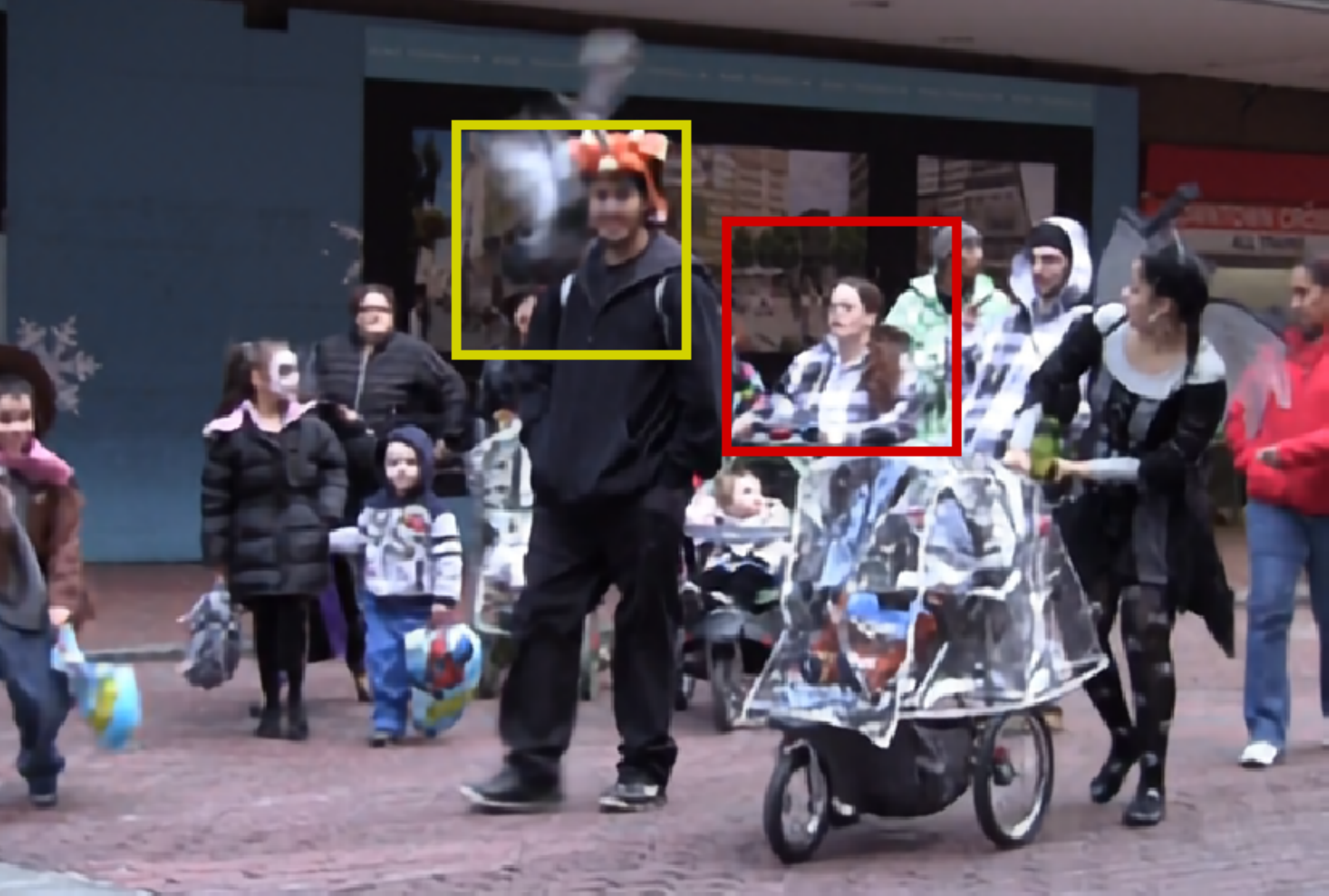}
			\label{fig:Walk5}
		\end{subfigure}
		\hspace*{-0.4em}
		\begin{subfigure}[b]{0.14\textwidth}
			\includegraphics[width=\textwidth]{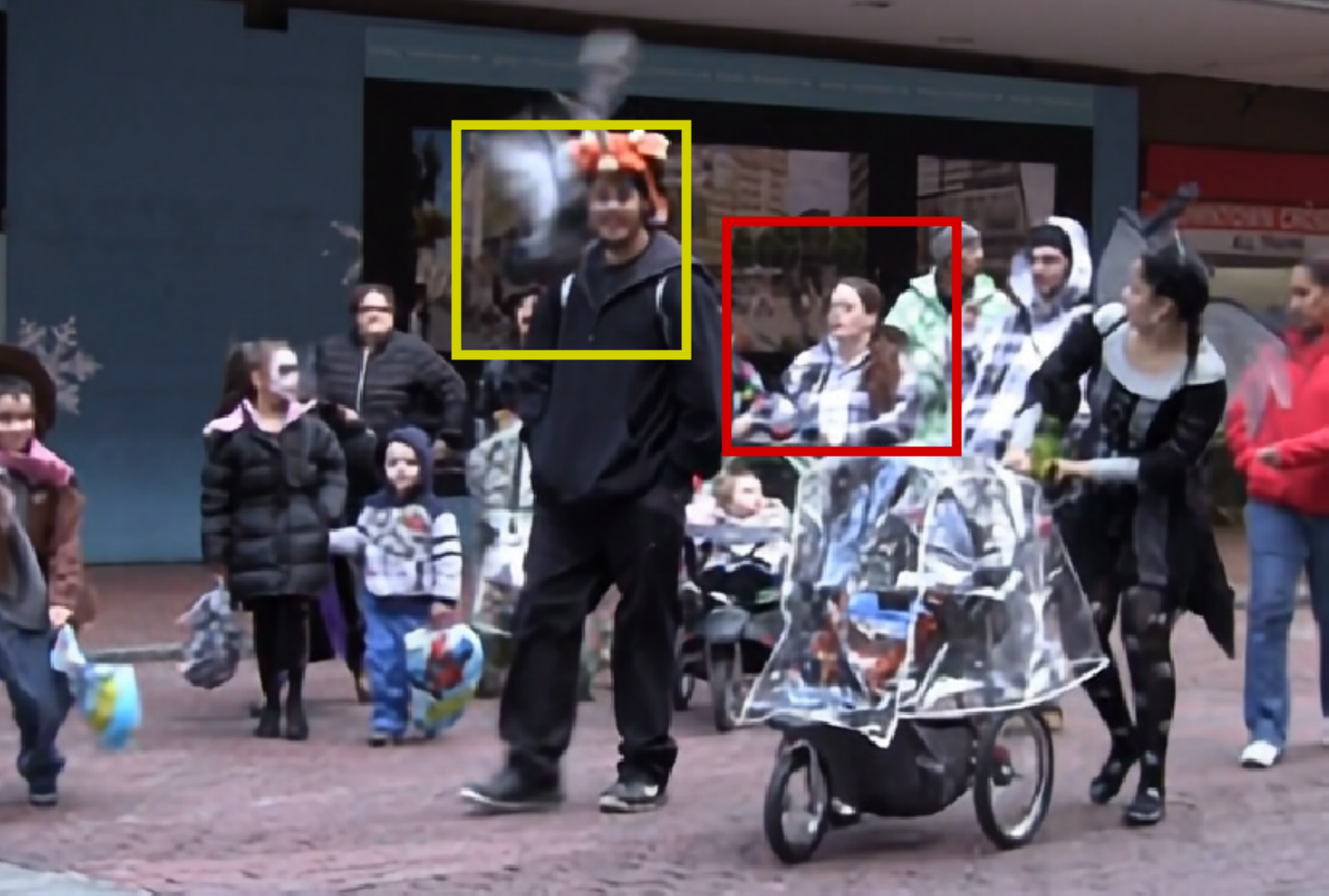}
			\label{fig:Walk6}
		\end{subfigure}
		\hspace*{-0.4em}
		\begin{subfigure}[b]{0.14\textwidth}
			\includegraphics[width=\textwidth]{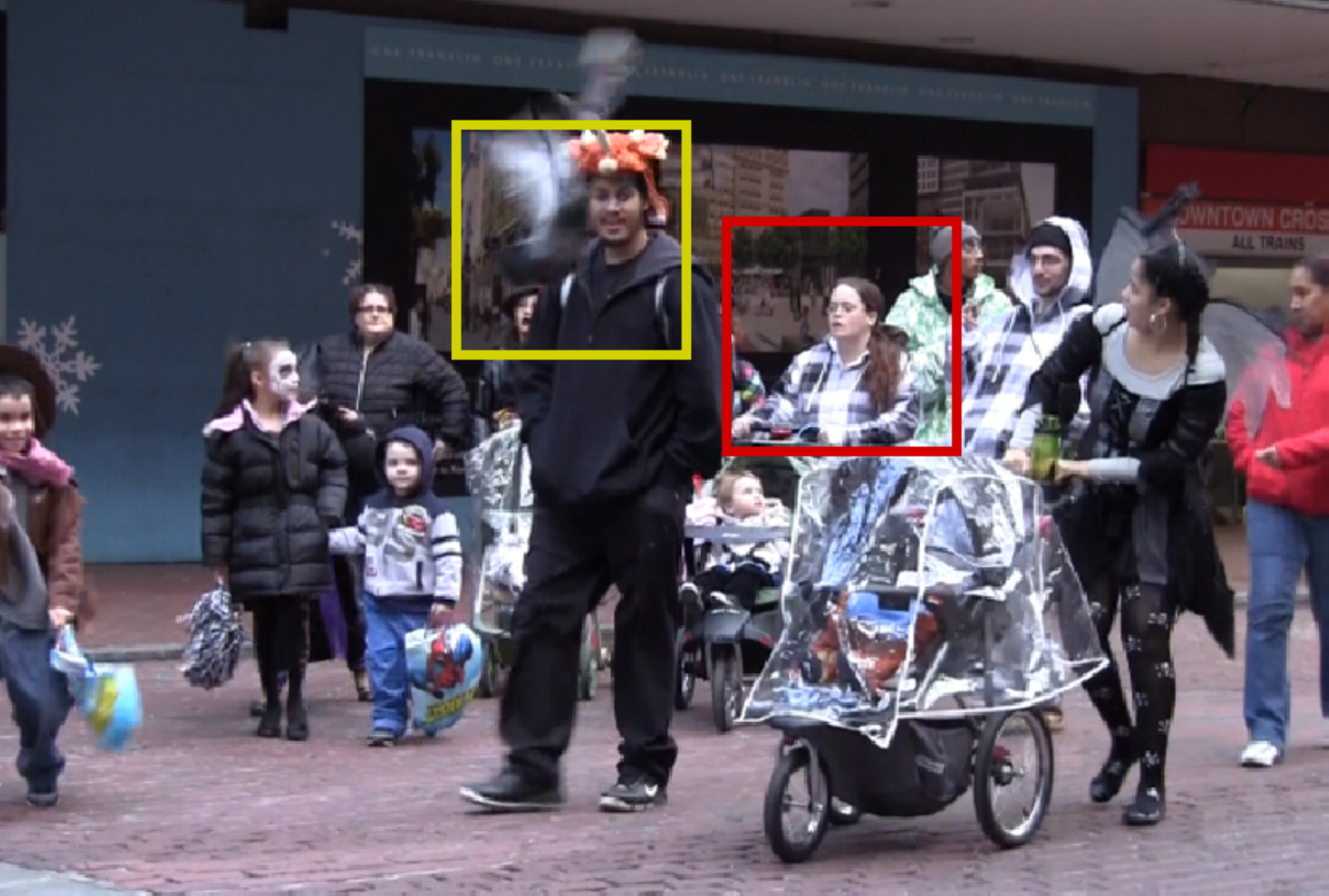}
			\label{fig:Walk7}
		\end{subfigure}
		\\
		\vspace*{-1.2em}
		\begin{subfigure}[b]{0.14\textwidth}
			\includegraphics[width=\textwidth]{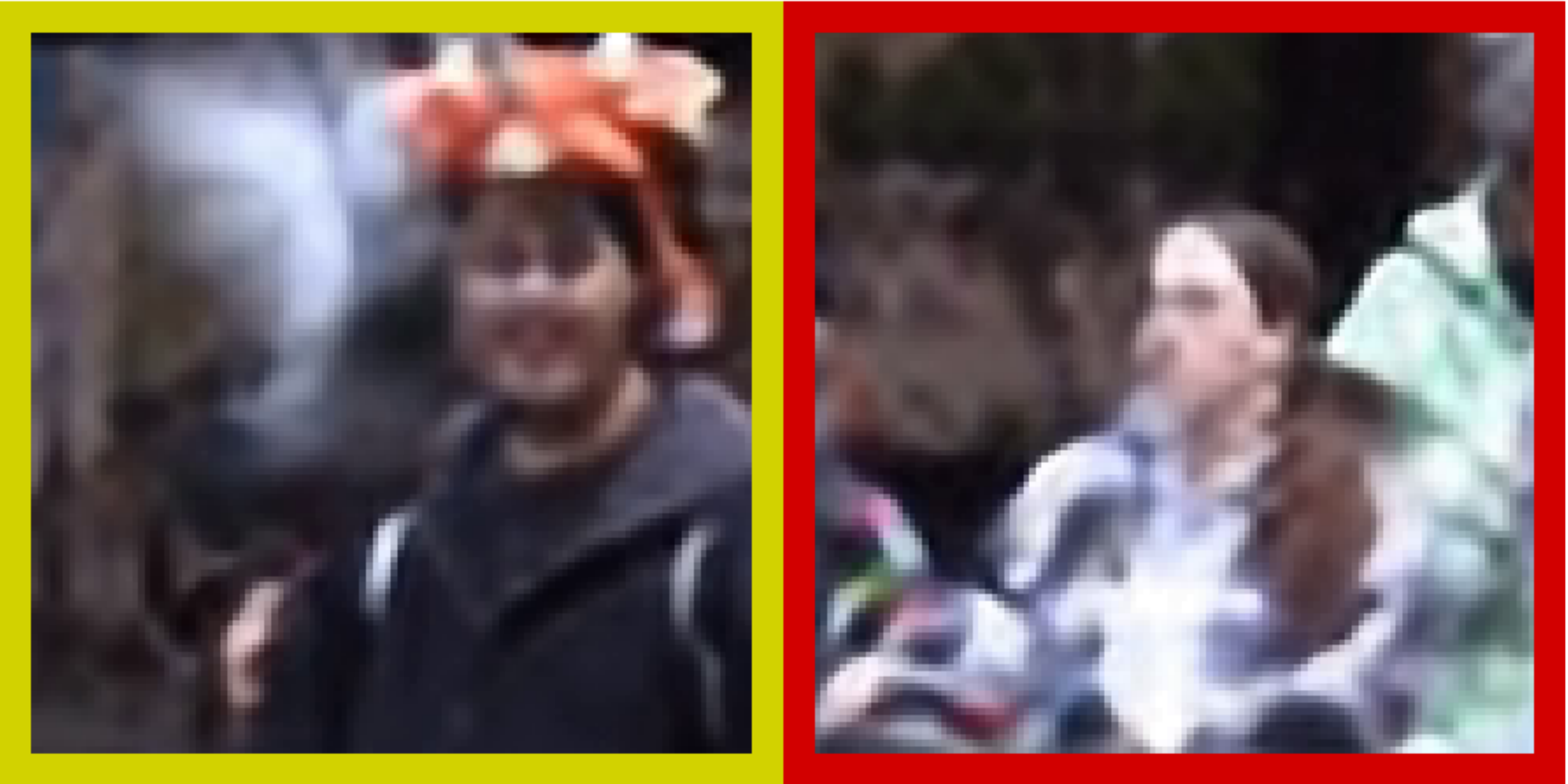}\vspace*{-1.5em}
			\label{fig:Walk_part1}
			\caption{VSRNet \cite{kappeler2016video}}
		\end{subfigure}
		\hspace*{-0.4em}
		\begin{subfigure}[b]{0.14\textwidth}
			\includegraphics[width=\textwidth]{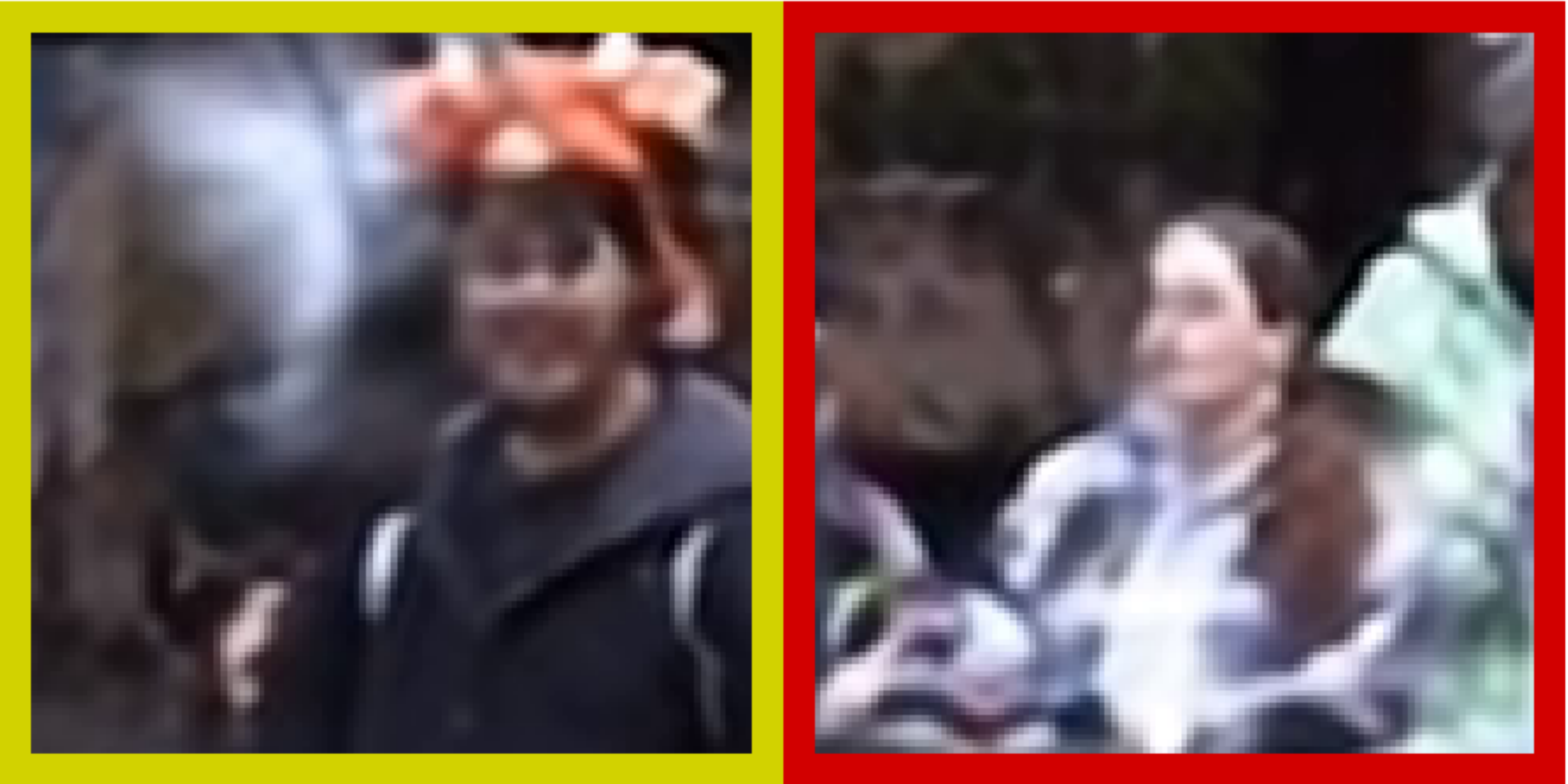}\vspace*{-1.5em}
			\label{fig:Walk_part2}
			\caption{VESPCN \cite{caballero2017real}}
		\end{subfigure}
		\hspace*{-0.4em}
		\begin{subfigure}[b]{0.14\textwidth}
			\includegraphics[width=\textwidth]{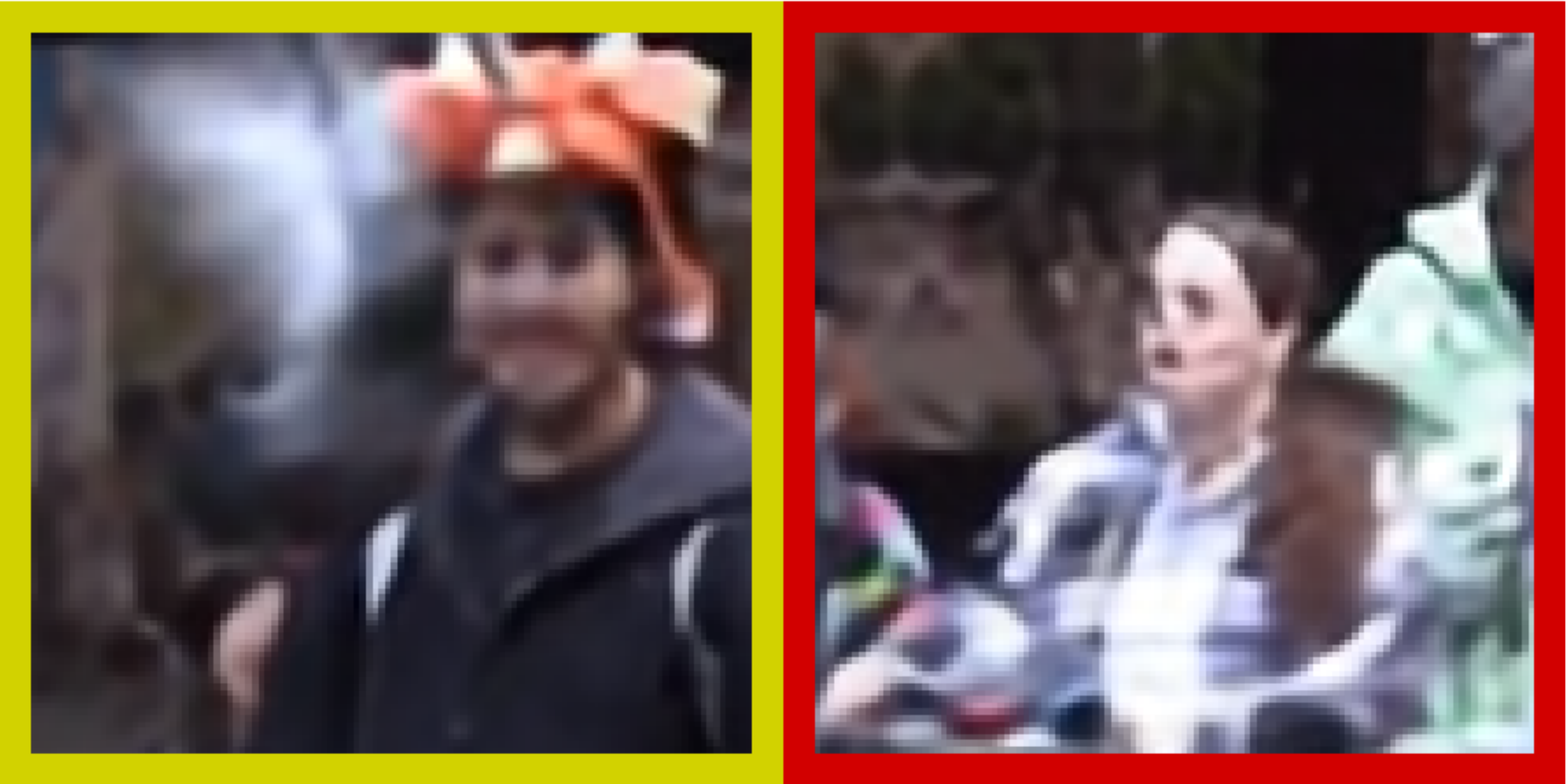}\vspace*{-1.5em}
			\label{fig:Walk_part3}
			\caption{SPMC \cite{tao2017detail}}
		\end{subfigure}
		\hspace*{-0.4em}
		\begin{subfigure}[b]{0.14\textwidth}
			\includegraphics[width=\textwidth]{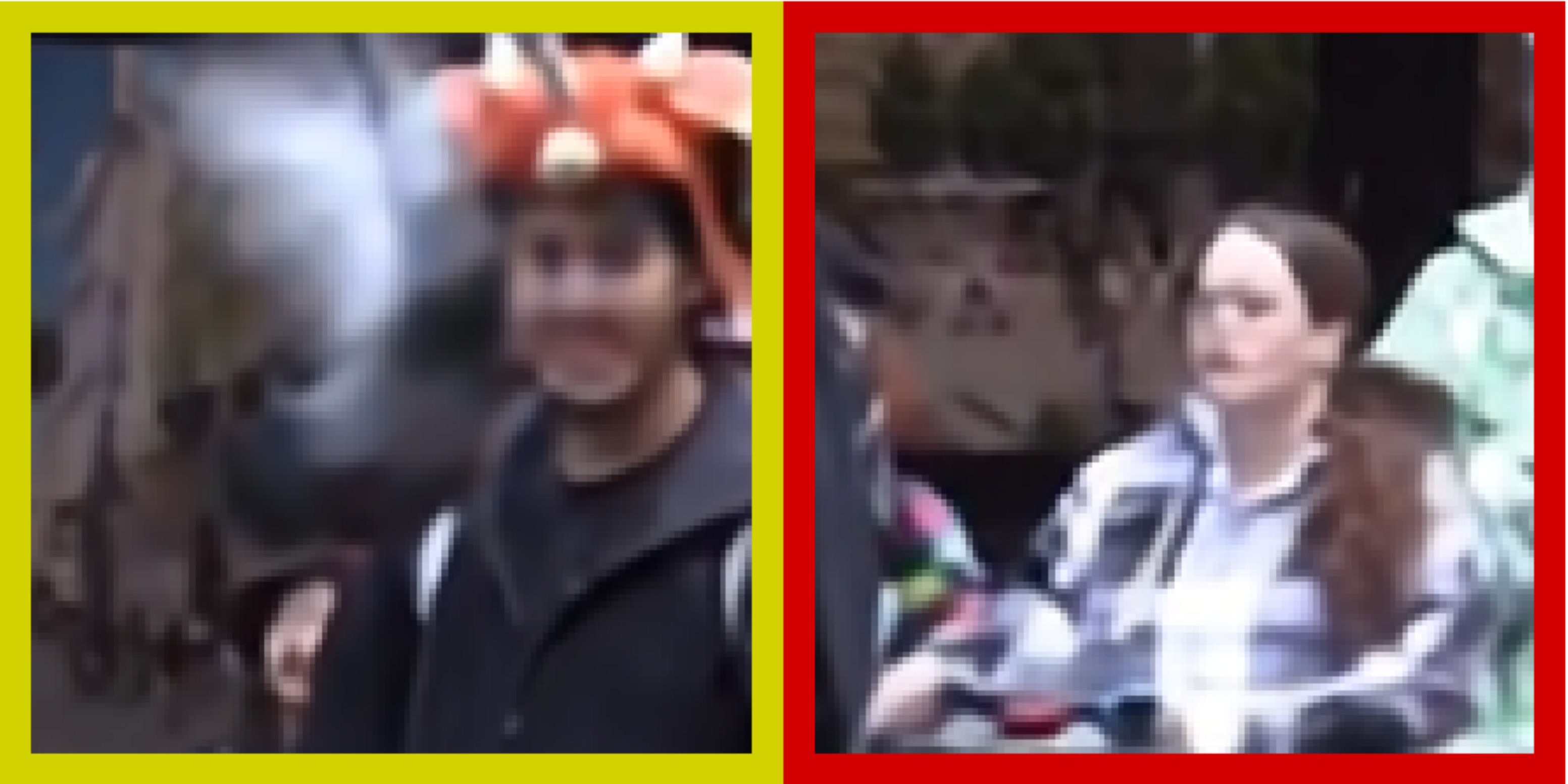}\vspace*{-1.5em}
			\label{fig:Walk_part4}
			\caption{FRVSR \cite{sajjadi2018frame}}
		\end{subfigure}
		\hspace*{-0.4em}
		\begin{subfigure}[b]{0.14\textwidth}
			\includegraphics[width=\textwidth]{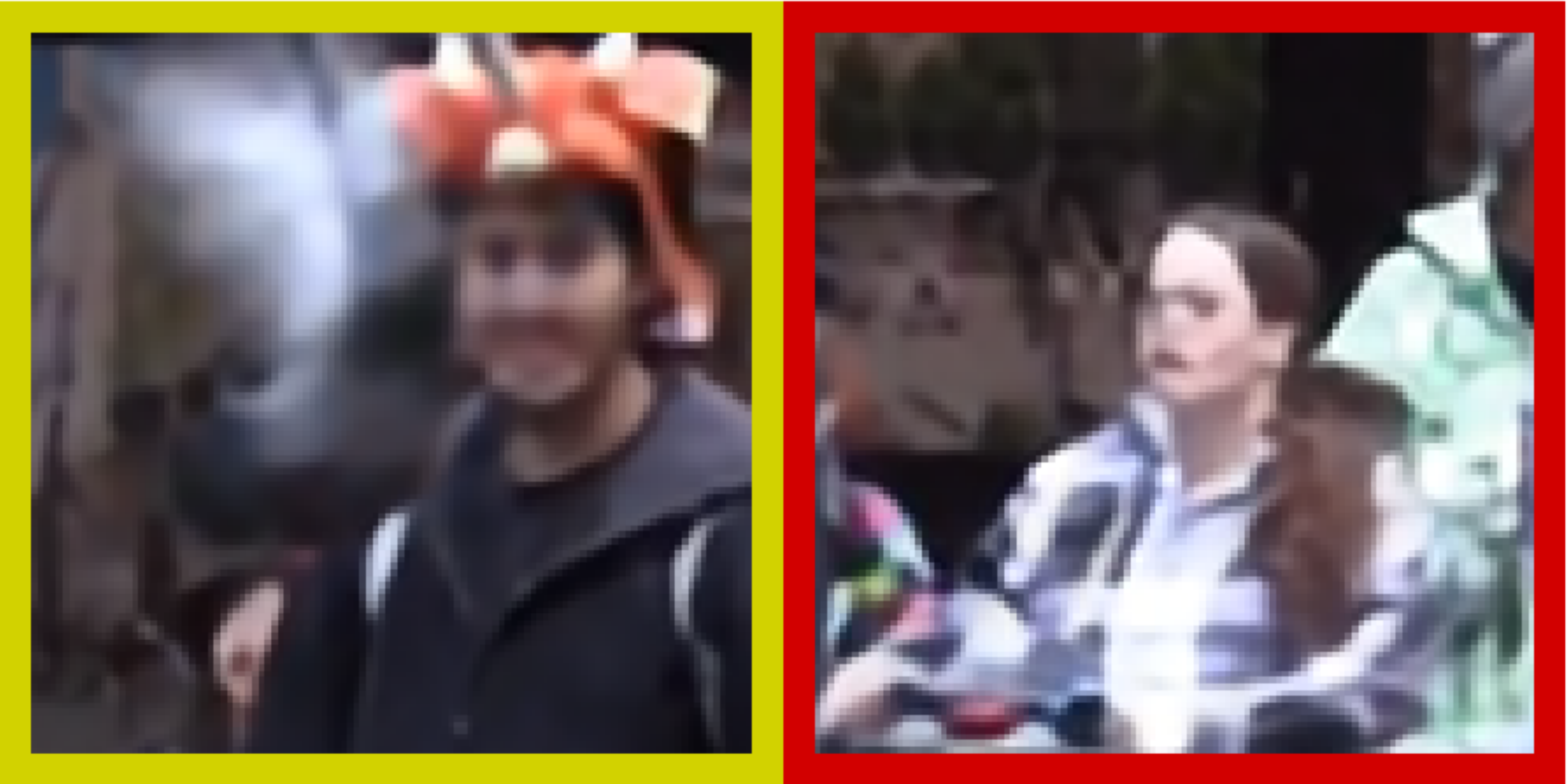}\vspace*{-1.5em}
			\label{fig:Walk_part5}
			\caption{DUF-52L \cite{jo2018deep}}
		\end{subfigure}
		\hspace*{-0.4em}
		\begin{subfigure}[b]{0.14\textwidth}
			\includegraphics[width=\textwidth]{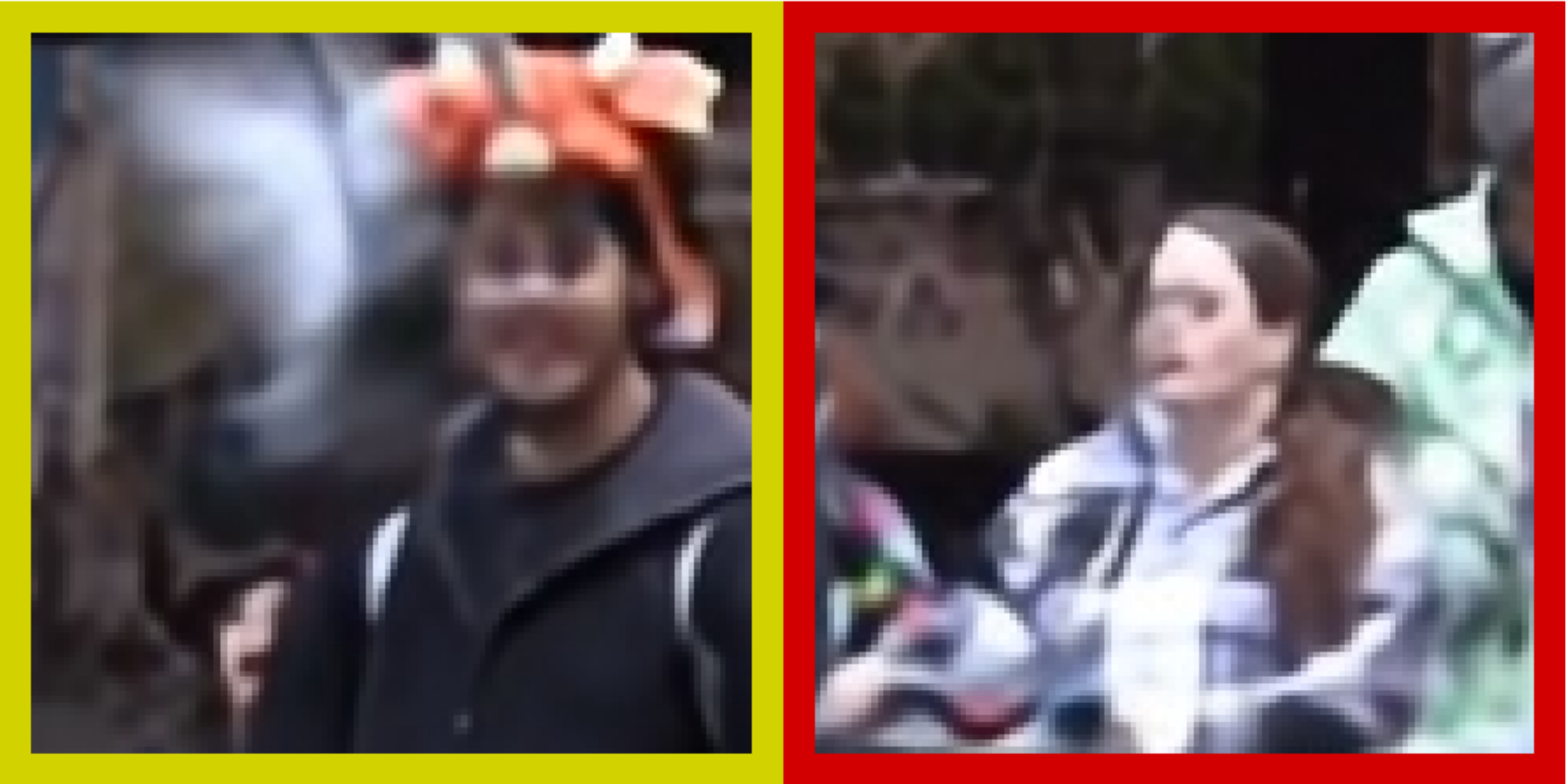}\vspace*{-1.5em}
			\label{fig:Walk_part6}
			\caption{Proposed}
		\end{subfigure}
		\hspace*{-0.4em}
		\begin{subfigure}[b]{0.14\textwidth}
			\includegraphics[width=\textwidth]{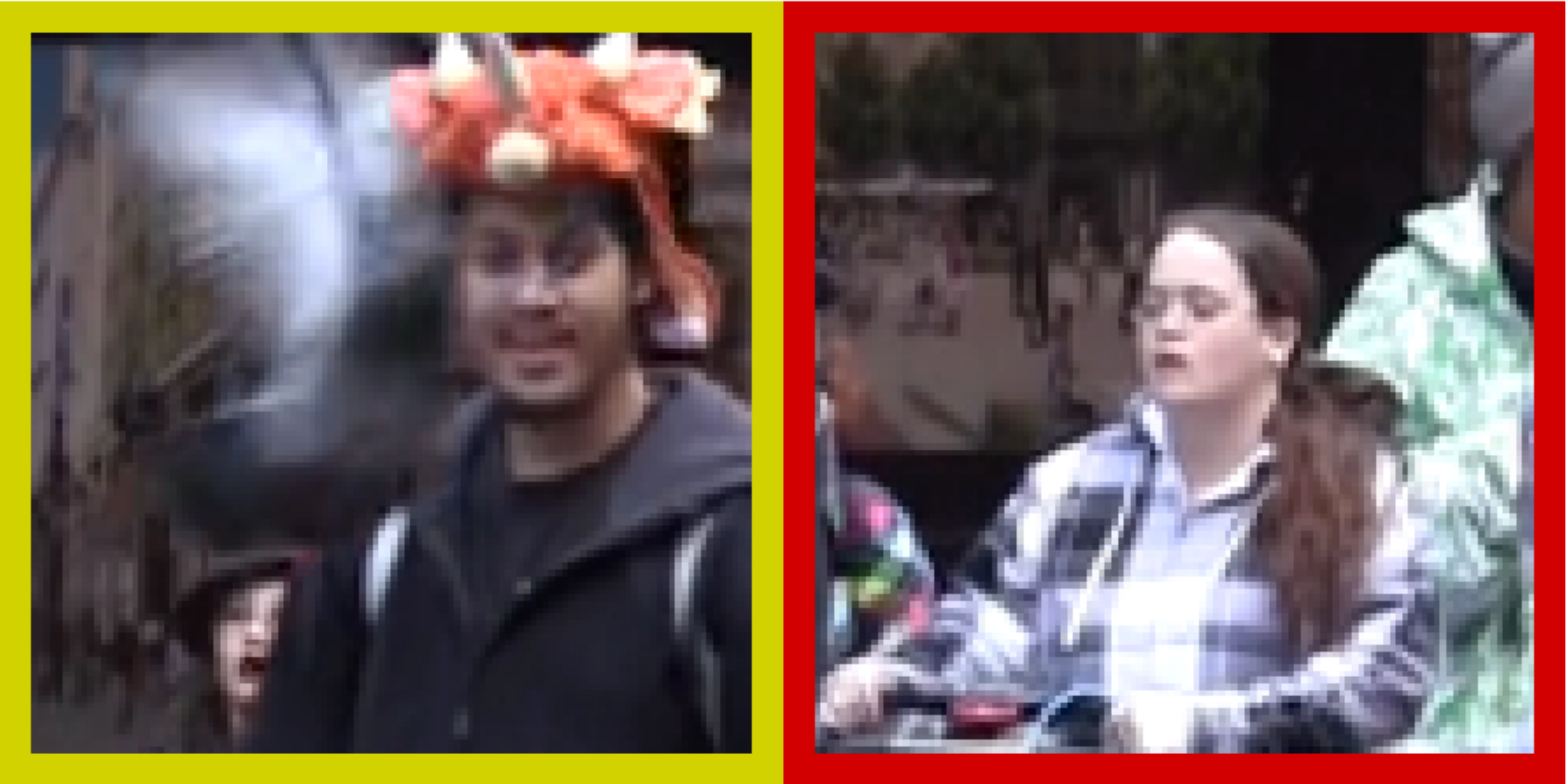}\vspace*{-1.5em}
			\label{fig:Walk_part7}
			\caption{GT}
		\end{subfigure}
		\caption{Visual comparisons on the Vid4 dataset with $r=4$.}
		\label{Fig:vid4}
	\end{center}
\end{figure*}
\begin{figure}[h]
	\centering
	\begin{subfigure}[b]{0.118\textwidth}
		\includegraphics[width=\textwidth]{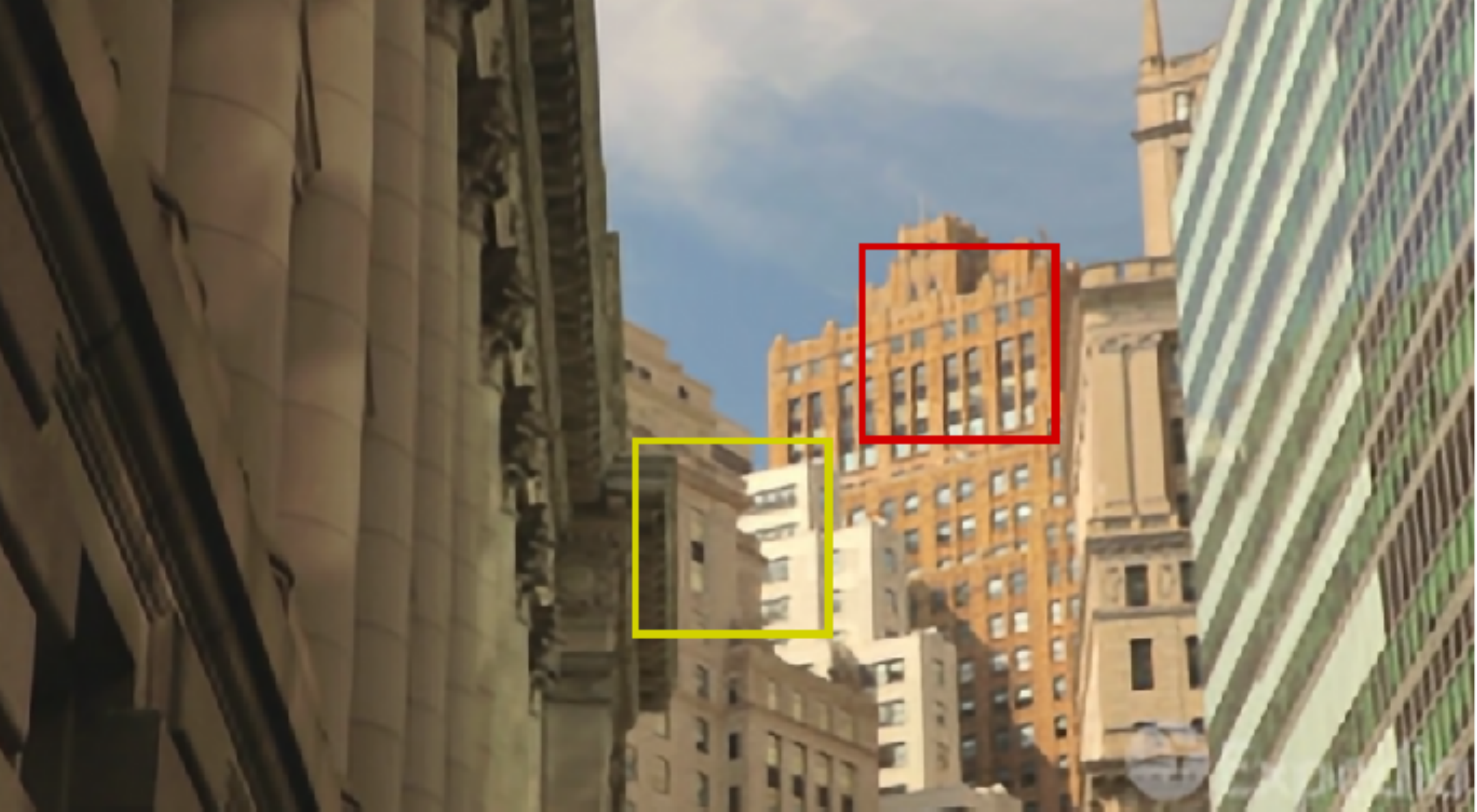}
		\label{fig:NYVTG1}
	\end{subfigure}
	\hspace*{-0.4em}
	\begin{subfigure}[b]{0.118\textwidth}
		\includegraphics[width=\textwidth]{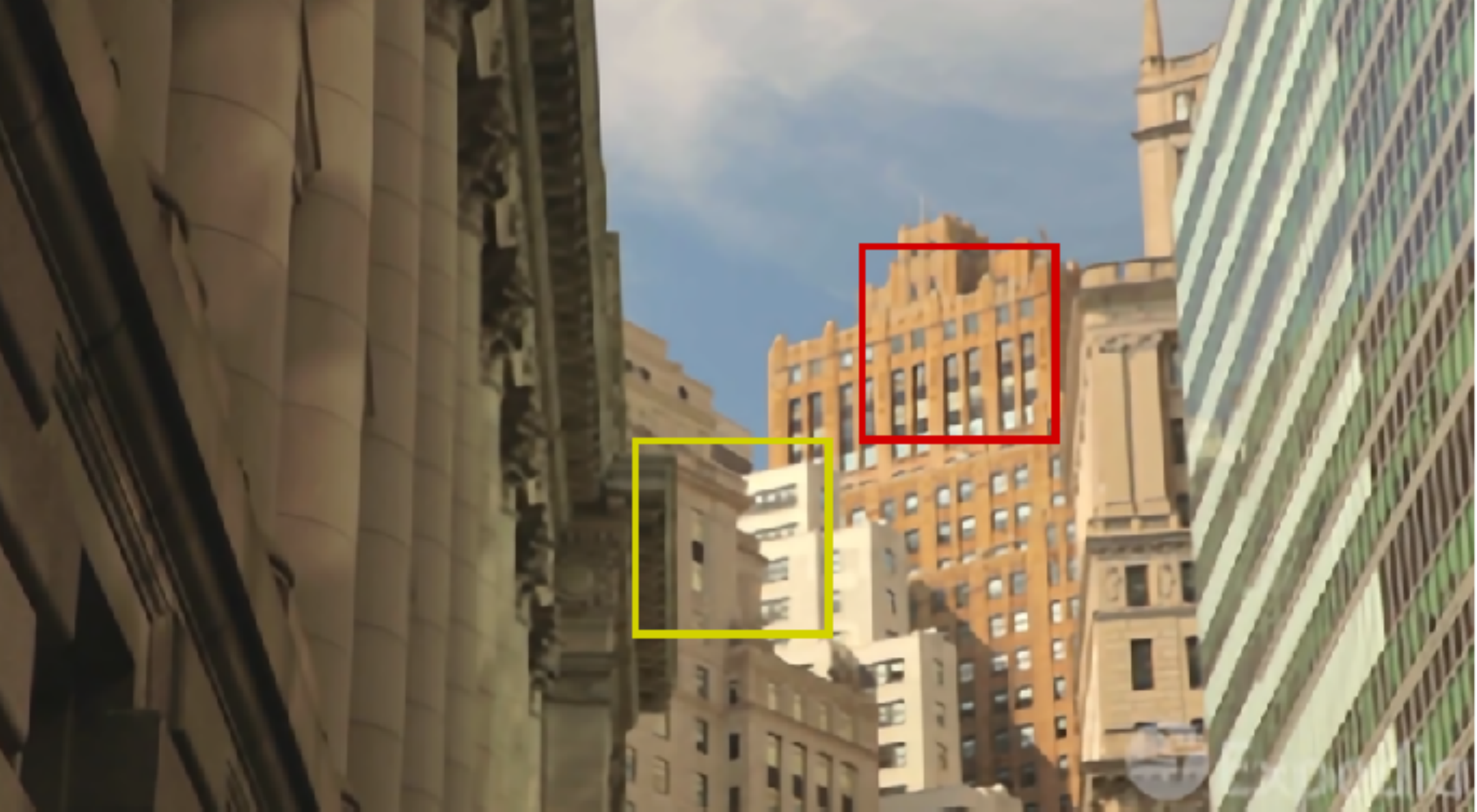}
		\label{fig:NYVTG2}
	\end{subfigure}
	\hspace*{-0.4em}
	\begin{subfigure}[b]{0.118\textwidth}
		\includegraphics[width=\textwidth]{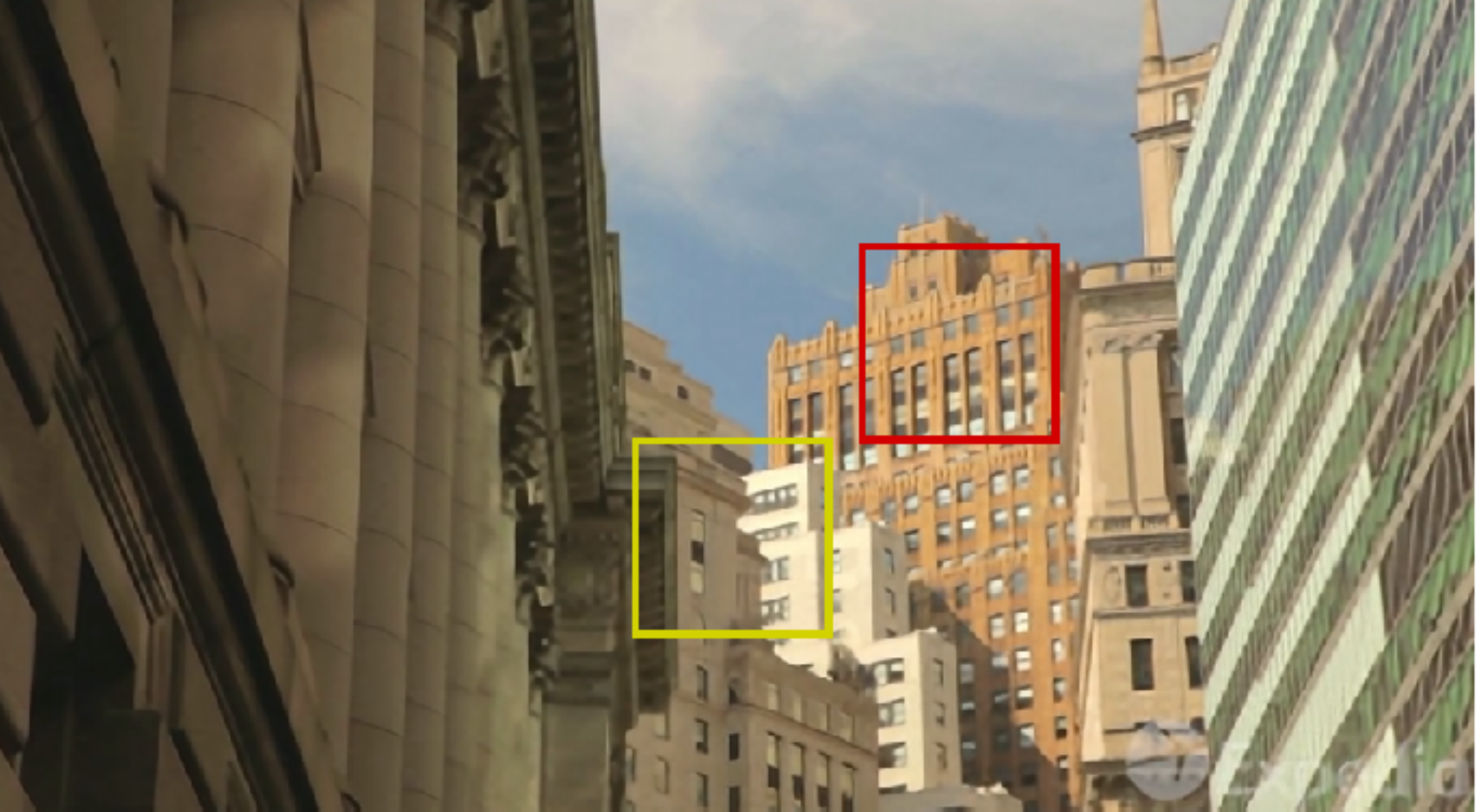}
		\label{fig:NYVTG3}
	\end{subfigure}
	\hspace*{-0.4em}
	\begin{subfigure}[b]{0.118\textwidth}
		\includegraphics[width=\textwidth]{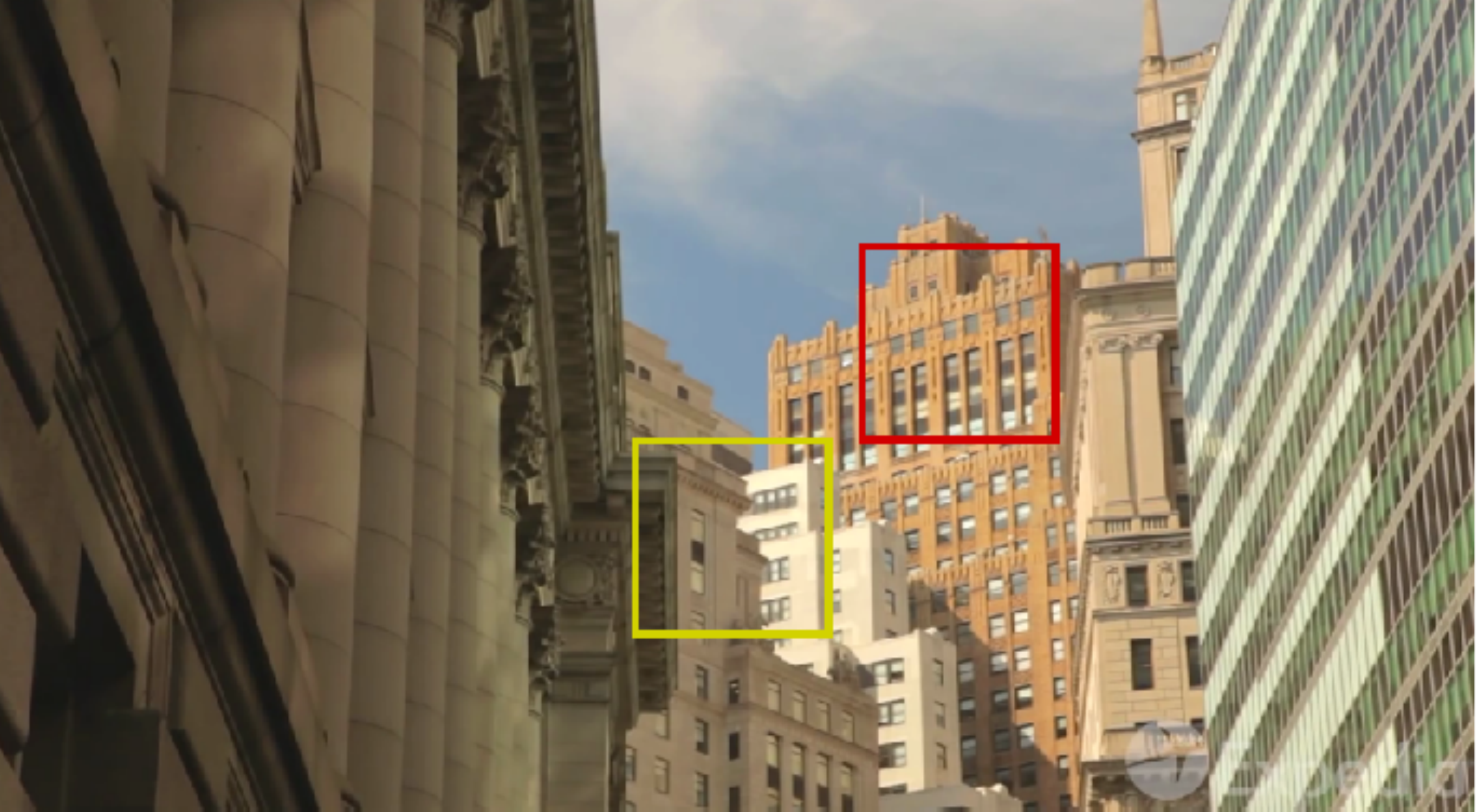}
		\label{fig:NYVTG4}
	\end{subfigure}
	\\
	\vspace*{-1.2em}	
	\begin{subfigure}[b]{0.118\textwidth}
		\includegraphics[width=\textwidth]{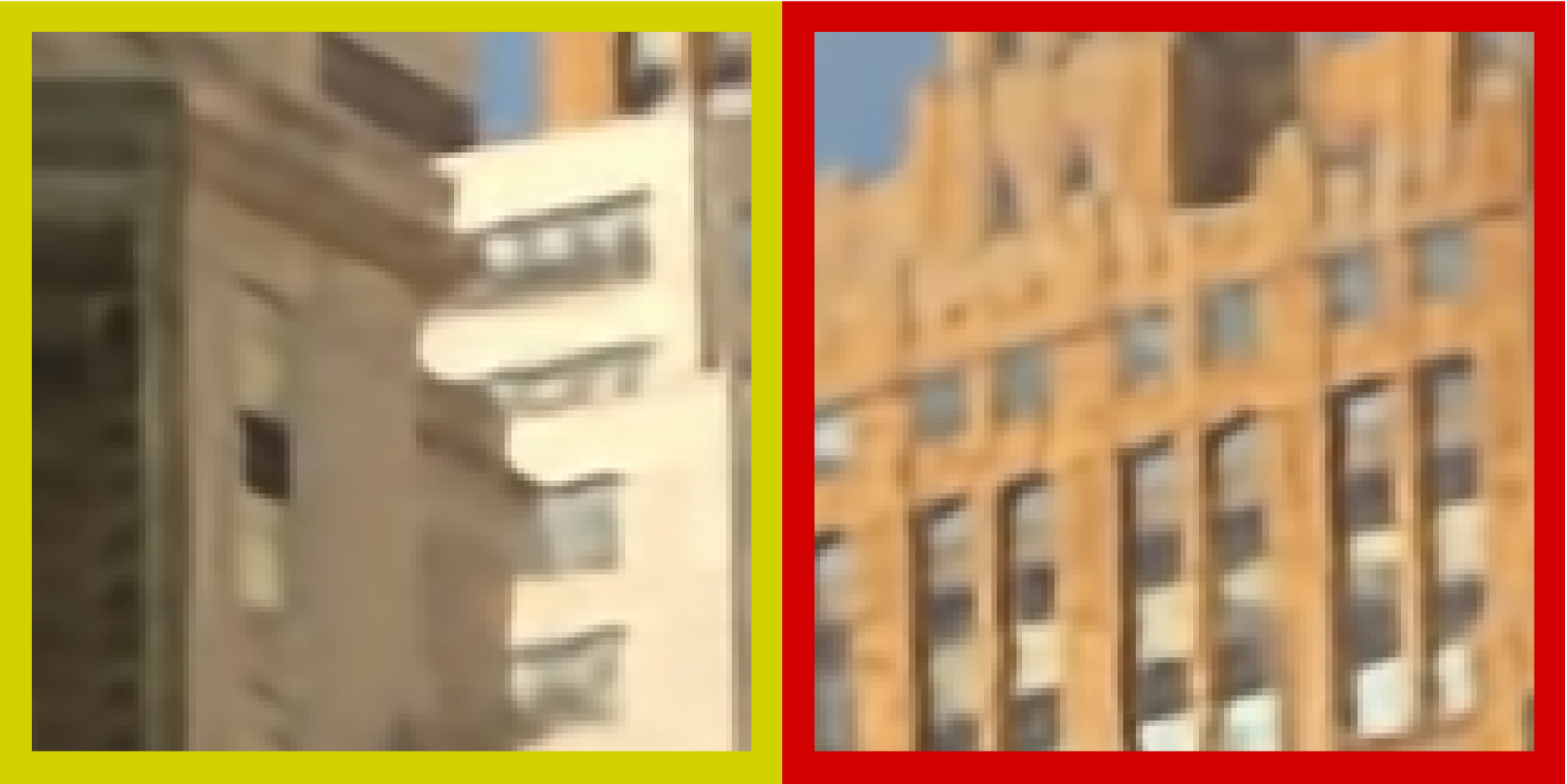}
		\label{fig:NYVTG_part1}
	\end{subfigure}
	\hspace*{-0.4em}
	\begin{subfigure}[b]{0.118\textwidth}
		\includegraphics[width=\textwidth]{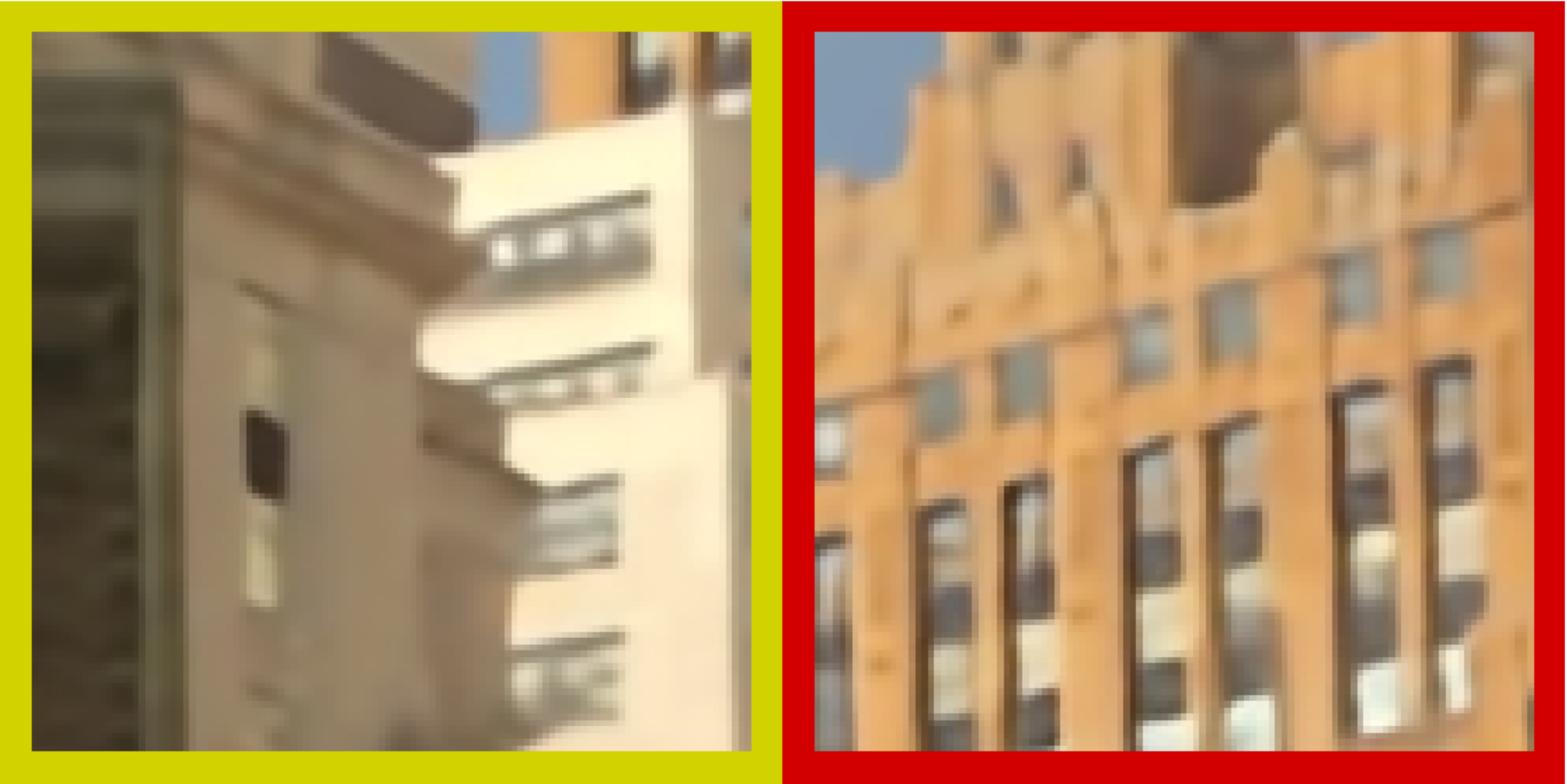}
		\label{fig:NYVTG_part2}
	\end{subfigure}
	\hspace*{-0.4em}
	\begin{subfigure}[b]{0.118\textwidth}
		\includegraphics[width=\textwidth]{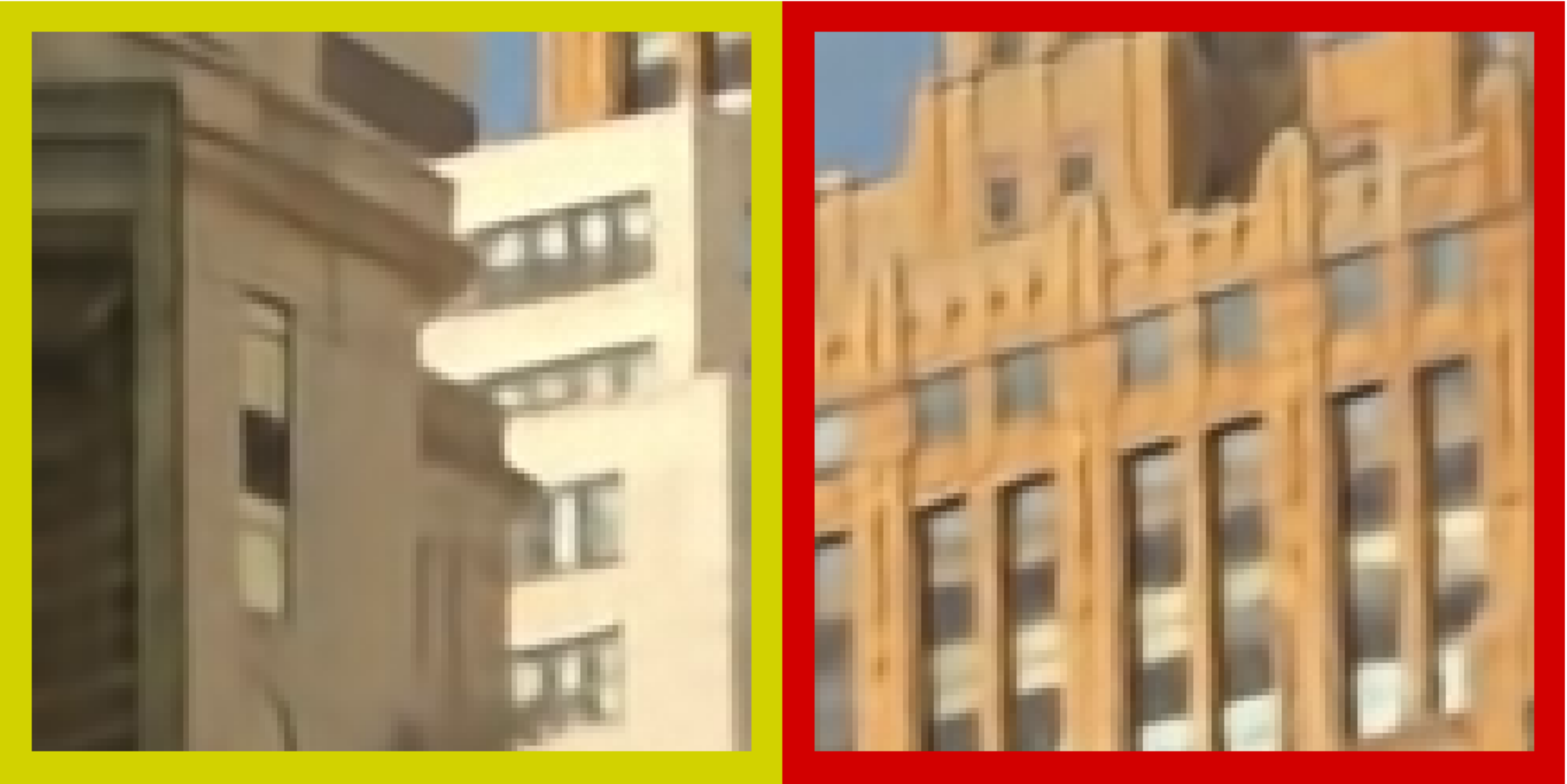}
		\label{fig:NYVTG_part3}
	\end{subfigure}
	\hspace*{-0.4em}
	\begin{subfigure}[b]{0.118\textwidth}
		\includegraphics[width=\textwidth]{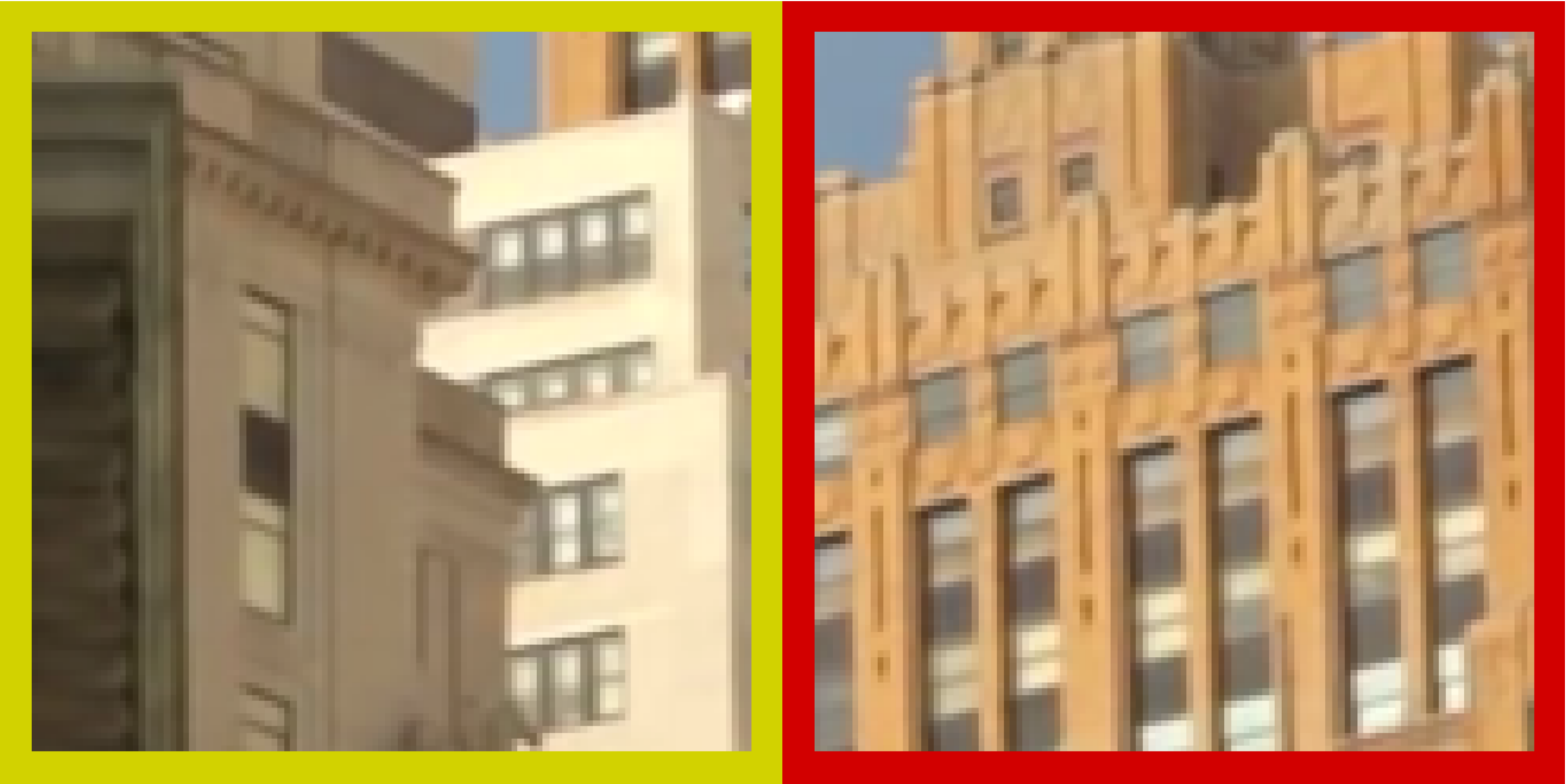}
		\label{fig:NYVTG_part4}
	\end{subfigure}
	\\
	\vspace*{-1em}	
	\begin{subfigure}[b]{0.118\textwidth}
		\includegraphics[width=\textwidth]{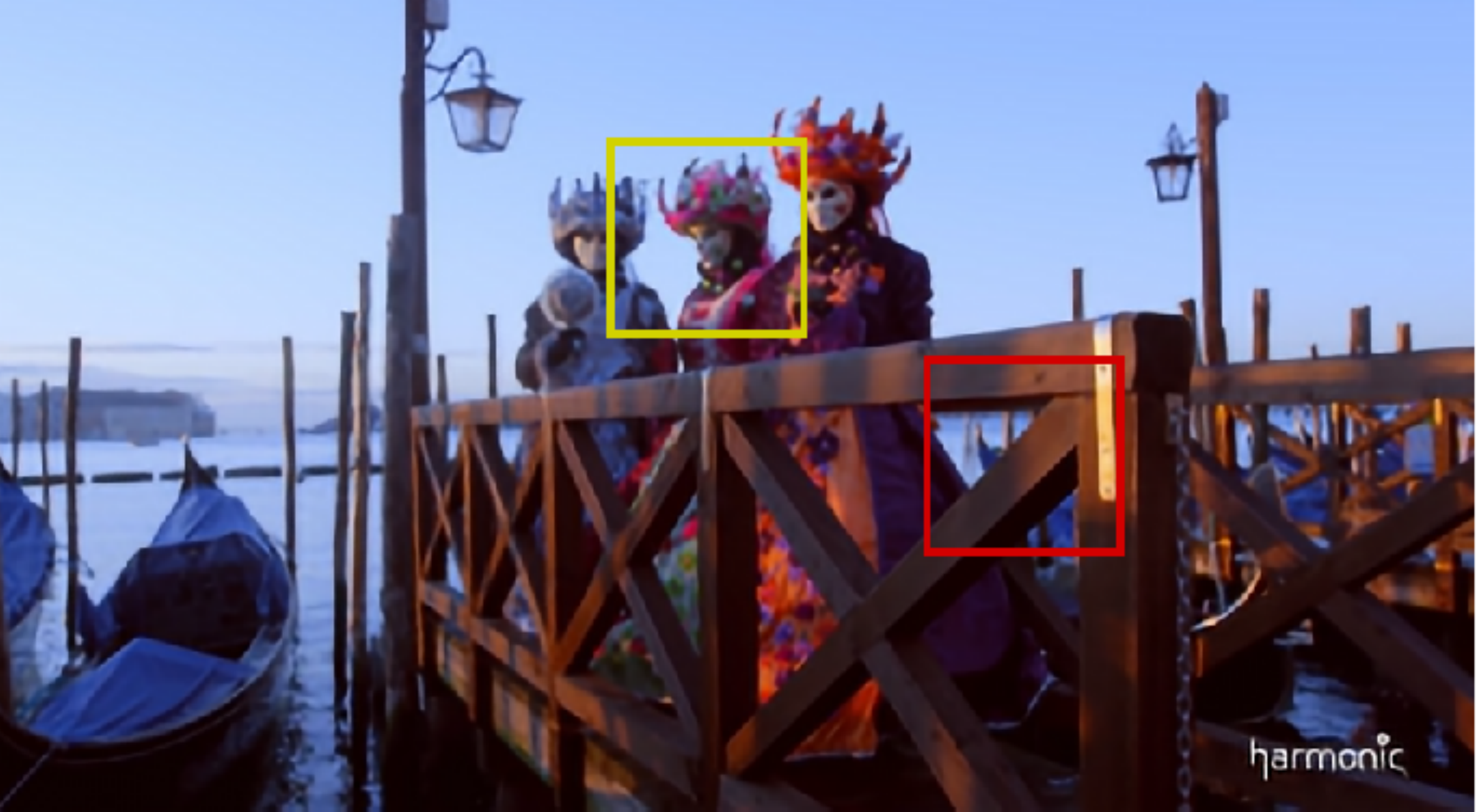}
		\label{fig:veni31}
	\end{subfigure}
	\hspace*{-0.4em}
	\begin{subfigure}[b]{0.118\textwidth}
		\includegraphics[width=\textwidth]{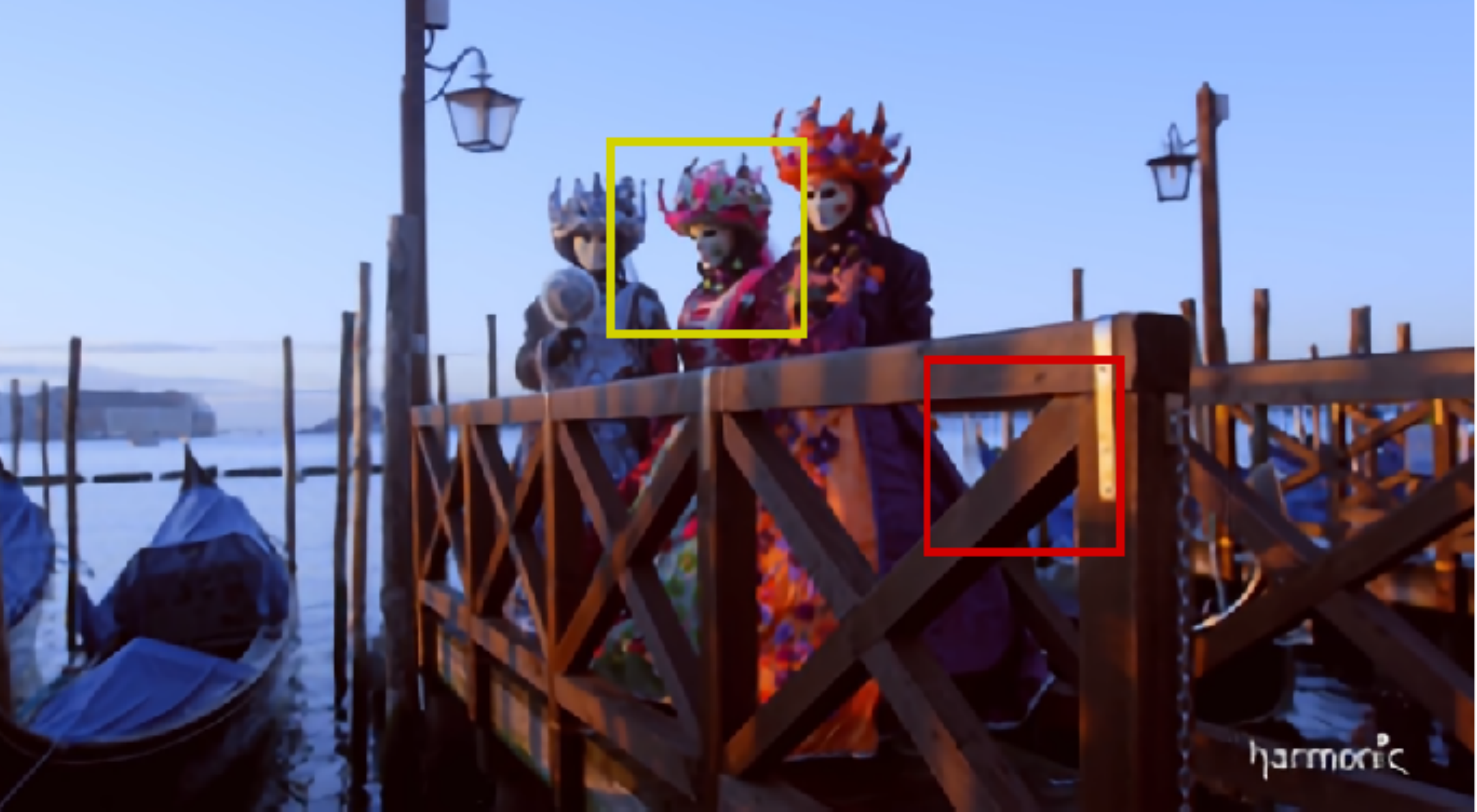}
		\label{fig:veni32}
	\end{subfigure}
	\hspace*{-0.4em}
	\begin{subfigure}[b]{0.118\textwidth}
		\includegraphics[width=\textwidth]{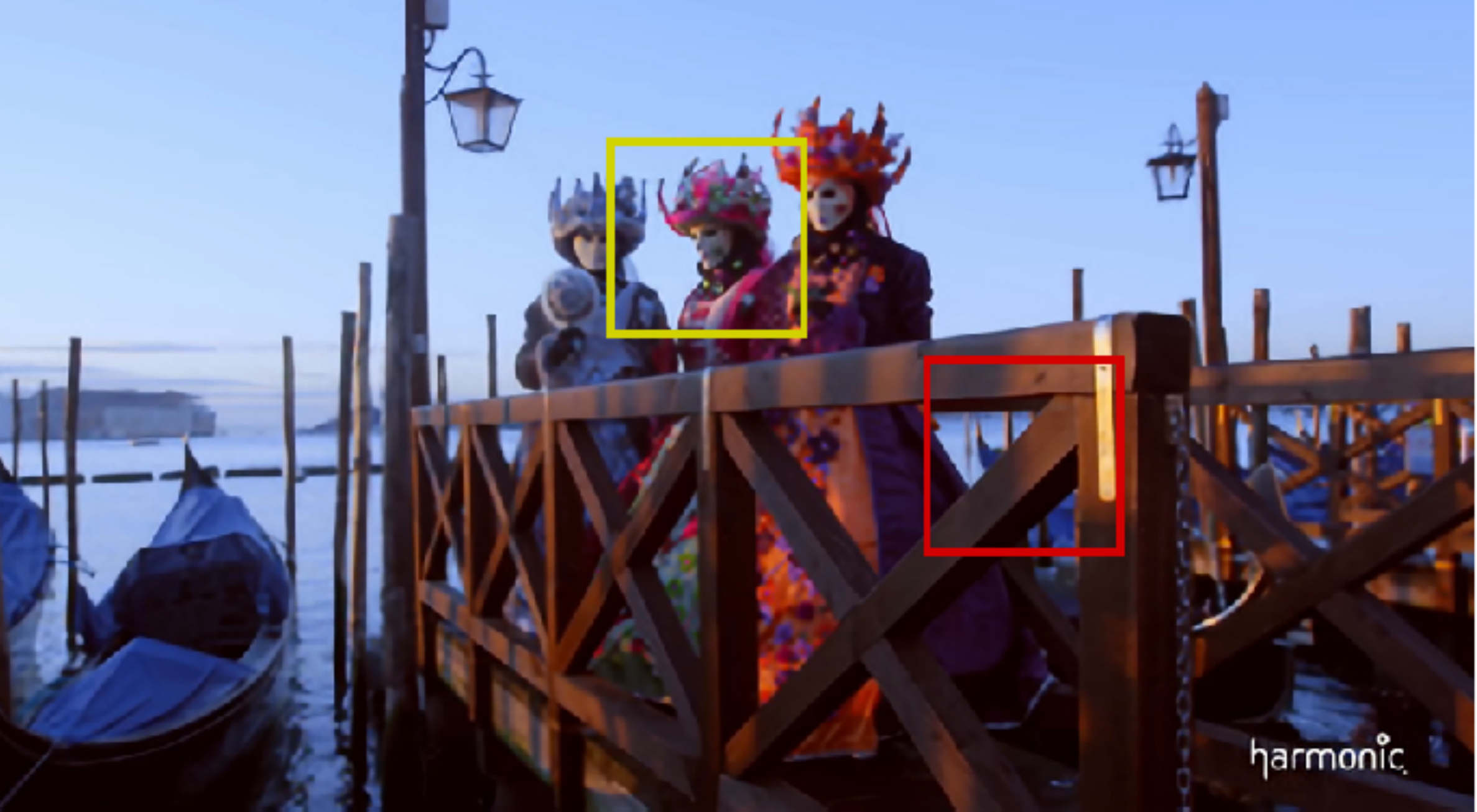}
		\label{fig:veni33}
	\end{subfigure}
	\hspace*{-0.4em}
	\begin{subfigure}[b]{0.118\textwidth}
		\includegraphics[width=\textwidth]{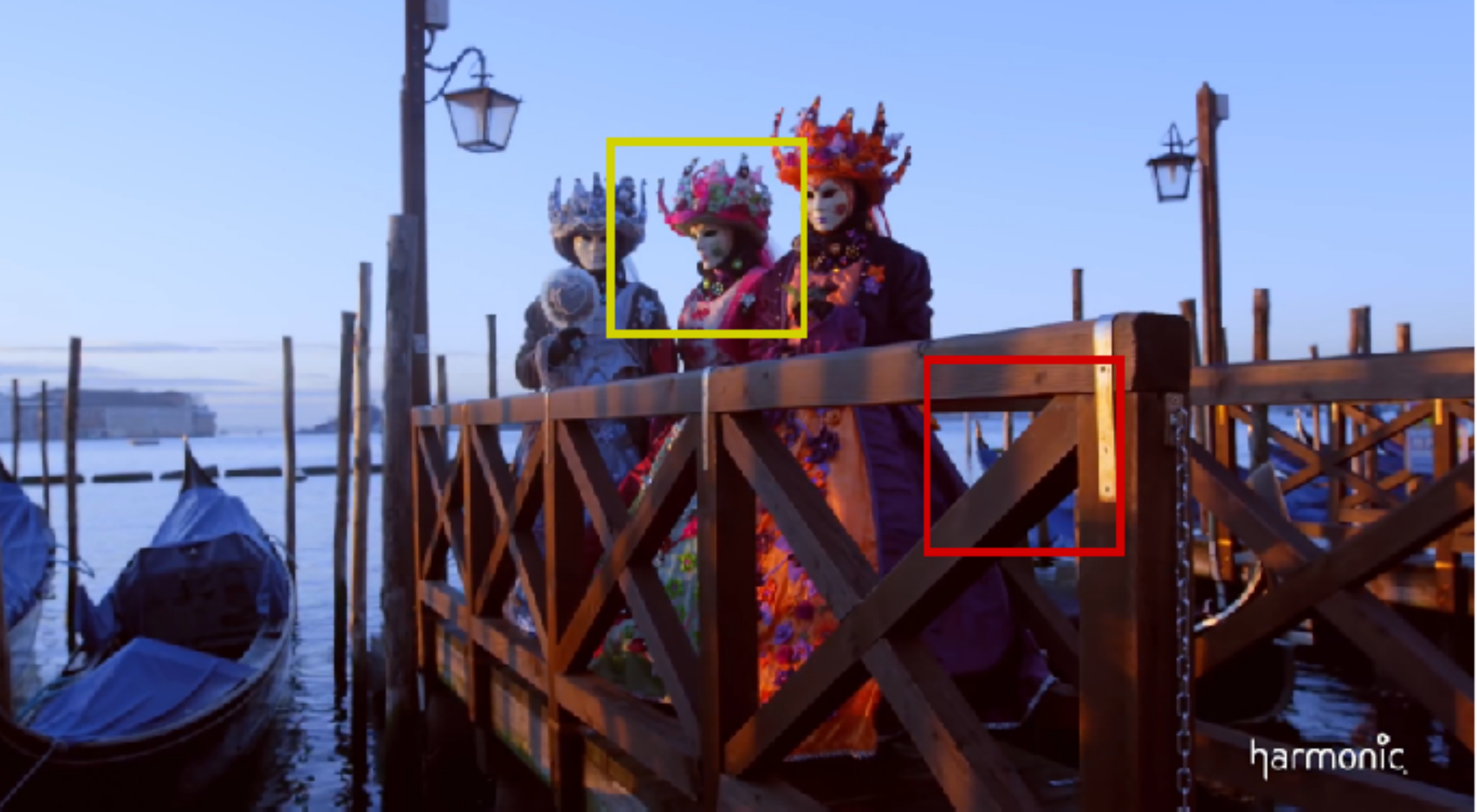}
		\label{fig:veni34}
	\end{subfigure}
	\\
	\vspace*{-1.2em}	
	\begin{subfigure}[b]{0.118\textwidth}
		\includegraphics[width=\textwidth]{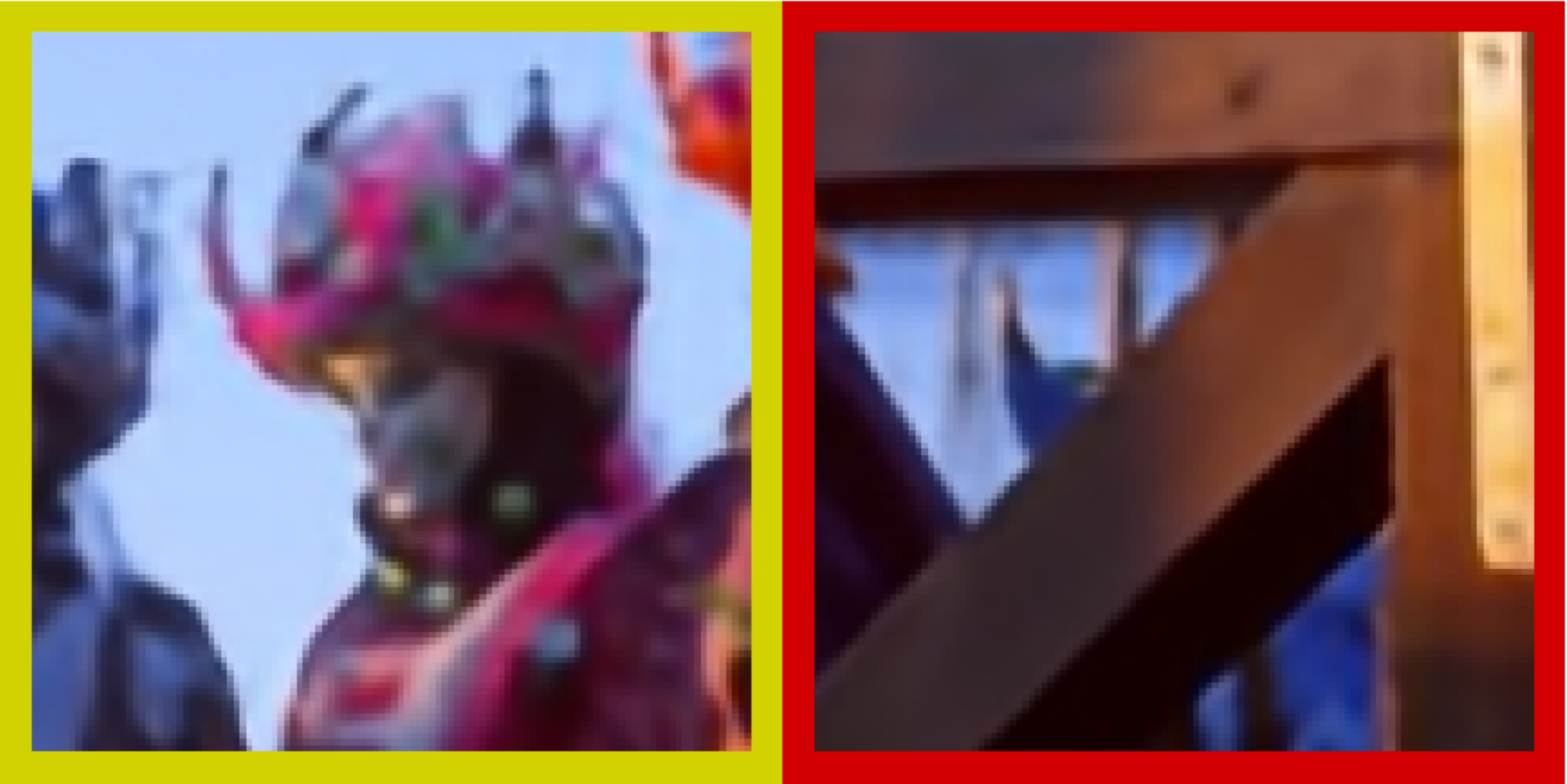}\vspace*{-1.5em}
		\label{fig:veni3_part1}
		\caption{SPMC \cite{tao2017detail}}
	\end{subfigure}
	\hspace*{-0.4em}
	\begin{subfigure}[b]{0.118\textwidth}
		\includegraphics[width=\textwidth]{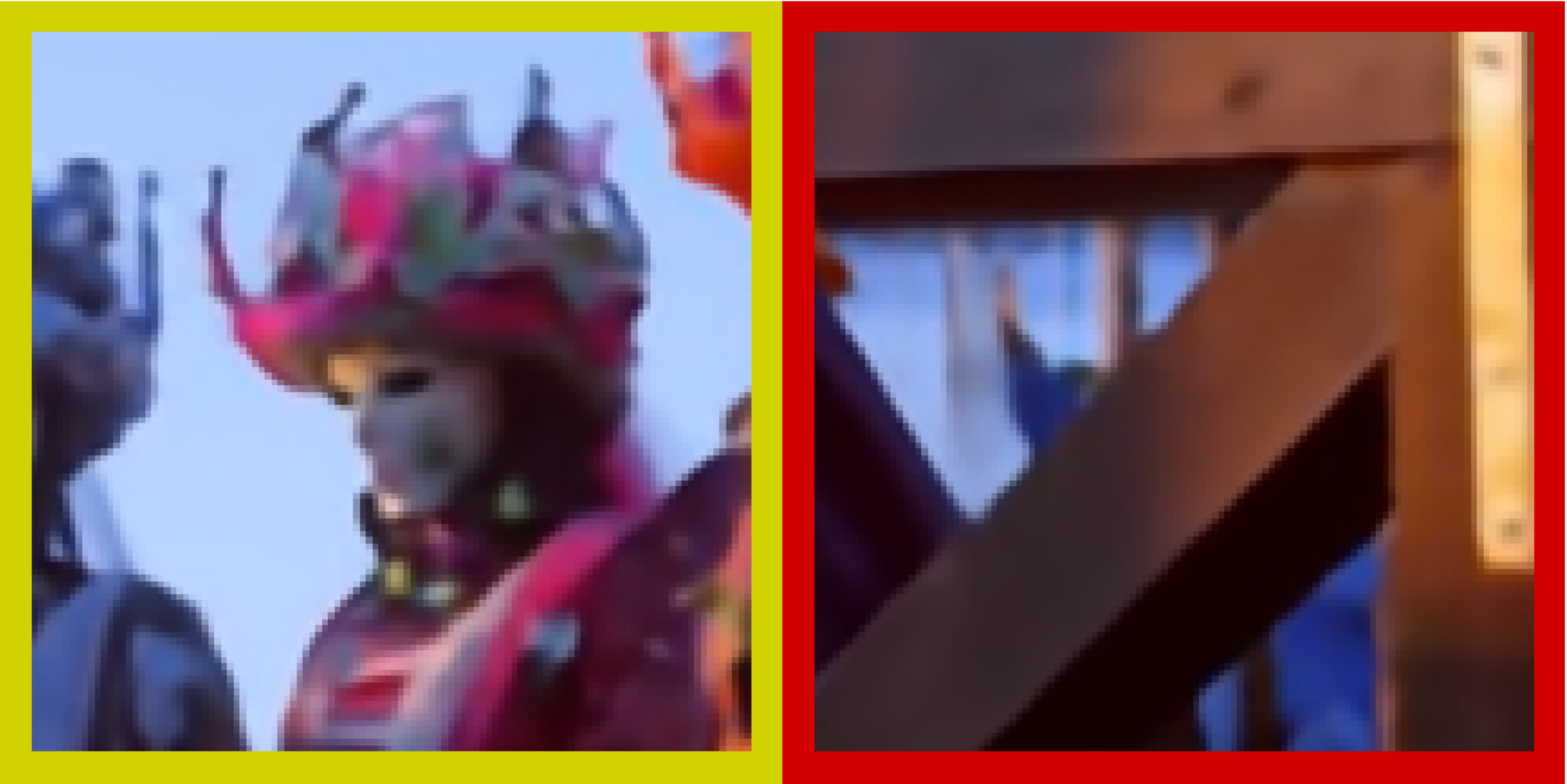}\vspace*{-1.5em}
		\label{fig:veni3_part2}
		\caption{DUF52 \cite{jo2018deep}}
	\end{subfigure}
	\hspace*{-0.4em}
	\begin{subfigure}[b]{0.118\textwidth}
		\includegraphics[width=\textwidth]{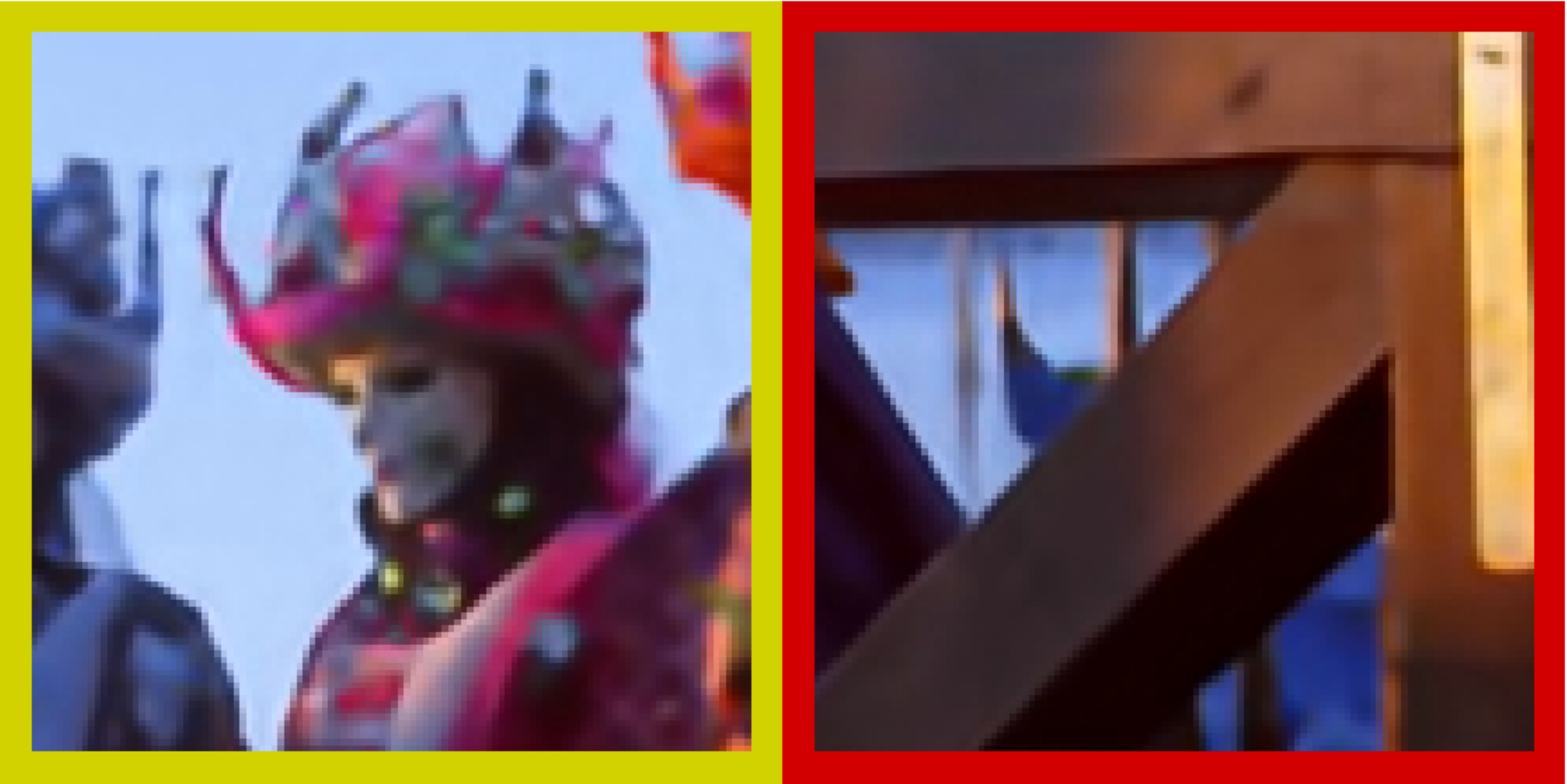}\vspace*{-1.5em}
		\label{fig:veni3_part3}
		\caption{Proposed}
	\end{subfigure}
	\hspace*{-0.4em}
	\begin{subfigure}[b]{0.118\textwidth}
		\includegraphics[width=\textwidth]{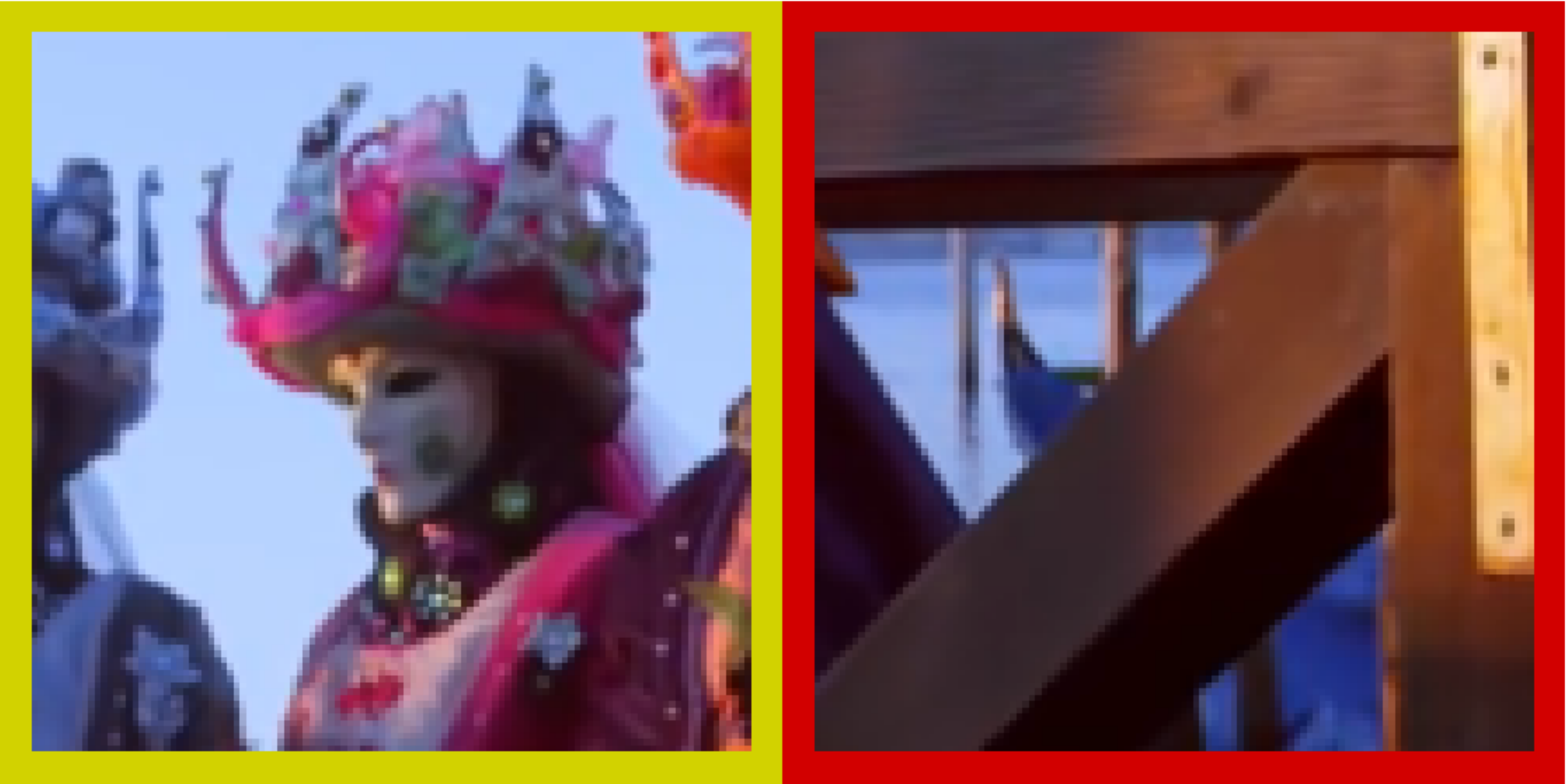}\vspace*{-1.5em}
		\label{fig:veni3_part4}
		\caption{GT}
	\end{subfigure}
	\\
	\vspace*{-1em}

	\caption{Visual comparisons for \textit{NewYork} and \textit{Venice} from the SPMCS dataset with  $r=4$.}
	\label{fig:tao_dataset}
\end{figure}
\subsection{Analysis of Temporal Consistency}
The Lack of temporal consistency may cause flickering artifacts that manifest visually, for instance, in the form of jagged edges. To validate that our proposed method maintains temporal consistency, we extract spatially co-located rows from 
consecutive super-resolved frames of \textit{Calendar} in the Vid4 dataset with the scale ratio $r=4$, and arrange them vertically to compose a temporal profile \cite{sajjadi2018frame}; we also compose a temporal profile for the \textit{City} frames in the same dataset. Fig.~\ref{fig:temporal} shows the comparison of our temporal profiles and the corresponding ones generated by five existing VSR methods. Overall, the proposed method presents the most consistent temporal profiles.
In fact, its maintenance of temporal consistency remains good even in some cases where the ground-truth frames have certain artifacts in this respect (see, e.g., \textit{City}). Moreover, it can be seen from the temporal profiles that our method produces the sharpest edges, as compared to the other ones, in all \textit{Calendar} and \textit{City} frames.
\subsection{Qualitative and Quantitative Comparisons}
The proposed method is compared to several existing VSR methods qualitatively and quantitatively with a particular focus on the \textit{VSRDUF}, which is the current state-of-the-art. 
In the \textit{VSRDUF}, the generated dynamic upsampling filters (DUF) are only applied on the central LR frame to reconstruct the SR one without joint consideration of all input frames. This mechanism would cause fuzzy edges and temporal inconsistencies (see Fig.~\ref{fig:temporal} (e)). In contrast, the generated dynamic local filters (DLF) are applied on all input LR frames to reconstruct the SR one, yielding better visual quality in terms of edge sharpness and temporal consistency (see Fig.~\ref{fig:temporal} (f)), and achieving higher PSNR and SSIM values. Moreover, the essence of the DUF is still a weight-sharing filter that has been widely used in conventional CNNs. However, the proposed DLF is spatially content adaptive (position-specific) within each frame and temporally distinct (sample-specific) among different frames. This is a new mechanism and is not like conventional CNNs. The reason we employ the DLF in VSR is based on the observation that each pixel in a video frame may exhibit a unique degradation pattern, which cannot be effectively exploited by the weight-sharing filters. Fig.~\ref{Fig:vid4} shows the qualitative comparisons of different methods on the Vid4 dataset with $r=4$, and Table~\ref{tab:vid4} demonstrates the quantitative comparisons in terms of average PSNR and SSIM values on the same dataset with $r=3,4$. Note that our SR results contain more fine details and restore sharper edges. Moreover, the PSNR value achieved by the proposed method is 0.61 dB  higher than that by the \textit{DUF-16L} when $r=3$ and  0.66 dB higher when $r=4$. Even compared with the  \textit{DUF-52L}, our result is still 0.13 dB higher when $r=4$. 
We have also performed the test on the SPMCS dataset, which contains 31 video clips, for further qualitative and quantitative comparisons. It can seen from Fig.~\ref{fig:tao_dataset} and Table~\ref{tab:tao} that the proposed method performs competitively on this dataset as well.

\begin{table*}[t]
	\centering
	\begin{adjustbox}{max width=\textwidth}
		\begin{tabular}{ |C{1.5cm} | C{1.5cm} || C{1.5cm} | C{1.5cm} | C{1.5cm} | C{1.5cm} | C{1.5cm} | C{1.5cm} |C{1.5cm} | C{1.5cm} | C{1.5cm} | C{1.5cm}| }
			\hline
			Vid4 & Metric & \textit{Bicubic} & \textit{Bayesian} \cite{liu2014bayesian} & \textit{VSRNet} \cite{kappeler2016video}& \textit{VESPCN} \cite{caballero2017real} & \textit{$B_{1,2,3}+T$} \cite{liu2017robust} & \textit{SPMC} \cite{tao2017detail} & \textit{FRVSR} \cite{sajjadi2018frame} & \textit{DUF-16L} \cite{jo2018deep} & \textit{DUF-52L} \cite{jo2018deep} &\textit{Proposed} \tstrut\bstrut\\
			\hline\hline
			\multirow{2}{*}{x3} & \bf PSNR & 25.28 & 25.82 & 26.79 & 27.25  & - & 27.49 & - & 28.90 & -  & \bf{29.51} \tstrut\bstrut\\
			& \bf SSIM & 0.7329 & 0.8323 & 0.8098 & 0.8447 & - & 0.8400 & - & 0.8898 & - & \bf{0.8964} \tstrut\bstrut\\ 
			\hline
			\multirow{2}{*}{x4} & \bf PSNR & 23.79 & 25.06 & 24.84 & 25.35 & 25.39 & 25.52 & 26.69 & 26.81 &27.34 & \bf{27.47} \tstrut\bstrut\\
			& \bf SSIM & 0.6332 & 0.7466 & 0.7049 & 0.7557 & 0.7490 & 0.7600 & 0.8220 & 0.8145 &0.8327 & \bf{0.8394} \tstrut\bstrut\\ 
			\hline
		\end{tabular}
	\end{adjustbox}
	\caption{Quantitative comparisons on the Vid4 dataset with $r=3,4$.}
	\label{tab:vid4}
\end{table*}
\begin{table}[h]
	\centering
	\resizebox{0.48\textwidth}{!}{
		\begin{tabular}{|c|c||c|c|c|c|}
			\hline
			SPMCS & Metric  & \textit{SPMC} \cite{tao2017detail} & \textit{DUF-16L} \cite{jo2018deep} & \textit{DUF-52L} \cite{jo2018deep} & \textit{Proposed} 
			\tstrut\\ 
			\hline\hline
			\multirow{2}{*}{x3} & \bf{PSNR} &32.10  & -  & - & \bf{33.91}
			\tstrut\\
			& \bf{SSIM} &0.9000  & - & - &\bf{0.9358}
			\tstrut\\
			\hline
			\multirow{2}{*}{x4} & \bf{PSNR} &29.89  &30.01 &30.39 &\bf{30.66}
			\tstrut\\
			& \bf{SSIM} &0.8400  &0.8355 &0.8646 &\bf{0.8711}
			\tstrut\\
			\hline
	\end{tabular}}
	\caption{Quantitative comparisons on the SPMCS dataset with $r=3,4$.}
	\label{tab:tao}
\end{table}

\begin{table}[t]
	\centering
	\resizebox{0.48\textwidth}{!}{
		\begin{tabular}{|c|c||c|c|c|c|}
			\hline
			Vid4 & Metric  & w/o LC Layers & w/o GRN & w/ U-Net \cite{ronneberger2015u} & Our full model
			\tstrut\\ 
			\hline\hline
			\multirow{2}{*}{x3} & \bf{PSNR} &27.27  & 28.13 & 29.20 & \bf{29.51} 
			\tstrut\\
			& \bf{SSIM} &0.8471  &0.8752 &0.8896 & \bf{0.8964} 
			\tstrut\\
			\hline
			\multirow{2}{*}{x4} & \bf{PSNR} &25.45  &26.20 &27.29 &\bf{27.47} 
			\tstrut\\
			& \bf{SSIM} &0.7530  &0.8134 &0.8352 &\bf{0.8394} 
			\tstrut\\
			\hline
	\end{tabular}}
	\caption{Quantitative comparisons on the Vid4 dataset for different variants of the proposed method with $r=3,4$.}
	\label{tab:variants}
\end{table}

\subsection{Ablation Study} \label{experiment_network_size}
To gain a better understanding of the effectiveness of the new mechanism for implicit motion estimation and compensation, we conduct an ablation study by directly feeding the output of the LFGN (via a convolutional layer) to the pixel-shuffle network; the resulting system is trained in the same way as before. Compared to the original system, this new system has (essentially) the same number of parameters but bypasses the action of dynamic local filters, via LC layers, on the input LR frames. As shown in Table~\ref{tab:variants}, this modification leads to significant performance degradation on the Vid4 dataset, suggesting that the mechanism adopted by the original system is more effective in terms of exploiting the  
learning capability of the LFGN  to realize
dynamic localized functionalities. We also conducted the analysis on the role of the GFN by 1) removing it from the proposed system and 2) replacing it with the U-Net \cite{ronneberger2015u}. It can be seen from Table~\ref{tab:variants} that these two variants incur significant performance loss compared with our full model. 


\begin{figure}[h]
	\begin{center}
		\includegraphics[width=0.8\linewidth]{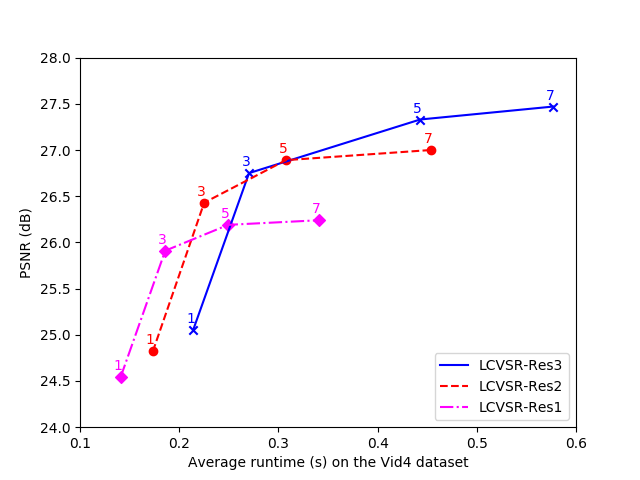}
	\end{center}
	\caption{PSNR vs. runtime for different configurations of the LCVSR system on the Vid4 dataset with the input length set to be 1, 3, 5 and 7.}
	\label{fig:time_analysis}
\end{figure}

We further investigate the performance-complexity trade-off for the proposed system by varying the input length and the network size. Specifically, by employing 1, 2 and 3 ResBlocks (the ones without shape change) in both DLFN and GRN, we construct three different configurations of the proposed system, denoted by LCVSR-Res1, LCVSR-Res2 and LCVSR-Res3, respectively. Fig.~\ref{fig:time_analysis} plots the PSNR against the average runtime for each configuration on
 the Vid4 dataset with the number of input LR frames set to be 1, 3, 5 and 7.  It can be seen that increasing the input length  leads to higher PSNRs at the cost of longer runtimes. When the input length is 1, the VSR task degenerates to the SISR task, which only exploits intra-frame dependencies. Increasing the input length from 1 to 3 significantly improves the PSNR values due to the additional freedom of exploring inter-frame dependencies via
 motion estimation and compensation. Further increasing the input length provides more spatial and temporal information that can be capitalized on, which helps to generate better SR results. However the improvement becomes negligible when the input length goes beyond 7. Employing more ResBlocks in DLFN and GRN has a similar effect. Indeed, it can be seen from Fig.~\ref{fig:time_analysis} that the PSNR value increases progressively from LCVSR-Res1 to LCVSR-Res3 for the same input length, and the runtime follows the same trend. 
 Note that the proposed LCVSR system corresponds to LCVSR-Res3 with input length 7. The above experimental results provide certain justifications for the design of the proposed system in consideration of the performance-complexity trade-off.

\begin{table}[t]
	\centering
	\resizebox{0.48\textwidth}{!}{
		\begin{tabular}{|c|c|c|c|c|c|}
			\hline
			Method & \textit{VSRNet} \cite{kappeler2016video}  & \textit{VESPCN} \cite{caballero2017real} & \textit{SPMC} \cite{tao2017detail} & \textit{DUF-52L} \cite{jo2018deep} & \textit{Proposed}
			\tstrut\\ 
			\hline\hline
			Params. & 0.39M &0.89M  &2.17M &5.82M & 5.81M
			\tstrut\\
			\hline
			Time (s) & 0.23 &0.29  &2.80 &2.68 &2.32
			\tstrut\\
			\hline
	\end{tabular}}
	\caption{The number of parameters and average runtime of different methods for 1080p frames.}
	\label{tab:time}
	\vspace{-4mm}
\end{table}

\subsection{Network parameters and runtime Analysis}
In table~\ref{tab:time}, the number of parameters and the runtime of different methods are demonstrated. Our method has slightly less parameters and faster runtime than the current state-of-the-art DUF-52L. Although the VSRNet has the least number of parameters and is close to real-time, it has the worst VSR performance.

\section{Conclusion}
In this paper, we have proposed an end-to-end trainable VSR method 
based on a new mechanism  for implicit motion estimation and compensation (realized through dynamic local filters and LC layers).  Our experimental results demonstrate that the proposed method outperforms the current state-of-the-art in terms of local transformation handling, edge sharpness and temporal consistency. 


\clearpage
\newpage

{\small
\bibliographystyle{ieee}
\bibliography{egbib}
}

\end{document}